\documentclass[10pt,twocolumn,letterpaper]{article}

\PassOptionsToPackage{table}{xcolor}
\usepackage[pagenumbers]{iccv} %

\usepackage{tabularx}
\usepackage{multirow}
\usepackage[per-mode=symbol, group-digits=integer, detect-all=true, input-symbols={()}]{siunitx}

\usepackage[percent]{overpic} %
\usepackage{fp} %
\usepackage{graphicx}
\usepackage{tikz}
\usetikzlibrary{positioning}
\usetikzlibrary{shapes.geometric, shapes, arrows, arrows.meta, bending, shapes.geometric, shadings}
\usetikzlibrary{backgrounds}
\usetikzlibrary{spy}
\usepackage{pgfplots}
\usepackage{bbding}
\usepackage{dsfont}
\usepackage{fontawesome5}
\usepackage{calc}
\pgfplotsset{compat=1.17}
\usepackage[outline]{contour} 
\contourlength{0.02em}
\usepackage[table]{xcolor}
\usepackage{makecell}
\usepackage{pifont}%
\usepackage{gradient-text}
\usepackage{newtxtt}

\newcommand*{\inparagraph}[1]{\smallskip\noindent\textbf{#1}\hspace{0.4em}}

\newcommand*{\inparagraphnohspace}[1]{\smallskip\noindent\textbf{#1}}

\definecolor{iccvblue}{rgb}{0.21,0.49,0.74}
\definecolor{tud0d}{RGB}{83,83,83}
\definecolor{tud0c}{RGB}{137,137,137}
\definecolor{tud0b}{RGB}{181,181,181}
\definecolor{tud0a}{RGB}{220,220,220}
\definecolor{tud1a}{RGB}{93,133,195}
\definecolor{tud2a}{RGB}{0,156,218}
\definecolor{tud3a}{RGB}{80,182,149}
\definecolor{tud4a}{RGB}{175,204,80}
\definecolor{tud5a}{RGB}{221,223,72}
\definecolor{tud6a}{RGB}{255,224,92}
\definecolor{tud7a}{RGB}{248,186,60}
\definecolor{tud8a}{RGB}{238,122,52}
\definecolor{tud9a}{RGB}{233,80,62}
\definecolor{tud10a}{RGB}{201,48,142}
\definecolor{tud11a}{RGB}{128,69,151}
\definecolor{tud1b}{RGB}{0,90,169}
\definecolor{tud2b}{RGB}{0,131,204}
\definecolor{tud3b}{RGB}{0,157,129}
\definecolor{tud4b}{RGB}{153,192,0}
\definecolor{tud5b}{RGB}{201,212,0}
\definecolor{tud6b}{RGB}{253,202,0}
\definecolor{tud7b}{RGB}{245,163,0}
\definecolor{tud8b}{RGB}{236,101,0}
\definecolor{tud9b}{RGB}{230,0,26}
\definecolor{tud10b}{RGB}{166,0,132}
\definecolor{tud11b}{RGB}{114,16,133}
\definecolor{tud1c}{RGB}{0,78,138}
\definecolor{tud2c}{RGB}{0,104,157}
\definecolor{tud3c}{RGB}{0,136,119}
\definecolor{tud4c}{RGB}{127,171,22}
\definecolor{tud5c}{RGB}{177,189,0}
\definecolor{tud6c}{RGB}{215,172,0}
\definecolor{tud7c}{RGB}{210,135,0}
\definecolor{tud8c}{RGB}{204,76,3}
\definecolor{tud9c}{RGB}{185,15,34}
\definecolor{tud10c}{RGB}{149,17,105}
\definecolor{tud11c}{RGB}{97,28,115}
\definecolor{tud1d}{RGB}{36,53,114}
\definecolor{tud2d}{RGB}{0,78,115}
\definecolor{tud3d}{RGB}{0,113,94}
\definecolor{tud4d}{RGB}{106,139,55}
\definecolor{tud5d}{RGB}{153,166,4}
\definecolor{tud6d}{RGB}{174,142,0}
\definecolor{tud7d}{RGB}{190,111,0}
\definecolor{tud8d}{RGB}{169,73,19}
\definecolor{tud9d}{RGB}{156,28,38}
\definecolor{tud10d}{RGB}{115,32,84}
\definecolor{tud11d}{RGB}{76,34,106}
\definecolor{unlabeled}{RGB}{0,0,0}
\definecolor{egovehicle}{RGB}{0,0,0}
\definecolor{rectification border}{RGB}{0,0,0}
\definecolor{outofroi}{RGB}{0,0,0}
\definecolor{static}{RGB}{0,0,0}
\definecolor{dynamic}{RGB}{111,74,0}
\definecolor{ground}{RGB}{81,0,81}
\definecolor{road}{RGB}{128,64,128}
\definecolor{sidewalk}{RGB}{244,35,232}
\definecolor{parking}{RGB}{250,170,160}
\definecolor{rail track}{RGB}{230,150,140}
\definecolor{building}{RGB}{70,70,70}
\definecolor{wall}{RGB}{102,102,156}
\definecolor{fence}{RGB}{190,153,153}
\definecolor{guard rail}{RGB}{180,165,180}
\definecolor{bridge}{RGB}{150,100,100}
\definecolor{tunnel}{RGB}{150,120,90}
\definecolor{pole}{RGB}{153,153,153}
\definecolor{polegroup}{RGB}{153,153,153}
\definecolor{trafficlight}{RGB}{250,170,30}
\definecolor{trafficsign}{RGB}{220,220,0}
\definecolor{vegetation}{RGB}{107,142,35}
\definecolor{terrain}{RGB}{152,251,152}
\definecolor{sky}{RGB}{70,130,180}
\definecolor{skylight}{RGB}{98,182,252}
\definecolor{person}{RGB}{220,20,60}
\definecolor{rider}{RGB}{255,0,0}
\definecolor{car}{RGB}{0,0,142}
\definecolor{truck}{RGB}{0,0,70}
\definecolor{bus}{RGB}{0,60,100}
\definecolor{caravan}{RGB}{0,0,90}
\definecolor{trailer}{RGB}{0,0,110}
\definecolor{train}{RGB}{0,80,100}
\definecolor{motorcycle}{RGB}{0,0,230}
\definecolor{bicycle}{RGB}{119,11,32}
\definecolor{licenseplate}{RGB}{0,0,142}
\definecolor{sscbench_car}{HTML}{6496F5}
\definecolor{sscbench_bicycle}{HTML}{FF0000}
\definecolor{sscbench_motorcycle}{HTML}{FF00FF}
\definecolor{sscbench_truck}{HTML}{FF96FF}
\definecolor{sscbench_other_vehicle}{HTML}{4B004B}
\definecolor{sscbench_person}{HTML}{AF004B}
\definecolor{sscbench_road}{HTML}{ffc800}
\definecolor{sscbench_sidewalk}{HTML}{969696}
\definecolor{sscbench_building}{HTML}{1E3C96}
\definecolor{sscbench_fence}{HTML}{501EB4}
\definecolor{sscbench_vegetation}{HTML}{086100}
\definecolor{sscbench_terrain}{HTML}{B83802}
\definecolor{sscbench_pole}{HTML}{FF8F2E}
\definecolor{sscbench_traffic_sign}{HTML}{70FF32}
\definecolor{sscbench_other_object}{HTML}{C20000}
\definecolor{sd1}{RGB}{107, 76, 233}
\definecolor{sd2}{RGB}{213, 154, 130}
\definecolor{sd3}{RGB}{233, 217, 2}

\newcommand{\MethodName}{SceneDINO\@\xspace} %

\newcommand\cbarm[4][tud1b]{\colorbox{tud7b}{\color{black}\framebox(#2,#3){}}\,\colorbox{white}{\color{white}\framebox(#4,#3){}}}
\newcommand\cbarp[4][tud1b]{\colorbox{white}{\color{white}\framebox(#2,#3){}}\,\colorbox{tud3a}{\color{black}\framebox(#4,#3){}}}

\usepackage[pagebackref,breaklinks,colorlinks,allcolors=iccvblue]{hyperref}

\def\paperID{---} %
\def\confName{ICCV}
\def\confYear{2025}

\title{Feed-Forward \gradientRGB{Scene}{107, 76, 233}{213, 154, 130}\gradientRGB{DINO}{213, 154, 130}{233, 217, 2} for Unsupervised Semantic Scene Completion}

\newcommand{\authorstep}{\hspace{0.75cm}}
\newcommand{\affiliationstep}{\hspace{0.5cm}}

\author{
	Aleksandar Jevti\'{c}\textsuperscript{\normalfont{}* 1}
	\authorstep Christoph Reich\textsuperscript{\normalfont{}* 1,2,4,5}
	\authorstep Felix Wimbauer\textsuperscript{\normalfont{} 1,4}\\
	Oliver Hahn\textsuperscript{\normalfont{} 2}
	\authorstep Christian Rupprecht\textsuperscript{\normalfont{} 3}
	\authorstep Stefan Roth\textsuperscript{\normalfont{} 2,5,6}
	\authorstep Daniel Cremers\textsuperscript{\normalfont{} 1,4,5}\\[1pt]
	\small{\textsuperscript{1}TU Munich \affiliationstep \textsuperscript{2}TU Darmstadt \affiliationstep \textsuperscript{3}University of Oxford \affiliationstep \textsuperscript{4}MCML\affiliationstep \textsuperscript{5}ELIZA\affiliationstep \textsuperscript{6}hessian.AI\affiliationstep
		\textsuperscript{*}equal contribution}\\[-1pt]\small {\url{https://visinf.github.io/scenedino}}}

\usepackage{fancyhdr}
\usepackage{setspace}

\fancypagestyle{firststyle}
{

	\fancyhf{}
	\lfoot{{\footnotesize\begin{spacing}{.5}\parbox{\linewidth}{\vspace{-0.85em}%
					To appear in \emph{Proceedings of the IEEE/CVF International Conference on Computer Vision}, Honolulu, Hawai'i, USA, 2025.%
					\\\hrule\vspace{\baselineskip}
					\copyright~2025 IEEE. Personal use of this material is permitted. Permission from IEEE must be obtained for all other uses, in any current or future media, including reprinting/republishing this material for advertising or promotional purposes, creating new collective works, for resale or redistribution to servers or lists, or reuse of any copyrighted component of this work in other works. 
	}\end{spacing}}}
}

\begin{document}
	\twocolumn[{%
		\renewcommand\twocolumn[1][]{#1}%
		\maketitle
		\vspace{-2.1em}
		{\centering\pgfdeclarehorizontalshading{scenedinoshading}{100bp}{
	color(0bp)=(sd1);
	color(50bp)=(sd2);
	color(100bp)=(sd3)
}
\pgfdeclarehorizontalshading{scenedinoshading2}{100bp}{
	color(0bp)=(vegetation!75);
	color(42bp)=(vegetation!75);
	color(43bp)=(road!75);
	color(64bp)=(road!75);
	color(65bp)=(car!75);
	color(100bp)=(car!75)
}
\begin{tikzpicture}[
	clip,
	baseline,
	>={Stealth[inset=0pt,length=8.0pt,angle'=35]},
	every node/.style={font=\sffamily\small},
	]
	
	\node at (0.375, 0) {\includegraphics[width=4.25cm]{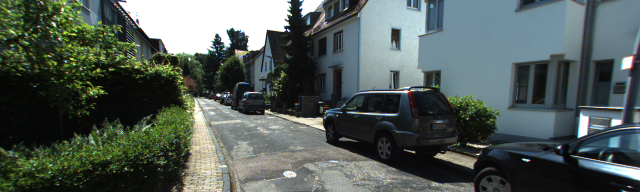}};
	\node[] at (0.375, -1.35) {\normalsize\textbf{Single Input Image}};
	
	\node[
	single arrow,
	shape border rotate=0,
	draw=none, 
	shading=axis, 
	shading=scenedinoshading,
	shading angle=0, 
	text=white,
	anchor=base, 
	align=center,
	single arrow head indent=0.075cm,
	text width=1.8cm] at (3.6, -0.1) {\MethodName};
	
	\node at (7.0, 0.0) {\includegraphics[width=4cm]{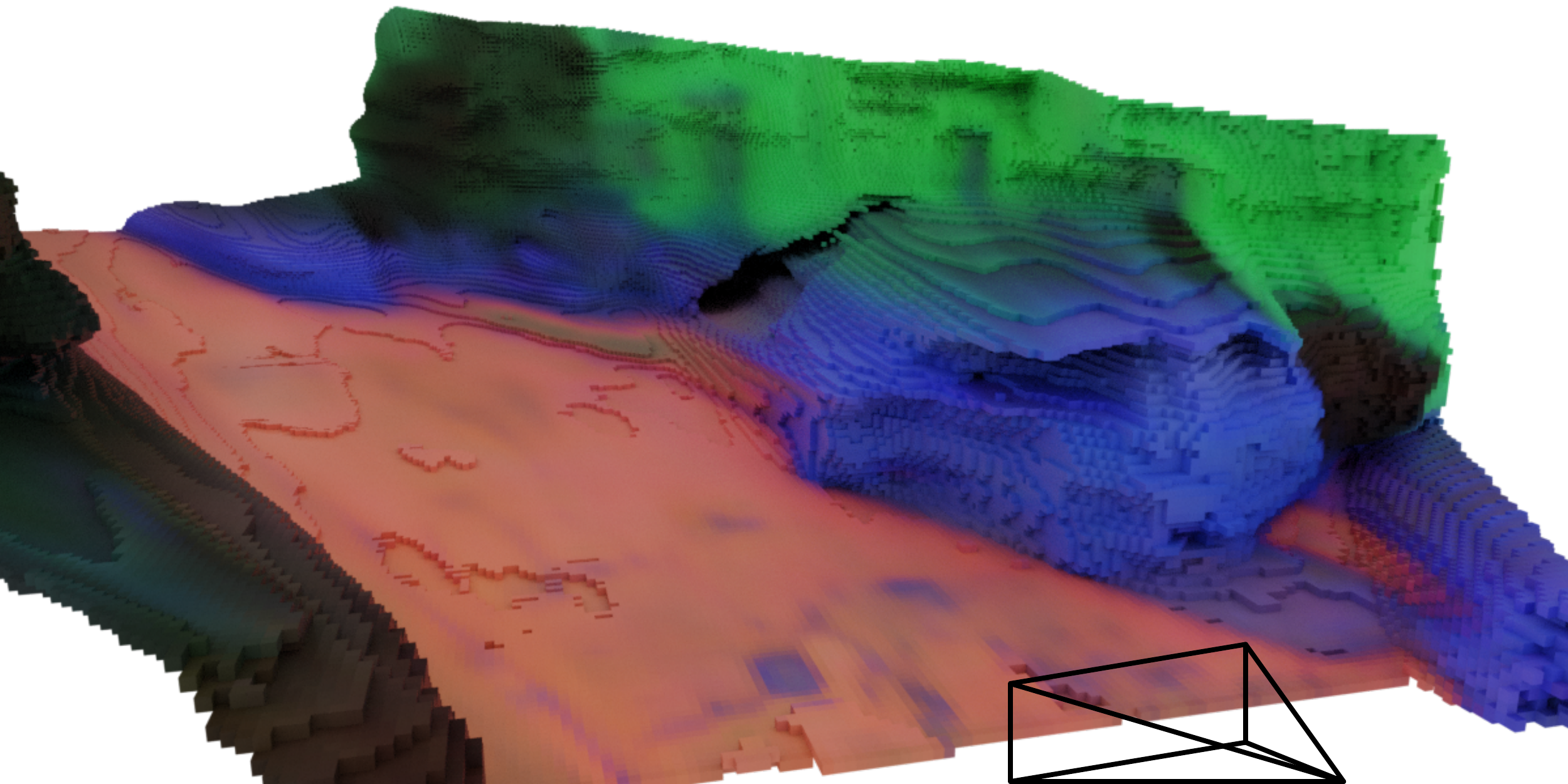}};
	\node[] at (7.0, -1.35) {\normalsize\textbf{3D Feature Field}\vphantom{g}};
	
	\node[
	single arrow,
	shape border rotate=0,
	draw=none, 
	shading=axis, 
	shading=scenedinoshading2,
	shading angle=0, 
	text=white,
	anchor=base, 
	align=center,
	single arrow head indent=0.075cm,
	text width=1.8cm] at (10.1, -0.1) {\!\!\!\!\!\!\!\!Distill\,\&\,Cluster\!\!\!\!\!\!\!\!\!\!};
	
	\node at (13.5, 0.0) {\includegraphics[width=4cm]{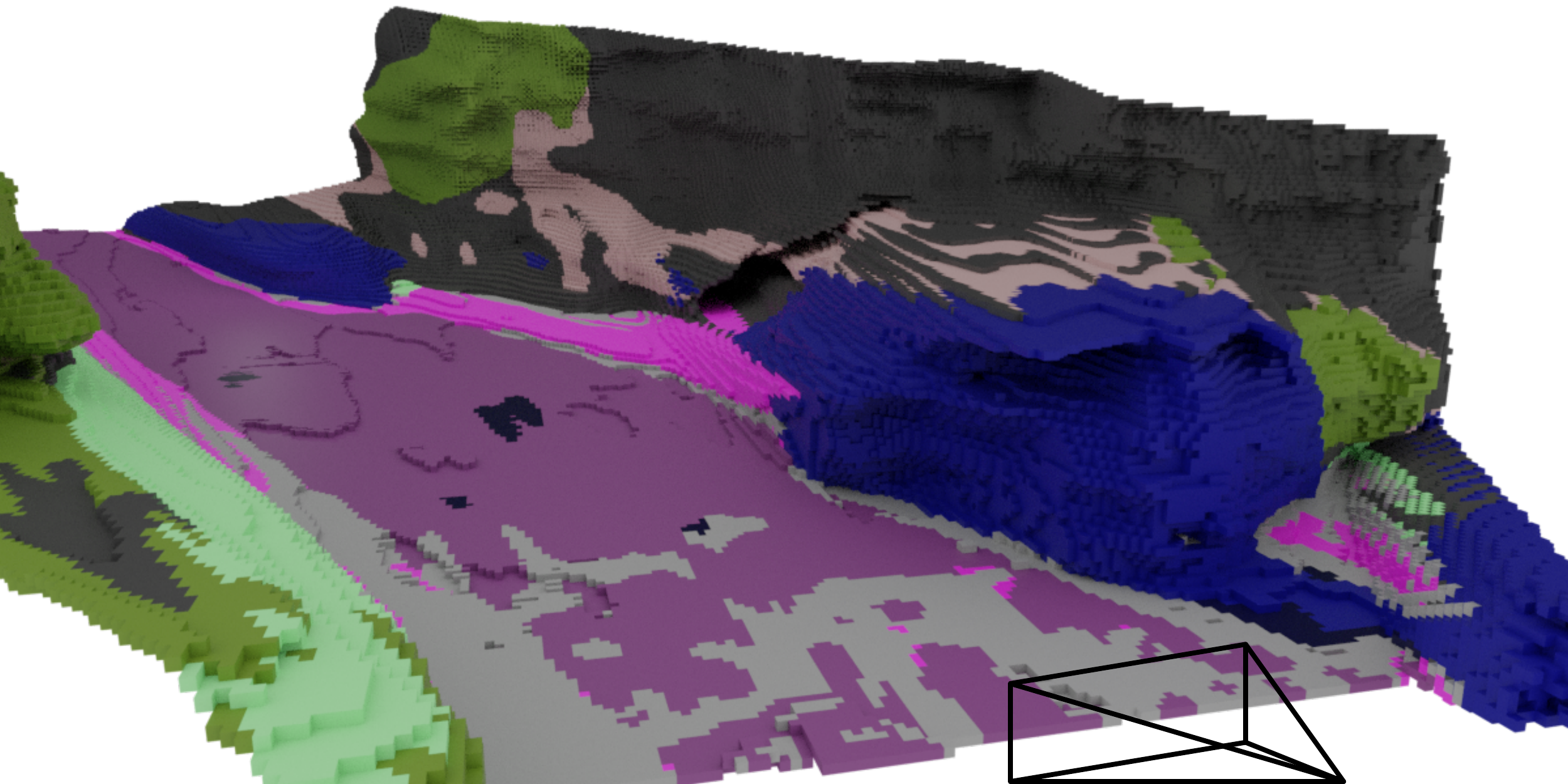}};
	\node[] at (13.5, -1.35) {\normalsize\textbf{SSC Prediction}\vphantom{g}};
	
\end{tikzpicture}
}
		\vspace{-0.75em}
		\captionof{figure}{\textbf{\MethodName overview.} Given a single input image \emph{(left)}, SceneDINO predicts both 3D scene geometry and 3D features in the form of a feature field \emph{(middle)} in a feed-forward manner, capturing the structure and semantics of the scene. Unsupervised distillation and clustering of \MethodName's feature space leads to unsupervised semantic scene completion predictions \emph{(right)}.\label{fig:teaser}}
		\vspace{1.5\baselineskip}
	}]
\begin{abstract}
Semantic scene completion (SSC) aims to infer both the 3D geometry and semantics of a scene from single images. In contrast to prior work on SSC that heavily relies on expensive ground-truth annotations, we approach SSC in an \emph{unsupervised} setting. Our novel method, \MethodName, adapts techniques from self-supervised representation learning and 2D unsupervised scene understanding to SSC. Our training exclusively utilizes multi-view consistency self-supervision without any form of semantic or geometric ground truth. Given a single input image, \MethodName infers the \emph{3D geometry} and expressive \emph{3D DINO features} in a feed-forward manner. Through a novel 3D feature distillation approach, we obtain unsupervised 3D semantics. In both 3D and 2D unsupervised scene understanding, \MethodName reaches state-of-the-art segmentation accuracy. Linear probing our 3D features matches the segmentation accuracy of a current \emph{supervised} SSC approach. Additionally, we showcase the domain generalization and multi-view consistency of \MethodName, taking the first steps towards a strong foundation for single image 3D scene understanding.%
\end{abstract} %

\thispagestyle{firststyle} 
\section{Introduction\label{sec:introduction}}

Understanding the geometry and semantics of 3D scenes from image observations is a fundamental computer vision task with broad applications in robotics~\cite{Geiger:2013:KIT}, autonomous driving~\cite{Janai:2020:AVS, Liao:2023:KND}, medical image analysis~\cite{Cicek:2016:U3D, Yang:2024:M3D}, and civil engineering~\cite{Ma:2018:R3D}. The Semantic Scene Completion (SSC) task unifies 3D geometry and semantic prediction from limited image observations~\cite{Song:2017:SSC, Li:2024:SSC, Roldao:2022:SSC}. Recent progress in SSC has been primarily driven by utilizing supervised learning~\cite{Song:2017:SSC, Hayler:2024:S4C, Roldao:2020:LM3}. However, acquiring large-scale 3D annotations is highly labor-intensive~\cite{Liao:2023:KND}. While significant resources have been invested in collecting human annotations for 2D tasks~\cite{Kirillov:2023:SAM, Ravi:2024:SAM}, annotating similar amounts of data in 3D remains unapproached. This motivates approaching SSC without the need for manually annotated data.

Existing SSC approaches rely on ground-truth semantic annotations and frequently utilize additional supervision from LiDAR scans~\cite{Song:2017:SSC, Miao:2023:ODM, Huang:2024:SES, Hayler:2024:S4C}. In contrast, we are the first to approach SSC in a \emph{fully unsupervised} setting, \ie without task supervision or other supervised components. In particular, we aim to approach SSC from a \emph{single image} without relying on any human annotations, only learning from unlabeled multi-view images using self-supervision. This setting is extremely challenging for two reasons: \emph{first}, the human-defined nature of semantic taxonomies is ambiguous, and \emph{second}, a single image only provides a partial observation of the scene with many invisible areas. We take inspiration from recent advances in self-supervised learning (SSL) of 2D representations and 3D reconstruction. 2D SSL representations, such as from DINO~\cite{Caron:2021:EPS}, have been shown effective for 2D unsupervised scene understanding~\cite{Hamilton:2022:USS, Wang:2023:CAL, Hahn:2025:UPS}. 3D reconstruction approaches successfully leveraged SSL from multi-view data to infer dense 3D geometry from a single image~\cite{Wimbauer:2023:BTS, Han:2024:BSV}. %

In this paper, we present \emph{\MethodName}, to the best of our knowledge, the first approach for unsupervised semantic scene completion. Trained using 2D SSL features from DINO~\cite{Caron:2021:EPS} and multi-view self-supervision~\cite{Wimbauer:2023:BTS}, \MethodName predicts both 3D geometry and 3D features from a single image during inference in a feed-forward manner. Our general 3D feature representations enable us to approach unsupervised 3D scene understanding. Harnessing our expressive 3D features, we propose a novel 3D feature distillation approach for obtaining unsupervised semantic predictions in 3D. %
While we focus on the task of unsupervised SSC, \MethodName's features are general, offering a foundation for different 3D scene-understanding tasks by building on our 3D feature field.

Specifically, we make the following contributions: 
\emph{(i)}~We introduce \MethodName, the first approach predicting dense 3D geometry \emph{and} expressive 3D features in a \emph{feed-forward manner} from a \emph{single image}. 
\emph{(ii)}~We effectively distill \MethodName's feature field representation in 3D, obtaining unsupervised semantic predictions. 
\emph{(iii)}~We demonstrate the first fully unsupervised SSC results. We build a simple, yet competitive unsupervised SSC baseline, lifting unsupervised 2D semantic predictions. Our \MethodName approach outperforms this SSC baseline in unsupervised SSC as well as established 2D approaches in 2D semantic segmentation. 
\emph{(iv)}~Finally, we also showcase the domain generalization ability and multi-view consistency of \MethodName.

\section{Related Work\label{sec:related_work}}

\inparagraph{Single-image scene reconstruction.} Estimating 3D geometry from image observations is a fundamental task in computer vision and has been studied for decades~\cite{Hartley:2003:MVG}. Traditional approaches, such as structure from motion~\cite{Schonberger:2016:SFM}, as well as recent neural radiance fields (NeRFs)~\cite{Mildenhall:2021:NER}, perform scene reconstruction using multiple images, as reviewed by multiple surveys~\cite{Ozyesil:2017:SFM, Han:2019:S3D, Xie:2022:NEF}. Recently, estimating dense 3D geometry from a single image have been approached~\cite{Oswald:2013:SSR, Tulsiani:2017:CVPR, Richter:2018:MNP, Wimbauer:2023:BTS, Yu:2021:PNE, Han:2024:BSV, Cao:2022:M3S, Szymanowicz:2024:F3D}. Unlike monocular depth estimation~\cite{Ming:2021:MDE}, these approaches predict the depth for visible and occluded regions, reconstructing a complete scene. Behind the Scenes (BTS)~\cite{Wimbauer:2023:BTS} introduced an approach for unsupervised single-image \emph{scene} reconstruction using multi-view self-supervision, which infers dense 3D geometry in a feed-forward manner. Our approach extends BTS by additionally lifting self-supervised features into 3D for unsupervised 3D scene understanding.

\inparagraph{Semantic scene completion (SSC),} also known as 3D semantic occupancy prediction, aims to jointly estimate the 3D geometry and semantics of a scene~\cite{Song:2017:SSC, Li:2024:SSC, Zhang:2023:OCC, Li:2023:VSV}. Initial approaches used 3D semantic and geometric annotations and addressed indoor scenes \cite{Cai:2021:SSC, Chen:2020:3SA, Li:2019:RBD, Li:2020:DBS, Li:2020:ACN, Liu:2028:STD, Zhang:2019:CCP}, outdoor scenes with LiDAR \cite{Cheng:2020:SSS, Li:2021:SIS, Rist:2022:SSC, Roldao:2020:LM3, Yan:2021:SSS}, or both domains~\cite{Cao:2022:M3S, Miao:2023:ODM}. Using birds-eye views has been proven effective for SSC~\cite{Li:2023:F3O, Tong:2023:SAO, Huang:2023:TPV}. To overcome the need for 3D annotations, approaches for using 2D annotations have been proposed~\cite{Huang:2024:SES, Hayler:2024:S4C, Pan:2024:RenderOcc}. While SelfOcc~\cite{Huang:2024:SES} and RenderOcc~\cite{Pan:2024:RenderOcc} use multiple inference views, S4C~\cite{Hayler:2024:S4C} performs single-image SSC. In particular, S4C~\cite{Hayler:2024:S4C} employs a supervised 2D model and lifts 2D multi-view semantic predictions into 3D. In contrast to using 2D annotations, GaussTR~\cite{Jiang:2024:GTR} uses 2D foundation models for SSC and multiple views during inference. However, GaussTR relies on heavily supervised foundation models, including SAM~\cite{Kirillov:2023:SAM} and Metric3Dv2~\cite{Hu:2024:M3D}, and uses weak supervision from image/text pairs. To the best of our knowledge, there is no method for approaching SSC without the need for any ground-truth annotations. Our work presents the first unsupervised SSC approach, utilizing lifted SSL features and a single RGB input image for inference.

\inparagraphnohspace{Self-supervised representation learning (SSL)} aims to extract general features from data without annotations, facilitating various downstream tasks such as segmentation~\cite{Ericsson:2022:SSL}. Recent SSL methods, often based on Vision Transformers (ViTs)~\cite{Dosovitskiy:2021:AIW}, leverage clustering~\cite{Asano:2020:SLC, Caron:2018:DCL, Caron:2020:ULV, Ji:2019:IIC, Li:2021:PCL}, masked modeling~\cite{He:2022:MAE, Gupta:2023:SMA, Nguyen:2024:RMA, Wei:2022:MFP, Darcet:2025:CPL}, contrastive learning~\cite{Bachman:2019:LRM, Hjelm:2019:LDR, Henaff:2020:CPP, Chen:2020:ISL, He:2020:MCU, Chen:2021:SSL}, or negative-free~\cite{Bardes:2022:VIL, Caron:2021:EPS, Grill:2020:BYL, Bardes:2022:VIC, Oquab:2023:DLR} pretext tasks~\cite{Doersch:2015:UVR, Noroozi:2016:JIG} for large-scale training. State-of-the-art models, \eg, DINO~\cite{Caron:2021:EPS}, produce semantically rich, dense features, driving recent advances in 2D unsupervised scene understanding~\cite{Hamilton:2022:USS, Wang:2023:CAL}. We here aim to bring expressive features from DINO~\cite{Caron:2021:EPS, Oquab:2023:DLR} to 3D for SSC.

\begin{figure*}[t]
    \centering
     \begin{subfigure}[b]{0.32\textwidth}
        \centering
        \vspace{-6pt}
        \def\svgwidth{0.9\columnwidth}
        \input{artwork/method/architechture_overview_paper}
        \subcaption{3D feature field \& semantic inference\label{fig:sd_inference}}
     \end{subfigure}
     \hfill
     \unskip\ \vrule\
     \begin{subfigure}[b]{0.32\textwidth}
        \centering
        \input{artwork/method/rendering}
        \subcaption{Volumetric feature \& image rendering\label{fig:sd_render}}
     \end{subfigure}
     \hfill
     \unskip\ \vrule\
     \begin{subfigure}[b]{0.32\textwidth}
        \centering
        \vspace{-6pt}
        \def\svgwidth{0.75\columnwidth}%
\scriptsize%
\sffamily%
\setlength{\tabcolsep}{1.75pt}%
\begingroup%
  \makeatletter%
  \providecommand\color[2][]{%
    \errmessage{(Inkscape) Color is used for the text in Inkscape, but the package 'color.sty' is not loaded}%
    \renewcommand\color[2][]{}%
  }%
  \providecommand\transparent[1]{%
    \errmessage{(Inkscape) Transparency is used (non-zero) for the text in Inkscape, but the package 'transparent.sty' is not loaded}%
    \renewcommand\transparent[1]{}%
  }%
  \providecommand\rotatebox[2]{#2}%
  \newcommand*\fsize{\dimexpr\f@size pt\relax}%
  \newcommand*\lineheight[1]{\fontsize{\fsize}{#1\fsize}\selectfont}%
  \ifx\svgwidth\undefined%
    \setlength{\unitlength}{1200bp}%
    \ifx\svgscale\undefined%
      \relax%
    \else%
      \setlength{\unitlength}{\unitlength * \real{\svgscale}}%
    \fi%
  \else%
    \setlength{\unitlength}{\svgwidth}%
  \fi%
  \global\let\svgwidth\undefined%
  \global\let\svgscale\undefined%
  \makeatother%
  \begin{picture}(1,0.5)%
    \lineheight{1}%
    \setlength\tabcolsep{0pt}%
    \put(0,0){\includegraphics[width=\unitlength,page=1]{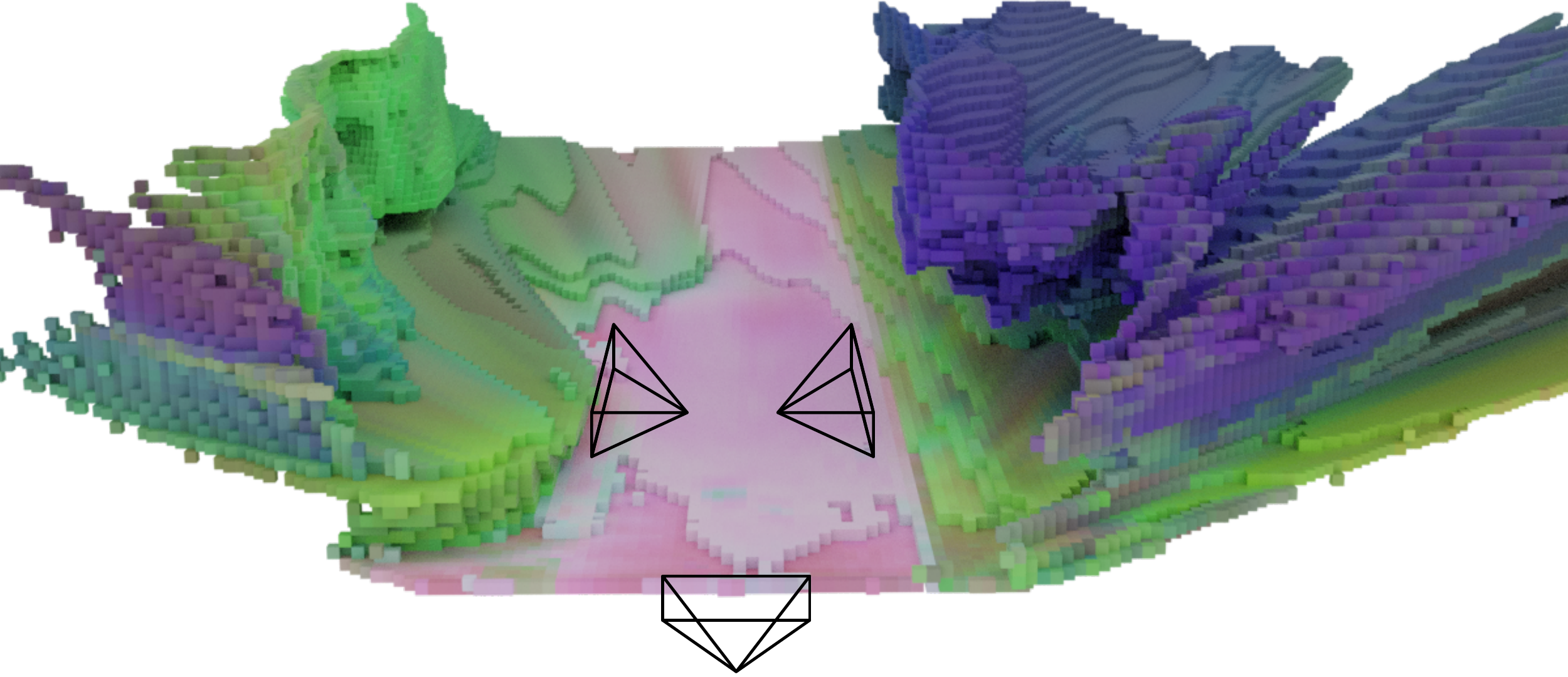}}%
    \put(0.40645264,0.16022705){\color[rgb]{0,0,0}\makebox(0,0)[lt]{\lineheight{1.25}\smash{\begin{tabular}[t]{l}\end{tabular}}}}%
    \put(0.46457764,0.10897705){\color[rgb]{0,0,0}\makebox(0,0)[lt]{\lineheight{1.25}\smash{\begin{tabular}[t]{l}\end{tabular}}}}%
    \put(0.51895264,0.16022705){\color[rgb]{0,0,0}\makebox(0,0)[lt]{\lineheight{1.25}\smash{\begin{tabular}[t]{l}\end{tabular}}}}%
  \end{picture}%
\endgroup%
\\%
\hspace{-0pt}\begin{tabular}{ccc}
    \multicolumn{3}{c}{Reconstructed views} \\[-1.0pt]
    \includegraphics[width=0.275\columnwidth]{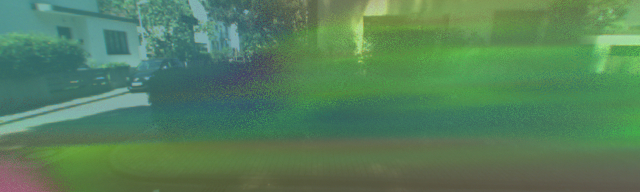} & \includegraphics[width=0.275\columnwidth]{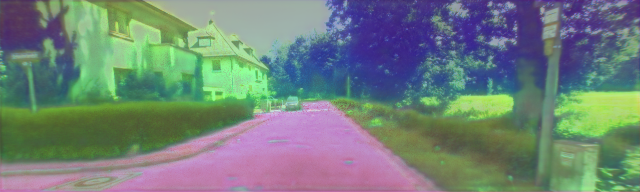} & \includegraphics[width=0.275\columnwidth]{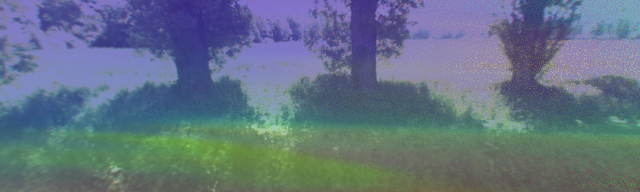} \\[-2.9pt]
    \begin{tikzpicture}[>={Stealth[inset=0pt,length=3.0pt,angle'=45]}] \draw[thick, ->] (0, 0) -- (0, -0.2); \end{tikzpicture} & \begin{tikzpicture}[>={Stealth[inset=0pt,length=3.0pt,angle'=45]}] \draw[thick, ->] (0, 0) -- (0, -0.2); \end{tikzpicture} & \begin{tikzpicture}[>={Stealth[inset=0pt,length=3.0pt,angle'=45]}] \draw[thick, ->] (0, 0) -- (0, -0.2); \end{tikzpicture} \\[-3.5pt]
    
    \multicolumn{3}{c}{\begin{tikzpicture}[>={Stealth[inset=0pt,length=3.0pt,angle'=45]}] \draw[fill=white] (0.0, 0.0) rectangle ++(4.6, 0.31) node[pos=.5] {Multi-view image \& feature reconstruction}; \end{tikzpicture}} \\[-3.5pt]
    
    \begin{tikzpicture}[>={Stealth[inset=0pt,length=3.0pt,angle'=45]}] \draw[thick, <-] (0, 0) -- (0, -0.2); \end{tikzpicture} & \begin{tikzpicture}[>={Stealth[inset=0pt,length=3.0pt,angle'=45]}] \draw[thick, <-] (0, 0) -- (0, -0.2); \end{tikzpicture} & \begin{tikzpicture}[>={Stealth[inset=0pt,length=3.0pt,angle'=45]}] \draw[thick, <-] (0, 0) -- (0, -0.2); \end{tikzpicture} \\[-2.9pt]
    \includegraphics[width=0.275\columnwidth]{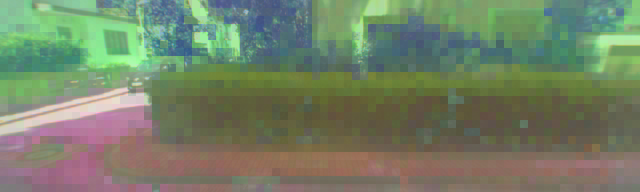} & \includegraphics[width=0.275\columnwidth]{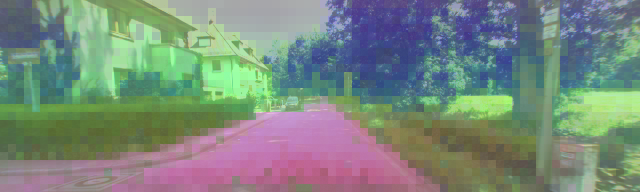} & \includegraphics[width=0.275\columnwidth]{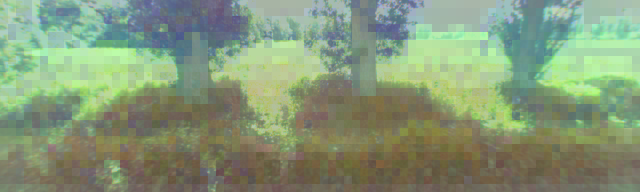} \\[-1.5pt]
    \multicolumn{3}{c}{Target views} \\
\end{tabular}

        \subcaption{Self-supervised multi-view training\label{fig:sd_train}}
     \end{subfigure}
    \centering
    \vspace{-0.5em}
    \caption{\textbf{\MethodName architecture, rendering, and training.} \emph{\subref{fig:sd_inference}} Inference: Given a single input image $\mathbf{I}_{0}$ during inference, a 2D encoder-decoder $\xi$ produces the embedding $\mathbf{E}$ from which the local embedding $\mathbf{e}_{\mathbf{u}}$ is interpolated. The MLP encoder $\phi$ takes in $\mathbf{e}_{\mathbf{u}}$ and 3D position $\mathbf{x}_{i}$, and predicts both the density $\sigma_{\mathbf{x}_{i}}$ and the 3D feature $f_{\mathbf{x}_{i}}$. Using a lightweight unsupervised segmentation head $h$, we can obtain semantic predictions $p_{\mathbf{x}_{i}}$ using $f_{\mathbf{x}_{i}}$. \emph{\subref{fig:sd_render}} Rendering: Our feature field allows for volume rendering by shooting rays through it, yielding depth $\hat{d}$ and $\hat{f}$ in 2D. Color $c_{i}$ is sampled from an another view (\eg, $\mathbf{I}_{1}$) using $\mathbf{u}_{s}$ and rendered to obtain the reconstructed color~$\hat{c}$. \emph{\subref{fig:sd_train}}~Multi-view training: We render 2D views (features \& images) from our feature field and reconstruct the training views.}
    \label{fig:architecture-overview}
    \vspace{-0.5em}
\end{figure*}

\inparagraph{2D-to-3D feature lifting.} The expressiveness of 2D visual representations has motivated lifting 2D features into 3D~\cite{Shen:2023:DFF, Yang:2024:EME}. Existing approaches utilize multi-view 2D features for 3D feature lifting~\cite{Tschernezki:2022:NFF, Kerr:2023:LER, Shafiullah:2023:CFI, Tsagkas:2023:VLF, Kobayashi:2022:DNE, Ha2023:SEA, Huang:2024:S3D, Mazur:2023:FRN, Peng:2023:OPS, Shen:2023:DFF, Takmaz:2023:OM3, Weder:2024:LAM, Yang:2024:EME, Yue:2024:FIT}. Lifting 2D features is effective in various tasks, including few-shot semantic occupancy prediction~\cite{Yang:2024:EME}, and refining 2D representations~\cite{Yue:2024:FIT}. However, existing feature-lifting approaches fit to a single scene~\cite{Tschernezki:2022:NFF, Yang:2024:EME, Yue:2024:FIT, Kerr:2023:LER, Shafiullah:2023:CFI, Tsagkas:2023:VLF, Kobayashi:2022:DNE, Shen:2023:DFF}, require RGB-D inputs~\cite{Mazur:2023:FRN, Ha2023:SEA, Huang:2024:S3D, Takmaz:2023:OM3, Weder:2024:LAM}, or work on 3D point cloud inputs~\cite{Peng:2023:OPS}. The only feed-forward approaches that use RGB inputs and lift 2D features, which we are aware of, are GaussTR~\cite{Jiang:2024:GTR} and MVSplat360~\cite{Chen:2024:MVS}. However, both approaches utilize multiple input images during inference, and MVSplat360~\cite{Chen:2024:MVS} only predicts low-dimensional feature representations, which are not suitable for unsupervised scene understanding. In contrast, we propose the first feed-forward approach for inferring lifted high-dimensional and rich 3D features using a single input image.

\inparagraphnohspace{2D unsupervised semantic segmentation} partitions images automatically into semantically meaningful regions without any form of human annotations. Early deep learning-based methods \cite{Ji:2019:IIC, Cho:2021:PUS, Harb:2021:IFS} approach the problem via representation learning. Leveraging SSL features from DINO as an inductive prior, STEGO~\cite{Hamilton:2022:USS} distills the feature representation into a lower-dimensional space for unsupervised probing. Building up on STEGO, subsequent methods \cite{Seong:2023:LHP, Kim:2024:EAL, Sick:2024:USS, Hahn:2024:BUS} propose enhancements to the distillation. Our approach follows the idea of STEGO~\cite{Hamilton:2022:USS}, extending it to 3D and integrating feature distillation using our 3D feature field to build the first unsupervised SSC approach.

\section{Unsupervised Semantic Scene Completion\label{sec:method}}

We approach semantic scene completion (SSC) without any form of manual supervision. To this end, we first describe \MethodName, predicting \emph{3D geometry} and expressive \emph{3D features} from a \emph{single image} in a \emph{feed-forward manner} (\cref{subsec:feature-lifting}), and \MethodName's multi-view training (\cref{subsec:training}). Next, we present our 3D feature distillation approach to obtain \emph{unsupervised 3D semantic} predictions (\cref{subsec:ssc-head}). An overview of our full pipeline, including inference, rendering, and multi-view self-supervision, is provided in \cref{fig:architecture-overview}.

\inparagraph{Notation.} Let $\mathbf{I}_0 \in {[0,1]}^{3 \times \rm H \times \rm W}$ be a single RGB input image (for both training \& inference) with corresponding pose $T_0 \in \mathbb{R}^{4 \times 4}$ and projection matrix $K_0 \in \mathbb{R}^{3 \times 4}$. 
For training, let $(\mathbf{I}_{v}, T_{v}, K_{v})$ with $v \in \{1, 2, \dots, n\}$, be $n$ additional views for multi-view self-supervision. 
Assuming a pinhole camera model, any 3D point $\mathbf{x} \in \mathbb{R}^3$ in world coordinates can be projected onto the image plane of view $v$ and the input view $v=0$ with the perspective projection $\pi_v(\mathbf{x})$.

\subsection{\MethodName}\label{subsec:feature-lifting}

Given a single input image $\mathbf{I}_0$, \MethodName represents the dense geometric structure and features of a scene as a continuous mapping from world coordinates $\mathbf{x} \in \mathbb{R}^3$ to a volumetric density $\sigma_{\mathbf{x}} \in \mathbb{R}_{+}$ and a feature $f_{\mathbf{x}} \in \mathbb{R}^{\rm D}$. This continuous output representation is often called a \emph{feature field}.
While \MethodName could represent any feature space, we aim for expressive SSL features from DINO~\cite{Caron:2021:EPS, Oquab:2023:DLR}.

\inparagraph{Architecture \& feature field inference.} Our \MethodName architecture comprises two main parts: a 2D encoder-decoder $\xi$ and an MLP decoder (\cf \cref{fig:sd_inference}), following BTS~\cite{Wimbauer:2023:BTS}. $\xi$ takes in $\mathbf{I}_0$ and produces a per-pixel embedding $\mathbf{E}\in\mathbb{R}^{{\rm D}_{\mathbf{E}} \times {\rm H} \times {\rm W}}$ with ${\rm D}_{\mathbf{E}}$ dimensions. Intuitively, every spatial element of $\mathbf{E}$ represents a camera ray through a pixel, capturing both local geometry and features.

To infer the feature at a 3D position $\mathbf{x}$, we employ a two-layer MLP decoder $\phi$ (\cf \cref{fig:sd_inference}). Given a position $\mathbf{x}$ within the camera frustum, we project $\mathbf{x}$ into the camera plane, obtaining the pixel location $\mathbf{u}=\pi_0(\mathbf{x})$. We query $\mathbf{E}$ at the position $\mathbf{u}$ using bilinear interpolation, obtaining the local embedding $\mathbf{e}_{\mathbf{u}}$. Based on the embedding $\mathbf{e}_{\mathbf{u}}$, the pixel position $\mathbf{u}$, and the distance $d_{\mathbf{x}}\in\mathbb{R}_{+}$ of $\mathbf{x}$ to the camera, we obtain the density $\sigma_{\mathbf{x}}$ and feature prediction $f_{\mathbf{x}}$ as
\begin{equation}\label{eq:denisty_features}
    (\sigma_{\mathbf{x}}, f_{\mathbf{x}})  = \phi(\mathbf{e}_{\mathbf{u}}, \gamma(\mathbf{u}, d_{\mathbf{x}})),
\end{equation}
where $\gamma$ denotes a positional encoding~\cite{Mildenhall:2021:NER}.

\inparagraph{Feature, depth \& color volume rendering.} \MethodName predicts a continuous feature field from a single image. This representation can be used to render features and depth in 2D from an arbitrary viewpoint (\cf \cref{fig:sd_render}), following the discretization strategy of Max \etal~\cite{Max:1995:VOL}. Given a viewpoint $(T, K)$, we sample $\rm L$ points $\mathbf{x}_{i}$ along the ray through pixel $\mathbf{u}_r$, with distance $\delta_{i}$ between $\mathbf{x}_{i}$ and $\mathbf{x}_{i + 1}$. Based on the volumetric densities $\sigma_{\mathbf{x}_i}$ (\cf Eq.\ \ref{eq:denisty_features}), we can compute the probabilities $\alpha_i$ of the ray ending between $\mathbf{x}_i$ and $\mathbf{x}_{i+1}$, and accumulate these into $V_i$, the probability of $\mathbf{x}_{i}$ being visible:
\begin{equation}\label{eq:render1}
     V_{i}=\prod_{j=1}^{i-1} \left(1-\alpha_{j}\right),\quad\text{with}\;\alpha_{i}=1-\exp\!\left(-\sigma_{\mathbf{x}_i}\delta_{i}\right).
\end{equation}
Using $V_{i}$ and $\alpha_{i}$, we render depth $\tilde{d}_{\mathbf{u}_r}$ and feature $\tilde{f}_{\mathbf{u}_r}$ from the estimated features $f_{\mathbf{x}_i}$ from \cref{eq:denisty_features} and distances $d_{\mathbf{x}_i}$ to $\mathbf{x}_i$ onto the image plane at position $\mathbf{u}_r$ as
\begin{equation}\label{eq:render2}
    \tilde{f}_{\mathbf{u}_r} = \sum_{i=1}^{\rm L} V_i \alpha_{i} f_{\mathbf{x}_i} \qquad \tilde{d}_{\mathbf{u}_r} = \sum_{i=1}^{\rm L} V_i \alpha_i d_{\mathbf{x}_i}.
\end{equation}
The differentiability of this rendering process enables us to self-supervise \MethodName using multi-view images and their 2D feature representations (\eg, from DINO~\cite{Caron:2021:EPS}). \MethodName predicts 3D geometry and features, but does not predict color as we focus on semantic downstream tasks. To obtain color for image reconstruction during training, we follow the color sampling approach of BTS~\cite{Wimbauer:2023:BTS}.

\subsection{3D feature field training\label{subsec:training}}

We train \MethodName using \emph{multi-view self-supervision} (\cf \cref{fig:sd_train}), aiming to obtain an expressive and view-consistent feature field without the need for any form of manual annotations. For self-supervision, we sample $n + 1$ views $\mathbf{I}_v$ with camera parameters\footnote{Note, camera poses can be obtained using unsupervised visual SLAM~\cite{Campos:2021:ORB}, strictly adhering to the fully unsupervised setting.} $T_v, K_v$ from the data and obtain dense 2D target features $\mathbf{F}_v$ from a self-supervised ViT (\eg, DINO~\cite{Caron:2021:EPS}).
Note that the 2D features entail a resolution of $\mathbf{F}_v\in\mathbb{R}^{{\rm D} \times \frac{\rm H}{p} \times \frac{\rm W}{p}}$, due to the ViT patch size $p$. 
The set of training views and features $\mathbb{V}=\{(\mathbf{I}_v, T_v, K_v, \mathbf{F}_v) \mid v=0,\ldots,n\}$ is randomly partitioned into two subsets $\mathbb{V}_{\text{source}}$ and $\mathbb{V}_{\text{target}}$.
Training reconstructs the views $\mathbb{V}_{\text{target}}$ using the views of $\mathbb{V}_{\text{source}}$.
In practice, we use a randomly sampled set of image patches that align with the ViT patches instead of the full image. In the following, we still refer to images and the full image resolution for the sake of brevity. %

\inparagraph{Image reconstruction.} We aim to learn the geometry of our feature field via multi-view photometric consistency. In particular, for every image $\mathbf{I}_{t} \in \mathbb{V}_{\text{target}}$ we derive a reconstructed image $\hat{\mathbf{I}}_{t, s}$ from every view $s$ in $\mathbb{V}_{\text{source}}$ using differentiable rendering (\cf Eq.\ \ref{eq:render2}) and color sampling~\cite{Wimbauer:2023:BTS}. Equipped with both the reconstructed image $\hat{\mathbf{I}}_{t, s}$ and the target image $\mathbf{I}_t$, we compute our photometric loss per view as
\begin{equation}\label{eq:p_loss}
    \mathcal{L}_{\text{p}} =\!\min_{\mathbf{I}_s \in \mathbb{V}_\text{source}}\!\!\!\left( \lambda_{1}\mathcal{L}_{1}(\mathbf{I}_t, \hat{\mathbf{I}}_{t,s}) + \lambda_{\text{SSIM}}\mathcal{L}_{\text{SSIM}}(\mathbf{I}_t, \hat{\mathbf{I}}_{t,s}) \right).
\end{equation}
We only consider the minimum loss across the views in $\mathbb{V}_\text{source}$, in practice across patches. The scalars $\lambda_{1}$ and $\lambda_{\text{SSIM}}$ weight the absolute error $\mathcal{L}_{1}$ and the SSIM loss $\mathcal{L}_{\text{SSIM}}$ \cite{Wang:2004:SSI}.

To regularize our 3D geometry, we impose smoothness using an edge-aware smoothness loss~\cite{Godard:2017:UDE}. We estimate 2D depth maps $\tilde{\mathbf{d}}_{t}$ for views in $\mathbb{V}_{\text{target}}$, using Eq.\ \ref{eq:render2}. From $\tilde{\mathbf{d}}_{t}$, we obtain the inverse mean-normalized depths $\tilde{\mathbf{d}}^{*}_{t}$ and compute the edge-aware smoothness loss $\mathcal{L}_{\text{s}}$ per view as
\begin{equation}\label{eq:s_loss}
    \vspace{-0.25pt}
    \mathcal{L}_{\text{s}} = |\nabla_{x}\,\tilde{\mathbf{d}}^{*}_{t}| \, e^{-|\nabla_{x}\mathbf{I}_{t}|} + |\nabla_{y}\, \tilde{\mathbf{d}}^{*}_{t}| \, e^{-|\nabla_{y} \mathbf{I}_{t}|},
    \vspace{-0.25pt}
\end{equation}
where $\nabla_{x}$ and $\nabla_{y}$ denote the first spatial derivatives.

\inparagraph{Feature reconstruction.} We learn a multi-view consistent and expressive 3D feature field using the 2D features $\mathbf{F}_{t}$ from $\mathbb{V}_{\text{target}}$. As we aim to learn a high-resolution (continuous) feature field, we render 2D features using Eq.\ \ref{eq:render2} at the full image resolution $\hat{\mathbf{F}}_{t}\in\mathbb{R}^{\rm D \times \rm H \times \rm W}$. To compensate for the reduced spatial dimension of $\mathbf{F}_{t}$, we employ the downsampler $\psi$ proposed by Fu \etal~\cite{Fu:2024:FUP} to our rendered features $\hat{\mathbf{F}}_{t}$. While current 2D SSL features capture semantics, they lack multi-view consistency, \emph{i.a.}, due to positional encodings used in ViTs~\cite{Yang:2024:DEN}, leading to different features for identical visual content at two distinct positions in an image. As we aim for multi-view consistency, we compensate for this by learning a constant decomposition $\overline{\mathbf{F}}\in\mathbb{R}^{{\rm D} \times \frac{\rm H}{p} \times \frac{\rm W}{p}}$ of features induced by positional encodings. Our feature loss is defined per view as
\begin{equation}
    \mathcal{L}_{\text{f}} = 1 - \operatorname{cos-sim}(\mathbf{F}_{t}, \psi(\hat{\mathbf{F}}_{t}) + \overline{\mathbf{F}}),
\end{equation}
where $\operatorname{cos-sim}$ is the cosine similarity between two features.

\begin{figure}[t]
    \centering
    \includegraphics[width=\columnwidth]{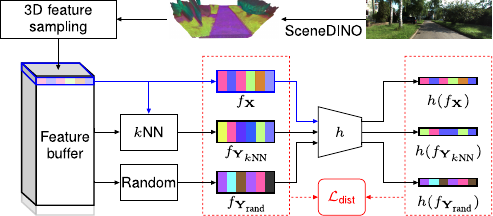}
    \vspace{-1.3em}
    \caption{\textbf{3D feature distillation.} Given an input image, \MethodName predicts a 3D feature field. 3D features $f_{\mathbf{X}}$ are sampled from the feature field. For $f_{\mathbf{X}}$, we obtain $f_{\mathbf{Y}_{k\text{NN}}}$ and $f_{\mathbf{Y}_{k\text{rand}}}$ from the feature buffer. The segmentation head $h$ distills the features into a low-dimensional space and is trained using $\mathcal{L}_{\text{dist}}$.}
    \label{fig:distillation}
    \vspace{-0.3em}
\end{figure}

As image edges correlate with semantic edges and to further impose consistency, we regularize the rendered features $\hat{\mathbf{F}}_{t}$ using an edge-aware smoothness loss per view
\begin{equation}
    \mathcal{L}_{\text{fs}} = |\nabla_x \hat{\mathbf{F}}_t| \, e^{-|\nabla_x \mathbf{I}_t|} + |\nabla_{y} \hat{\mathbf{F}}_t| \, e^{-|\nabla_{y} \mathbf{I}_t|}.
\end{equation}
Our final loss for training \MethodName is a weighted sum of the photometric loss, the feature loss, and both smoothness losses, $\mathcal{L}_{\text{\MethodName}} =\lambda_{\text{p}} \mathcal{L}_{\text{p}} + \lambda_{\text{s}} \mathcal{L}_{\text{s}} + \lambda_{\text{f}} \mathcal{L}_{\text{f}} + \lambda_{\text{fs}} \mathcal{L}_{\text{fs}}$, averaged over all pixels, features, and views.

\subsection{3D feature distillation for unsupervised SSC \label{subsec:ssc-head}}

Given the expressive feature field representation, we aim to obtain unsupervised semantic predictions for SSC. While naïve $k$-means~\cite{Lloyd:1982:PCM, MacQueen:1967:KME} can yield meaningful pseudo semantics, distilling features into a lower-dimensional space has been shown to be more effective in 2D semantic segmentation~\cite{Hamilton:2022:USS, Koenig:2023:IWS}. To this end, we present a novel 3D feature distillation approach (\cf \cref{fig:distillation}). We train a point-wise segmentation head $h$, mapping $f_{\mathbf{x}} \in \mathbb{R}^{\rm D}$ to a lower-dimensional distilled representation $z_{\mathbf{x}} \in \mathbb{R}^{\rm K}$, with $\rm K \ll \rm D$. The resulting distilled space is clustered to obtain pseudo-semantic predictions $p_{\mathbf{x}}\in [0, 1]^{\rm C}$, with $\rm C$ pseudo classes.

Existing work in 2D unsupervised semantic segmentation has shown that SSL feature correspondence captures semantic class co-occurence~\cite{Hamilton:2022:USS}. 
This correspondence between two batches of $\rm N$ sample points $\mathbf{X}=[\mathbf{x}_1, \ldots, \mathbf{x}_{\rm N}]$ and $\mathbf{Y}=[\mathbf{y}_1, \ldots, \mathbf{y}_{\rm N}]$ can be expressed by pairwise feature similarity $S_{i,j} = \operatorname{cos-sim}(f_{\mathbf{x}_{i}}, f_{\mathbf{y}_{j}})\in[-1,1]$. Similarly, we can express the correspondence in the distilled feature space by $S^h_{i,j} = \operatorname{cos-sim}(h(f_{\mathbf{x}_{i}}), h(f_{\mathbf{y}_{j}}))\in[-1,1]$.
We describe the sampling of the $\mathbf{x}_i$ and $\mathbf{y}_j$ below.

\inparagraph{Feature distillation.} We aim to distill features such that similar features align while dissimilar features are separated. To this end, we use the contrastive correlation loss $\mathcal{L}_{\text{corr}}$, introduced by STEGO~\cite{Hamilton:2022:USS} and defined as %
\begin{equation}\label{eq:corr_loss}
    \mathcal{L}_{\text{corr}}(f_{\mathbf{X}}, f_{\mathbf{Y}}, b) = - \sum_{i,j} (S_{i,j} - b) \max(S^h_{i,j}, 0),
\vspace{-0.3em}\end{equation}
where $f_{\mathbf{X}}$, $f_{\mathbf{Y}}$ are the features of the two sample batches.
This loss pushes $S^h_{i,j}$ towards $1$ in case $S_{i,j}$ exceeds the threshold $b$. Otherwise, $\mathcal{L}_{\text{corr}}$ pushes the $S^h_{i,j}$ below $0$.

The correlation loss $\mathcal{L}_{\text{corr}}$ requires informative pairs of sampled features, balancing attractive and repulsive signals. Following STEGO~\cite{Hamilton:2022:USS}, we consider three different relations: \emph{(1)} feature pairs from the same image $(f_{\mathbf{X}}, f_{\mathbf{X}})$, \emph{(2)} feature pairs from an image and its $k$-nearest neighbors in feature space $(f_{\mathbf{X}}, f_{\mathbf{Y}_{k\text{NN}}})$, and \emph{(3)} feature pairs from an image and a randomly sampled other image $(f_{\mathbf{X}}, f_{\mathbf{Y}_{\text{rand}}})$. Note that each pair is obtained from \MethodName's 3D feature field, see below. Equipped with the three feature sample pairs, we compute the full distillation loss as
\begin{equation}\label{eq:distill_loss}
    \begin{aligned}
        \mathcal{L}_{\text{dist}} = &\lambda_{\text{self}} \mathcal{L}_{\text{corr}}(f_{\mathbf{X}}, f_{\mathbf{X}},b_{\text{self}})\\[-1.5pt]
        &+\lambda_{k\text{NN}} \mathcal{L}_{\text{corr}}(f_{\mathbf{X}}, f_{\mathbf{Y}_{k\text{NN}}},b_{k\text{NN}})\\[-1.5pt]
        &+\lambda_{\text{rand}} \mathcal{L}_{\text{corr}}(f_{\mathbf{X}}, f_{\mathbf{Y}_{\text{rand}}},b_{\text{rand}}),
    \end{aligned}
\end{equation}
where $\lambda_{\text{self}}$, $\lambda_{k\text{NN}}$, and $\lambda_{\text{rand}}$ denote the scalar loss weights. $b_{\text{self}}$, $b_{k\text{NN}}$, and $b_{\text{rand}}$ are the contrastive thresholds.

\inparagraph{Feature sampling in 3D.} While obtaining feature pairs using 2D rendered features is straightforward~\cite{Hamilton:2022:USS}, we aim to take advantage of our learned 3D geometry of the scene. To this end, we introduce a novel 3D feature sampling approach for the distillation loss $\mathcal{L}_{\text{dist}}$ from \cref{eq:distill_loss}.
Our goal is to sample features both similar and dissimilar in terms of the encoded semantic concept, which should capture \emph{rich semantics} as well as \emph{different semantic concepts}. 

\begin{figure}[t]
    \centering
    \def\svgwidth{0.625\columnwidth}
    \begingroup%
  \makeatletter%
  \providecommand\color[2][]{%
    \errmessage{(Inkscape) Color is used for the text in Inkscape, but the package 'color.sty' is not loaded}%
    \renewcommand\color[2][]{}%
  }%
  \providecommand\transparent[1]{%
    \errmessage{(Inkscape) Transparency is used (non-zero) for the text in Inkscape, but the package 'transparent.sty' is not loaded}%
    \renewcommand\transparent[1]{}%
  }%
  \providecommand\rotatebox[2]{#2}%
  \newcommand*\fsize{\dimexpr\f@size pt\relax}%
  \newcommand*\lineheight[1]{\fontsize{\fsize}{#1\fsize}\selectfont}%
  \ifx\svgwidth\undefined%
    \setlength{\unitlength}{114.80314961bp}%
    \ifx\svgscale\undefined%
      \relax%
    \else%
      \setlength{\unitlength}{\unitlength * \real{\svgscale}}%
    \fi%
  \else%
    \setlength{\unitlength}{\svgwidth}%
  \fi%
  \global\let\svgwidth\undefined%
  \global\let\svgscale\undefined%
  \makeatother%
  \begin{picture}(1,0.64814815)%
    \lineheight{1}%
    \setlength\tabcolsep{0pt}%
    \put(0,0){\includegraphics[width=\unitlength,page=1]{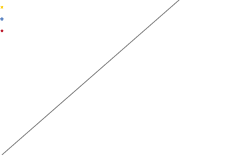}}%
    \put(0.028,0.608){\color[rgb]{0,0,0}\makebox(0,0)[lt]{\lineheight{1.25}\smash{\begin{tabular}[t]{l}\sffamily\scriptsize Center point $\mathbf{X}_{i}$\end{tabular}}}}%
    \put(0.028,0.56){\color[rgb]{0,0,0}\makebox(0,0)[lt]{\lineheight{1.25}\smash{\begin{tabular}[t]{l}\sffamily\scriptsize Accepted samples\end{tabular}}}}%
    \put(0,0){\includegraphics[width=\unitlength,page=2]{artwork/method/sampling_.pdf}}%
    \put(0.028,0.51){\color[rgb]{0,0,0}\makebox(0,0)[lt]{\lineheight{1.25}\smash{\begin{tabular}[t]{l}\sffamily\scriptsize Rejected samples\end{tabular}}}}%
    \put(0.61390699,0.51783226){\color[rgb]{1,1,1}\makebox(0,0)[lt]{\lineheight{1.25}\smash{\begin{tabular}[t]{l}\scriptsize $r$\end{tabular}}}}%
    \put(0,0){\includegraphics[width=\unitlength,page=3]{artwork/method/sampling_.pdf}}%
  \end{picture}%
\endgroup%

    \vspace{-0.4em}
    \caption{\textbf{3D feature sampling.} We first sample a center point $\mathbf{X}_{i}$ from all visible surface points. Further points are sampled within the radius $r$ around the center point $\mathbf{X}_{i}$. Sampled points with sufficient density are accepted; otherwise rejected. The accepted points are used to obtain the feature batch $f_\mathbf{X}$.}
    \label{fig:feature_sampling}
    \vspace{-0.6em}
\end{figure}

First, we obtain all $\rm G$ visible 3D surface points $\mathbf{V}\in\mathbb{R}^{3 \times \rm G}$ and their depth $d_{\mathbf{V}}\in\mathbb{R}^{\rm G}_{+}$ from the camera. To sample points that cover different semantic concepts, we use depth as a cue and sample different depth ranges. In particular, we sort the surface points $\mathbf{V}$ based on $d_{\mathbf{V}}$. The sorted surface points $\hat{\mathbf{V}}$ are partitioned into $\rm M$ equally-sized chunks; we uniformly sample a single 3D point from each chunk, resulting in $\rm M$ center points $\mathbf{X}\in\mathbb{R}^{3 \times \rm M}$.

Equipped with the center points $\mathbf{X}$, we aim to extract rich semantic features from the feature field. While we could just obtain the features for $\mathbf{X}$, we query positions in the neighborhood of $\mathbf{X}$ to increase semantic richness and better capture the 3D structure of the scene for distillation. In particular, for each center point, we randomly sample a point within a radius of $r=0.5\,\text{m}$. To account for samples falling into unoccupied regions in our feature field, we only keep samples with a sufficient density $\sigma > 0.5$. We repeat this sampling process until we obtain $N$ valid samples per center point. Using these samples, we query our feature field, resulting in a feature batch $f_{\mathbf{X}}\in\mathbb{R}^{\rm D \times \rm N}$ for each of the $\rm G$ center points in each scene (\cf \cref{fig:feature_sampling}).

To obtain $f_{\mathbf{Y}_{k\text{NN}}}$ and $f_{\mathbf{Y}_{\text{rand}}}$, we utilize a feature buffer that efficiently stores the sampled features of multiple scenes. Given a new input image, we obtain $G$ feature batches $f_{\mathbf{X}}$ as just described. For each $f_{\mathbf{X}}$, we randomly sample another feature batch from the buffer to obtain $f_{\mathbf{Y}_{\text{rand}}}$. To obtain $f_{\mathbf{Y}_{k\text{NN}}}$, we search in the feature buffer for the $k$-nearest neighbors of $f_{\mathbf{X}}$, using the average feature of each batch. From these $k$-nearest neighbors, we randomly pick a feature batch to obtain $f_{\mathbf{Y}_{k\text{NN}}}$ and compute the distillation loss $\mathcal{L}_{\text{dist}}$. After repeating this process for each of the current $G$ feature batches, we add the current feature batches to the feature buffer and remove the oldest batches.

\inparagraph{Unsupervised probing.} To obtain semantics, we probe the distilled $\rm K$-dim.\ feature space using $k$-means~\cite{MacQueen:1967:KME, Lloyd:1982:PCM}. In particular, we update cluster centers $\theta\in\mathbb{R}^{\rm K \times \rm C}$ using cosine similarity-based mini-batch $k$-means~\cite{Sculley:2010:MBK} during distillation. We compute $p_{\mathbf{x}}=\operatorname{softmax}(\operatorname{cos-sim}(h(f_{\mathbf{x}}),\theta))$ to infer $\rm C$ pseudo semantic class predictions.

\section{Experiments\label{sec:experiments}}

We evaluate \MethodName on SSC and compare it to a simple unsupervised SSC baseline (\cref{subsec:results_3d}). We also report results for 2D unsupervised segmentation, including domain generalization results (\cref{subsec:results_2d}). Finally, we explore multi-view feature consistency (\cref{subsec:results_mvc}) and present an analysis of \MethodName and our 3D distillation (\cref{subsec:analyzing_scenedino}). %

\inparagraph{Datasets.} We train using KITTI-360~\cite{Liao:2023:KND}, composed of clips from a moving vehicle equipped with cameras. For consistency, we follow S4C~\cite{Hayler:2024:S4C} by sampling eight views and using the dataset's camera poses. We further provide results with estimated poses.
We also show experiments for training on RealEstate10k~\cite{Zhou:2018:SML}, composed of monocular videos. Here, we follow the setup of BTS~\cite{Wimbauer:2023:BTS}, obtaining three views. If not noted differently, we report results obtained with training on KITTI-360.
For SSC and 2D semantic segmentation validation, we use the SSCBench-KITTI-360 test split~\cite{Li:2024:SSC}. 
Cityscapes~\cite{Cordts:2016:TCD} and BDD100K~\cite{Yu:2020:BDD} val are used for domain generalization results. To enable evaluation in 3D and 2D, we use the \num{19}-class taxonomy of Cityscapes and perform 2D evaluation on Cityscapes, BDD100K, and KITTI-360 on \num{19} classes. For SSCBench, we combine classes to adhere to the \num{15} SSCBench classes.

\begin{figure*}[t]
    \centering
    \newcommand{\imgwidth}{0.27}
\newcommand{\dddviswidth}{0.182}

\scriptsize
\sffamily
\setlength{\tabcolsep}{0pt}
\renewcommand{\arraystretch}{0.66}
\begin{tabular}{>{\centering\arraybackslash} m{\imgwidth\textwidth} 
                >{\centering\arraybackslash} m{\dddviswidth\textwidth} 
                >{\centering\arraybackslash} m{\dddviswidth\textwidth} 
                >{\centering\arraybackslash} m{\dddviswidth\textwidth}
                >{\centering\arraybackslash} m{\dddviswidth\textwidth}}

\multirow{2}{*}{\vspace{-0.5em}\textbf{Input Image}} & \multicolumn{2}{c}{\textbf{\MethodName \textit{(Ours)}}} & \textbf{S4C + STEGO} & \multirow{2}{*}{\vspace{-0.5em}\textbf{Ground Truth}} \\
\cmidrule(l{0.5em}r{0.5em}){2-3} \cmidrule(l{0.5em}r{0.5em}){4-4}
& Feature Field & SSC Prediction & SSC Prediction & \\[-2pt]

\includegraphics[width=\linewidth]{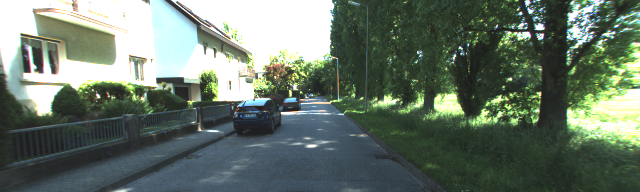} &
\includegraphics[width=\linewidth]{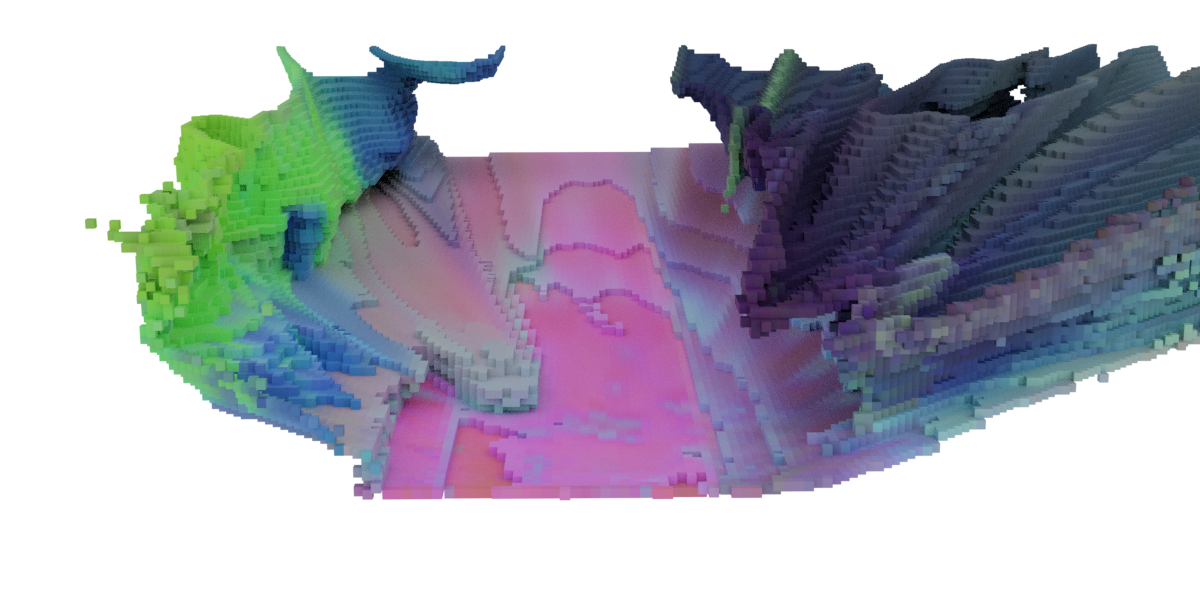} &
\includegraphics[width=\linewidth]{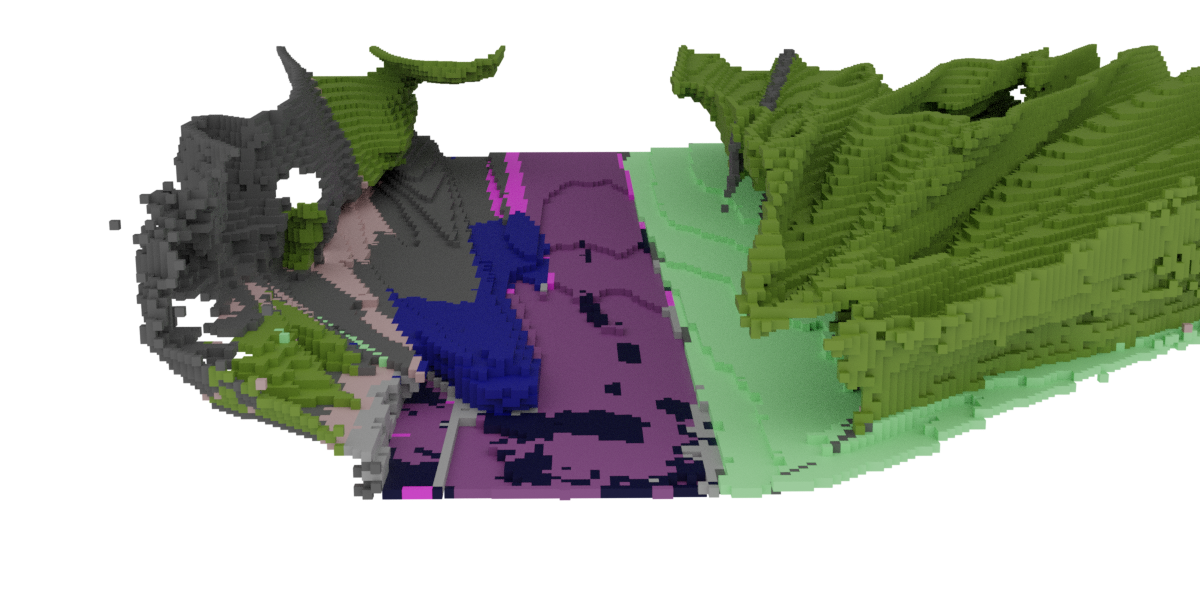} &
\includegraphics[width=\linewidth]{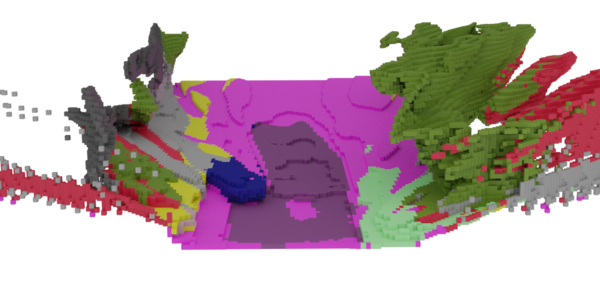} &
\includegraphics[width=\linewidth]{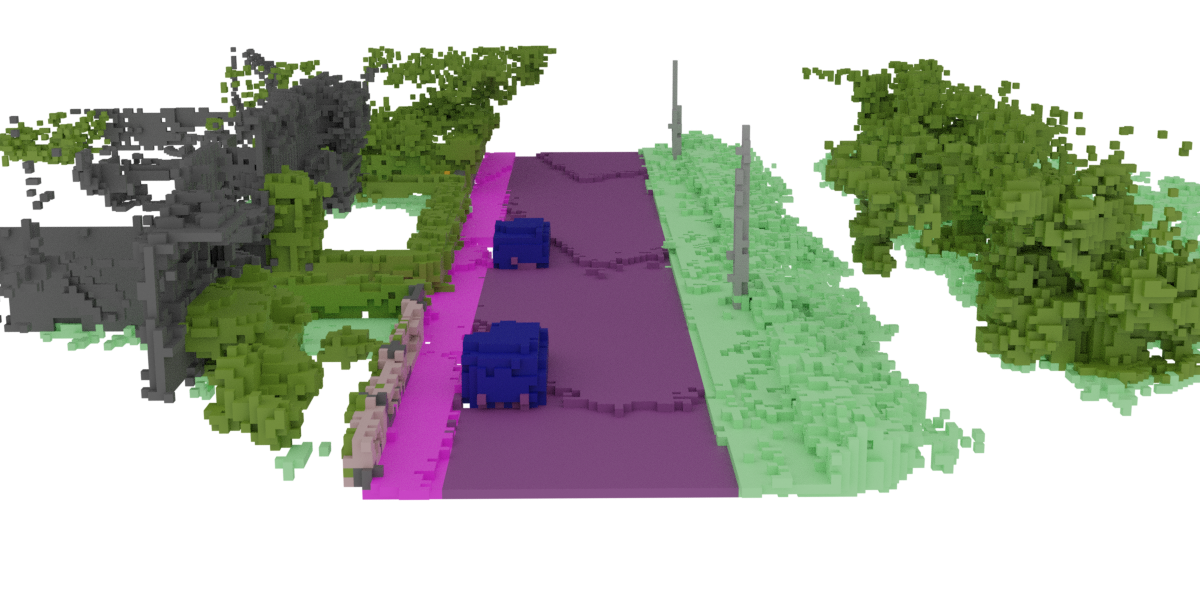} \\[-6.55pt]

\includegraphics[width=\linewidth]{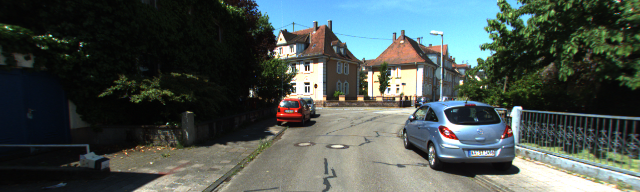} &
\includegraphics[width=\linewidth]{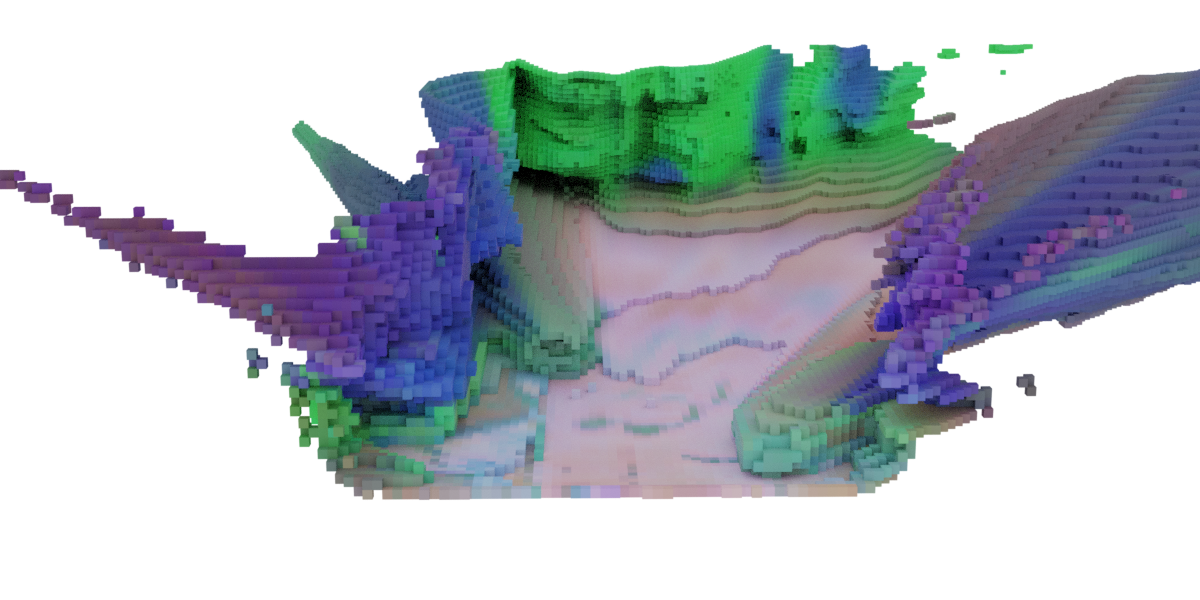} &
\includegraphics[width=\linewidth]{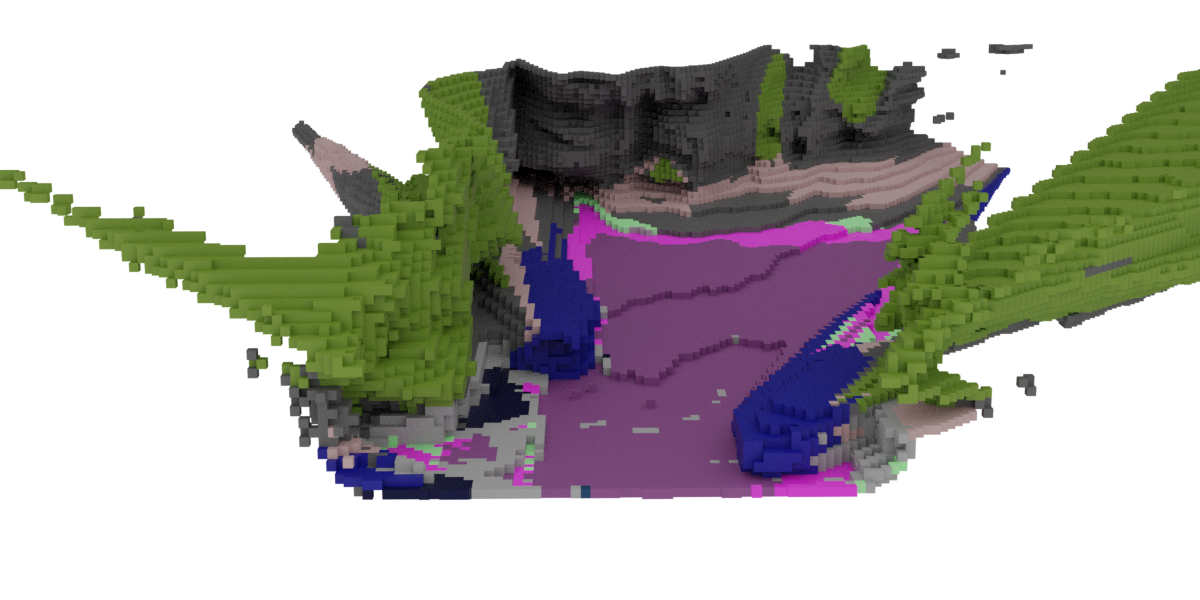} &
\includegraphics[width=\linewidth]{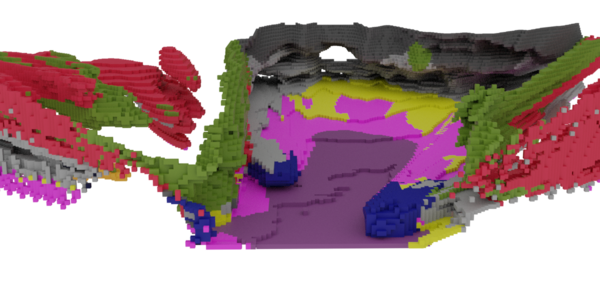} &
\includegraphics[width=\linewidth]{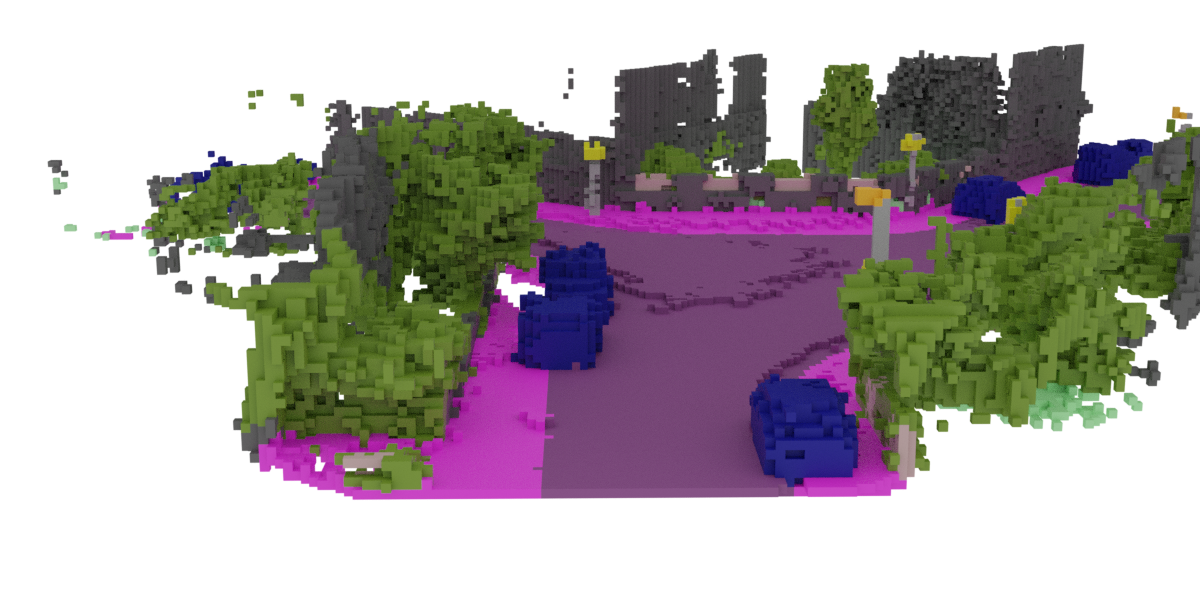} \\

\end{tabular}\\[-5.5pt]%
\tiny%
\renewcommand{\arraystretch}{1.3}%
\begin{tabularx}{0.9982\textwidth}{*{16}{>{\centering\arraybackslash}X}}  
    \cellcolor{road}\textcolor{white}{Road}
    & \cellcolor{sidewalk}\textcolor{white}{Sidewalk}
    & \cellcolor{building}\textcolor{white}{Building}
    & \cellcolor{fence}\textcolor{white}{Fence}
    & \cellcolor{pole}\textcolor{white}{Pole}
    & \cellcolor{trafficlight}\textcolor{white}{Other~Object}
    & \cellcolor{trafficsign}\textcolor{white}{Traffic~Sign}
    & \cellcolor{vegetation}\textcolor{white}{Vegetation}
    & \cellcolor{terrain}\textcolor{white}{Terrain}
    & \cellcolor{person}\textcolor{white}{Person}
    & \cellcolor{car}\textcolor{white}{Car}
    & \cellcolor{truck}\textcolor{white}{Other Vehicle}
    & \cellcolor{motorcycle}\textcolor{white}{Motorcycle}
    & \cellcolor{bicycle}\textcolor{white}{Bicycle}
\end{tabularx}

    \vspace{-0.4em}
    \caption{\textbf{Qualitative SSC comparison on KITTI-360.} We show the input image, \MethodName's feature field using the first three principal components and SSC prediction, the prediction of our baseline S4C\,+\,STEGO, and the ground truth. We only visualize surface voxels. Qualitative results show the expressiveness of our feature field and \MethodName's capabilities to accurately reconstruct and label a scene.\label{fig:ssbench}}%
    \vspace{-0.8em}
\end{figure*}

\inparagraph{3D evaluation.} Given our unsupervised setup, we predict pseudo-semantic classes that must be aligned with the ground truth for evaluation. We follow standard practice in 2D unsupervised semantic segmentation \cite{Cho:2021:PUS, Hamilton:2022:USS, Seong:2023:LHP, Kim:2024:EAL, Sick:2024:USS, Hahn:2024:BUS} by applying Hungarian matching~\cite{Kuhn:1955:THM} to align our pseudo semantics. For validating the aligned semantics, we follow the standardized SSCBench~\cite{Li:2024:SSC} protocol and report both semantic performance using the mean Intersection-over-Union (mIoU) and geometric performance using IoU, precision, and recall. We report all metrics on \mbox{SSCBench} ranges \SI{12.8}{m}, \SI{25.6}{m}, and \SI{51.2}{m}. 

\inparagraph{2D evaluation.} Following the established evaluation protocol in 2D unsupervised semantic segmentation \cite{Cho:2021:PUS, Hamilton:2022:USS, Seong:2023:LHP, Kim:2024:EAL, Sick:2024:USS, Hahn:2024:BUS}, we use the all-pixel accuracy (Acc) and mean Intersection-over-Union (mIoU) metrics. Likewise, in line with prior work, %
2D segmentation predictions of all models are refined using a dense Conditional Random Field~\cite{Krahenbuhl:2011:EFF} before computing Acc and mIoU.

\inparagraph{Multi-view feature consistency evaluation.} We aim to evaluate the multi-view consistency of our feature field. As we are not aware of any general feed-forward 3D feature field approach, we compare against 2D SSL models. To measure multi-view consistency in 2D, we use two video frames and estimate optical flow and occlusions with RAFT~\cite{Teed:2020:RAF}. We backward warp 2D features of the second frame to the first. On the aligned features, we compute the feature similarity using absolute error (L\textsubscript{1}), the Euclidean distance (L\textsubscript{2}), and the cosine similarity, ignoring occlusions.

\inparagraph{Baselines.} We are not aware of any existing unsupervised SSC approach. To allow for comparisons, we construct a competitive baseline for unsupervised SSC. In particular, we train the S4C approach with unsupervised semantics of the established STEGO~\cite{Hamilton:2022:USS} approach. For 2D semantic segmentation, we use U2Seg~\cite{Niu:2024:U2S} and STEGO~\cite{Hamilton:2022:USS} as established unsupervised baselines. Note that U2Seg is trained on ImageNet~\cite{Deng:2009:IMN} and COCO~\cite{Lin:2014:MSC} using STEGO pseudo-labels. We use STEGO~\cite{Hamilton:2022:USS} with DINO~\cite{Caron:2021:EPS} (ViT-B/8), DINOv2~\cite{Oquab:2023:DLR} (ViT-B/14), and FiT3D~\cite{Yue:2024:FIT} (ViT-B/14) features. FiT3D offers multi-view refined DINOv2 features~\cite{Yue:2024:FIT}. Note that FiT3D reports results, concatenating the refined features with DINOv2 features. We report results using both plain features only and the concatenation. We also use rendered 2D segmentations of our S4C + STEGO baseline for 2D validation. For multi-view feature consistency, we utilize DINO~\cite{Caron:2021:EPS}, DINOv2, and FiT3D~\cite{Yue:2024:FIT} features as baselines.

\begin{table}[t]\centering
    \caption{\textbf{SSCBench-KITTI-360 results.} Semantic results using mIoU and per class IoU, and geometric results using IoU, Precision, and Recall (all in \%, $\uparrow$) on SSCBench-KITTI-360 test using three depth ranges. We compare our baseline S4C + STEGO to our \MethodName. We report S4C as a 2D supervised baseline.}
    \vspace{-0.5em}
    \renewcommand\tabcolsep{1.56pt}
\scriptsize
\renewcommand{\arraystretch}{0.865}
\begin{tabularx}{\columnwidth}{lS[table-format=2.2]S[table-format=2.2]S[table-format=2.2]|S[table-format=2.2]S[table-format=2.2]S[table-format=2.2]|S[table-format=2.2]S[table-format=2.2]S[table-format=2.2]}
\toprule
\textbf{Method} & \multicolumn{3}{c|}{\textbf{S4C + STEGO}} & \multicolumn{3}{c|}{\textbf{\MethodName (Ours)}} & \multicolumn{3}{c}{\textbf{\color{tud0c!95} S4C}} \\
\midrule
\textbf{Supervision} & \multicolumn{6}{c}{\cellcolor{LimeGreen!70}\textbf{Unsupervised}} & \multicolumn{3}{c}{\cellcolor{RedOrange!70}\textbf{2D supervision}} \\
\midrule
\textbf{Range} & {\SI{12.8}{\meter}} & {\SI{25.6}{\meter}} & {\SI{51.2}{\meter}} & {\SI{12.8}{\meter}} & {\SI{25.6}{\meter}} & {\SI{51.2}{\meter}} & \color{tud0c!95} {\SI{12.8}{\meter}} & \color{tud0c!95} {\SI{25.6}{\meter}} & \color{tud0c!95} {\SI{51.2}{\meter}} \\
\specialrule{0.5pt}{1.5pt}{2pt}
\multicolumn{10}{c}{\textit{Semantic validation}} \\
\specialrule{0.5pt}{1.5pt}{2pt}
\cellcolor{tud0c!20}\textbf{mIoU} & \cellcolor{tud0c!20}10.53 & \cellcolor{tud0c!20}9.26 & \cellcolor{tud0c!20}6.60 & \cellcolor{tud0c!20}\bfseries 10.76 & \cellcolor{tud0c!20}\bfseries 10.01 & \cellcolor{tud0c!20}\bfseries 8.00 & \cellcolor{tud0c!20}\color{tud0c!95}16.94 & \cellcolor{tud0c!20}\color{tud0c!95}13.94 & \cellcolor{tud0c!20}\color{tud0c!95}10.19 \\
car           & 18.57 & 14.09 & 9.22 & 21.24 & 15.94 & 11.21 & \color{tud0c!95}22.58 & \color{tud0c!95}18.64 & \color{tud0c!95}11.49 \\
bicycle       & 0.01 & 0.01 & 0.01 &  0.00 & 0.00& 0.00  & \color{tud0c!95}0.00 & \color{tud0c!95}0.00 &  \color{tud0c!95}0.00 \\
motorcycle    & 0.00 & 0.00 & 0.00 &  0.00 & 0.00& 0.00  & \color{tud0c!95}0.00 & \color{tud0c!95}0.00 & \color{tud0c!95}0.00 \\
truck         & 0.11 & 0.04 & 0.02 &  0.00 & 0.00& 0.00  & \color{tud0c!95}7.51 &\color{tud0c!95} 4.37 & \color{tud0c!95}2.12 \\
other-v. & 0.01 & 0.05 & 0.02 &  0.00 & 0.00& 0.00  & \color{tud0c!95}0.00 & \color{tud0c!95}0.01 & \color{tud0c!95}0.06 \\
person        & 0.01 & 0.01 & 0.01 &  0.00 & 0.00& 0.00  & \color{tud0c!95}0.00 & \color{tud0c!95}0.00 & \color{tud0c!95}0.00 \\
road          & 61.97 & 52.47 & 38.15 &  51.10 & 49.12 & 39.82  & \color{tud0c!95}69.38 & \color{tud0c!95}61.46 & \color{tud0c!95}48.23 \\
sidewalk      & 18.74 & 20.95 & 18.21 &  20.26 & 22.31 & 18.97  & \color{tud0c!95}45.03 & \color{tud0c!95}37.12 & \color{tud0c!95}28.45 \\
building      & 14.75 & 24.44 & 17.81 &  12.33 & 18.27 & 14.32  & \color{tud0c!95}26.34 & \color{tud0c!95}28.48 & \color{tud0c!95}21.36 \\
fence         & 1.41 & 0.20 & 0.11 &  1.91 & 0.90 & 0.58  & \color{tud0c!95}9.70 & \color{tud0c!95}6.37 & \color{tud0c!95}3.64 \\
vegetation    & 15.83 & 16.58 & 11.30 &  31.22 & 25.57 & 19.85  & \color{tud0c!95}35.78 & \color{tud0c!95}28.04 & \color{tud0c!95}21.43 \\
terrain       & 26.49 &  9.95 & 4.17 &  23.26 & 18.02 & 15.22  & \color{tud0c!95}35.03 & \color{tud0c!95}22.88 & \color{tud0c!95}15.08 \\
pole          & 0.08 & 0.04 & 0.04 &  0.05 & 0.05 & 0.05  & \color{tud0c!95}1.23 & \color{tud0c!95}0.94 & \color{tud0c!95}0.65 \\
traffic-sign  & 0.00 & 0.00 & 0.00 &  0.00 & 0.00 & 0.00  & \color{tud0c!95}1.57 & \color{tud0c!95}0.83 & \color{tud0c!95}0.36 \\
other-obj. & 0.05 & 0.04 & 0.02 &  0.00 & 0.00 & 0.00  & \color{tud0c!95}0.00 & \color{tud0c!95}0.00 & \color{tud0c!95}0.00 \\
\specialrule{0.5pt}{1.5pt}{2pt}
\multicolumn{10}{c}{\textit{Geometric validation}} \\
\specialrule{0.5pt}{1.5pt}{2pt}
\cellcolor{tud0c!20}\textbf{IoU}           & \cellcolor{tud0c!20}49.32 & \cellcolor{tud0c!20}41.08 & \cellcolor{tud0c!20}36.39 & \cellcolor{tud0c!20}\bfseries 49.54 & \cellcolor{tud0c!20}\bfseries 42.27 & \cellcolor{tud0c!20}\bfseries 37.60 & \cellcolor{tud0c!20}\color{tud0c!95}54.64 & \cellcolor{tud0c!20}\color{tud0c!95}45.57 & \cellcolor{tud0c!20}\color{tud0c!95}39.35 \\
Precision     & 54.04 & 46.23 & 41.91 &  53.27 & 46.10 & 41.59  & \color{tud0c!95}59.75 & \color{tud0c!95}50.34 & \color{tud0c!95}43.59  \\
Recall        & 84.95 & 78.69 & 73.43 & 87.61 & 83.59 & 79.67  & \color{tud0c!95}86.47 & \color{tud0c!95}82.79 & \color{tud0c!95}80.16  \\
\bottomrule
\end{tabularx}

    \label{tab:sscbench}
    \vspace{-0.8em}
\end{table}

\inparagraph{Implementation details.} Our encoder-decoder uses a DINO-B/8~\cite{Caron:2021:EPS} backbone and a dense prediction decoder~\cite{Ranftl:2021:DPT}. The MLP decoder $\phi$ entails two layers with \num{128} hidden features. As rendering features is expensive, $\phi$ predicts \num{64} features. We employ another MLP to up-project again to the full dimensionality ${\rm D}=768$. If not stated differently, our target features are obtained from DINO-B/8~\cite{Caron:2021:EPS}. We train using a batch size of \num{4} and extract \num{32} patches of size $8 \times 8$ from each image to compute $\mathcal{L}_{\text{\MethodName}}$. Volume rendering samples each ray at $\mathrm{L}=32$ uniformly spaced points in inverse depth within $[3\,\text{m},80\,\text{m}]$. We train for \SI{100}{k} steps using Adam~\cite{Kingma:2015:AMS} with a base learning rate of $10^{-4}$. Training takes ca.\ 2 days on a \emph{single} V100 GPU. We distill using a batch size of \num{4}, \num{5} center points, a feature batch of size \num{576}, and cluster with $K=19$. For $k$NN sampling, we use $k=4$. The feature buffer holds \num{256} feature batches. Refer to the supplement for more details.

\begin{table}[t]
    \centering
    \caption{\textbf{2D unsupervised semantic segmentation results on KITTI-360}. Comparing \MethodName to existing 2D methods and our S4C + STEGO 3D baseline, using Accuracy and mean IoU (in \%, $\uparrow$) on the SSCBench-KITTI-360 test split. \textdagger\,denotes the use of plain FiT3D features. $\ddagger$\ denotes training on ImageNet and COCO.}
    \vspace{-0.5em}
    \sisetup{table-number-alignment=center}

\setlength{\tabcolsep}{8.05pt}
\footnotesize
\renewcommand{\arraystretch}{0.865}
\begin{tabularx}{\columnwidth}{>{\hspace{-\tabcolsep}\raggedright\columncolor{white}[\tabcolsep][\tabcolsep]}l>{\centering\arraybackslash}lp{0.15cm}S[table-format=2.2]S[table-format=2.2]}
	\toprule
    {\textbf{Method}} & {\textbf{Features}} &  & {\textbf{Acc}} & {\textbf{mIoU}}\\
    \midrule
    U2Seg$^\ddagger$~\cite{Niu:2024:U2S} & \hphantom{Feat}-- &  & 72.89 & 23.43 \\
    STEGO~\cite{Hamilton:2022:USS} & DINO~\cite{Caron:2021:EPS} &  & 73.32 & 23.57 \\
    STEGO~\cite{Hamilton:2022:USS} & DINOv2~\cite{Oquab:2023:DLR} &  & 64.54 & 24.82 \\
    STEGO~\cite{Hamilton:2022:USS} & FiT3D~\cite{Yue:2024:FIT} &  & 54.19 & 22.29 \\
    STEGO~\cite{Hamilton:2022:USS} & FiT3D\textsuperscript{\textdagger}~\cite{Yue:2024:FIT} &  & 57.25 & 18.95 \\
    S4C~\cite{Hayler:2024:S4C} + STEGO~\cite{Hamilton:2022:USS} & DINO~\cite{Caron:2021:EPS} &  & 65.16 & 21.67 \\
    \specialrule{0.5pt}{0pt}{2pt}
    \rowcolor{tud0c!20} \MethodName (Ours) & DINO~\cite{Caron:2021:EPS} &  & \bfseries 77.74 & \bfseries 25.81 \\
	\bottomrule
\end{tabularx}%

    \label{tab:semseg}
    \vspace{-0.6em}
\end{table}%

\subsection{3D semantic scene completion\label{subsec:results_3d}}

We assess the unsupervised SSC and geometric accuracy of \MethodName with our 3D feature distillation approach on SSCBench-KITTI-360. In particular, \cref{tab:sscbench} compares \MethodName against our unsupervised SSC baseline S4C~\cite{Hayler:2024:S4C} + STEGO~\cite{Hamilton:2022:USS}. \MethodName achieves a (semantic) mIoU of \SI{8.0}{\percent} for the range of \SI{51.2}{m}, significantly improving over our unsupervised baseline (\SI{6.6}{\percent}). This demonstrates that \MethodName effectively lifts DINO features into 3D. In terms of geometric accuracy, \MethodName moderately improves over S4C + STEGO. 
Despite being \emph{fully unsupervised}, \MethodName comes within \SI{2.2}{\percent} points mIoU of the 2D-supervised S4C.

\Cref{fig:ssbench} provides qualitative samples on SSCBench-KITTI-360. {\MethodName's} unsupervised SSC predictions are less noisy and capture finely resolved semantics compared to S4C + STEGO. Compared to the ground truth, we observe that \MethodName captures both the geometry and general semantics of the scene. We visualize \MethodName's feature field (before distillation) using the first three principal components. In PCA space, we observe that our feature field captures semantically meaningful regions.

\begin{table}[t]
    \centering
    \caption{\textbf{2D unsupervised semantic segmentation domain generalization results.} Comparing \MethodName to existing 2D unsupervised semantic segmentation methods and S4C + STEGO 3D baseline, using Accuracy and mean IoU (in \%, $\uparrow$). We train on KITTI-360 images and report domain generalization results on Cityscapes and BDD-100K val. \textdagger\,denotes plain FiT3D features.}
    \vspace{-0.5em}
    \sisetup{table-number-alignment=center}
\setlength{\tabcolsep}{3.55pt}
\footnotesize
\renewcommand{\arraystretch}{0.865}
\begin{tabularx}{\columnwidth}{>{\hspace{-\tabcolsep}\raggedright\columncolor{white}[\tabcolsep][\tabcolsep]}l>{\centering\arraybackslash}lS[table-format=2.2]S[table-format=2.2]S[table-format=2.2]S[table-format=2.2]}
	\toprule
    & & \multicolumn{2}{c}{\textbf{Cityscapes}} & \multicolumn{2}{c}{\textbf{BDD-100K}} \\
    \cmidrule(l{0.2em}r{0.2em}){3-4} \cmidrule(l{0.2em}r{0.2em}){5-6}
    \multirow{-2}{*}{\vspace{0.5em}\textbf{Method}} & \multirow{-2}{*}{\vspace{0.5em}\textbf{Features}} & {\textbf{Acc}} & {\textbf{mIoU}} & {\textbf{Acc}} & {\textbf{mIoU}}\\
    \midrule
    U2Seg~\cite{Niu:2024:U2S} & \hphantom{Feat}-- & \bfseries 75.57 & 18.62 & 69.00 & 17.99 \\
    STEGO~\cite{Hamilton:2022:USS} & DINO~\cite{Caron:2021:EPS} & 71.21 & 19.42 & \bfseries 75.02 & 21.41 \\
    STEGO~\cite{Hamilton:2022:USS} & DINOv2~\cite{Oquab:2023:DLR} & 68.41 & 19.73 & 65.72 & 21.77 \\
    STEGO~\cite{Hamilton:2022:USS} & FiT3D~\cite{Yue:2024:FIT} & 66.94 & 21.01 & 65.96 & 20.99 \\
    STEGO~\cite{Hamilton:2022:USS} & FiT3D\textsuperscript{\textdagger}~\cite{Yue:2024:FIT} & 64.76 & 17.17 & 60.83 & 19.09 \\
    S4C~\cite{Hayler:2024:S4C} + STEGO~\cite{Hamilton:2022:USS} & DINO~\cite{Caron:2021:EPS} & 54.80 & 14.04 & 44.98 & 11.62 \\
    \specialrule{0.5pt}{0pt}{2pt}%
    \rowcolor{tud0c!20} \MethodName (Ours) & DINO~\cite{Caron:2021:EPS} & 73.17 & \bfseries 22.81 & 72.28 & \bfseries 22.09 \\
	\bottomrule
\end{tabularx}%

    \label{tab:zero_shot}
    \vspace{-0.6em}
\end{table}

\subsection{2D semantic segmentation\label{subsec:results_2d}}

\Cref{tab:semseg} compares the semantic predictions of \MethodName to recent 2D approaches and our 3D baseline. We obtain 2D semantic segmentations from \MethodName and our S4C + STEGO baseline using semantic rendering~\cite{Hayler:2024:S4C}. \MethodName with our 3D distillation approach outperforms STEGO with DINO features, an established 2D unsupervised semantic segmentation approach. In particular, the mIoU of \MethodName is \SI{2.24}{\percent} points higher than for STEGO (w/ DINO). Utilizing 3D refined features from FiT3D deteriorates the baseline relative to DINO, indicating that the FiT3D refinement reduces feature expressiveness. Notably, our unsupervised 3D baseline S4C + STEGO transfers significantly worse to 2D than \MethodName.

We also validate \MethodName, trained on KITTI-360, on Cityscapes and BDD10K, demonstrating domain generalization. The results are reported in \cref{tab:zero_shot}. \MethodName outperforms all baselines in mIoU on both datasets while only falling short in Acc. S4C + STEGO falls short in generalization. We suspect this poor generalization is caused by the fact that S4C does not rely on general SSL features in the final model, while our feature field generalizes.

\subsection{Multi-view feature consistency\label{subsec:results_mvc}}

We analyze the multi-view consistency of our feature field against existing 2D SSL features in \cref{tab:mvc}. We report the results of \MethodName trained on KITTI-360 and RealEstate10K. \MethodName trained using DINO features exhibits significant improvements in multi-view feature consistency over standard DINO features. We also train \MethodName using target features from DINOv2~\cite{Oquab:2023:DLR}. Compared to standard DINOv2 and FiT3D features, \MethodName's feature field yields significantly better multi-view consistency. Notably, compared against plain 3D refined features of FiT3D, \MethodName shows a better multi-view consistency on both datasets and all metrics while also offering more expressiveness (\cf \cref{tab:semseg}).

\begin{table}[t]
    \centering
    \caption{\textbf{Multi-view consistency results}. Comparing multi-view consistency of \MethodName to existing 2D SSL features, using L\textsubscript{1} distance ($\downarrow$), L\textsubscript{2} distance ($\downarrow$), and cosine similarity ($\uparrow$) on KITTI-360 and RealEstate10K. We compare DINO (\emph{top}) and DINOv2-based (\emph{bottom}) features. \textdagger\,denotes plain FiT3D features.}
    \vspace{-0.5em}
    \setlength{\tabcolsep}{2.3pt}
\footnotesize
\renewcommand{\arraystretch}{0.865}
\begin{tabularx}{\columnwidth}{>{\hspace{-\tabcolsep}\raggedright\columncolor{white}[\tabcolsep][\tabcolsep]}lS[table-format=2.2]S[table-format=2.2]S[table-format=2.2]S[table-format=2.2]S[table-format=2.2]S[table-format=2.2]}
	\toprule
    & \multicolumn{3}{c}{\textbf{KITTI-360}} & \multicolumn{3}{c}{\textbf{RealEstate10K}} \\
    \cmidrule(l{0.2em}r{0.2em}){2-4} \cmidrule(l{0.2em}r{0.2em}){5-7}
    \multirow{-2}{*}{\vspace{0.5em}\textbf{Method}} & {\textbf{L\textsubscript{1}}} & {\textbf{L\textsubscript{2}}} & {\textbf{cos-sim}} & {\textbf{L\textsubscript{1}}} & {\textbf{L\textsubscript{2}}} & {\textbf{cos-sim}} \\
    \midrule
    DINO~\cite{Caron:2021:EPS} & 16.06 & 0.74 & 0.70 & 14.41 & 0.66 & 0.75 \\
    \rowcolor{tud0c!20} \MethodName (w/ DINO) & \bfseries 6.45 & \bfseries 0.33 & \bfseries 0.93 & \bfseries 5.87 & \bfseries 0.28 & \bfseries 0.95 \\
    \midrule
    DINOv2~\cite{Oquab:2023:DLR} & 15.83 & 0.73 & 0.70 & 14.20 & 0.66 & 0.75 \\
    FiT3D~\cite{Yue:2024:FIT} & 22.86 & 0.81 & 0.82 & 19.88 & 0.72 & 0.85 \\
    FiT3D\textsuperscript{\textdagger}~\cite{Yue:2024:FIT} & 7.02 & 0.33 & 0.93 & 5.67 & 0.27 & 0.95 \\
    \rowcolor{tud0c!20} \MethodName (w/ DINOv2) & \bfseries 5.24 & \bfseries 0.24 & \bfseries 0.96 & \bfseries 4.87 & \bfseries 0.22 & \bfseries 0.97 \\
	\bottomrule
\end{tabularx}%

    \vspace{-0.6em}%
    \label{tab:mvc}%
\end{table}

\subsection{Analyzing \MethodName \label{subsec:analyzing_scenedino}}

\begin{table}[t]
    \centering
    \footnotesize
    \setlength{\tabcolsep}{1.7pt}
    \setlength\fboxsep{0pt}
    \renewcommand{\arraystretch}{0.865}
    \caption{\textbf{\MethodName analysis.} We analyze the role of decomposing positional encodings, the choice of downsampling features during training, the effectiveness of the feature smoothness loss, the effect of estimated camera poses, and the choice of target features. We report the mean IoU (in \%, $\uparrow$) using a range of \SI{51.2}{m} on SSCBench-KITTI-360 test. $\Delta\,$mIoU reports the absolute difference in \% points to our standard model with DINO target features.\label{tab:analysis}}
    \vspace{-0.5em}
    \begin{tabularx}{\linewidth}{@{}>{\raggedleft\arraybackslash}m{31mm}|S[table-format=1.2]|X}
    \toprule
        \multicolumn{1}{>{\centering\arraybackslash}m{31mm}|}{$\Delta$ \textbf{mIoU}} & \textbf{mIoU} & \textbf{Configuration} \\
        \midrule
        -1.18 \cbarm{23.9}{5}{21.6} \hphantom{+1.08} & 6.82 & No downsampler (bilinear up.\,+\,aug.) \\
        -1.17 \cbarm{23.4}{5}{21.6} \hphantom{+1.08} & 6.83 & No feature smoothness loss ($\lambda_{\text{fs}}=0$)\!\!\!\!\! \\
        -0.74 \cbarm{14.8}{5}{21.6} \hphantom{+1.08} & 7.26 & No pos.\ enc.\ decomposition \\
        -0.12 \cbarm{2.4}{5}{21.6} \hphantom{+1.08} & 7.88 & w/ estimated ORB-SLAM3 poses \\
        \rowcolor{tud0c!20} \multicolumn{1}{>{\centering\arraybackslash}m{31mm}|}{---} & 8.00 & Full framework (\MethodName) \\
        \cbarp{0}{5}{21.6} +1.08 & 9.08 & DINOv2 target features (\vs DINO) \\
        \bottomrule
    \end{tabularx}
    \vspace{-0.6em}
\end{table}

To understand what core components contribute to obtaining an expressive feature field of \MethodName, we omit or replace individual components and report the results in \cref{tab:analysis}. Replacing the downsampling approach with bilinear upsampling and multi-crop augmentations, similar to~\cite{Araslanov:2021:SSA}, to obtain high-resolution target features decreases the SSC mIoU by \SI{1.18}{\percent}. Omitting the feature smoothness loss leads to a similar mIoU drop. Abolishing the constant decomposition of positional encodings leads to an mIoU drop of \SI{0.74}{\percent}. Training using unsupervised camera poses estimated by ORB-SLAM3~\cite{Campos:2021:ORB} results in an insignificant mIoU drop of only \SI{0.12}{\percent} over using KITTI-360 poses. Going from DINO to DINOv2 target features leads to an increased mIoU of \SI{1.08}{\percent}, demonstrating that \MethodName can benefit from more expressive 2D target features.

\begin{table}[t]
    \centering
    \footnotesize
    \setlength{\tabcolsep}{1.7pt}
    \setlength\fboxsep{0pt}
    \renewcommand{\arraystretch}{0.815}
    \caption{\textbf{Feature distillation analysis.} We analyze the effectiveness of distilling \MethodName's features, the $k$NN-correlation loss, our neighborhood sampling, and our 3D sampling approach over standard 5-crop sampling. We report the mean IoU (in \%, $\uparrow$) using a range of \SI{51.2}{m} on SSCBench-KITTI-360 test.\label{tab:distillation}}
    \vspace{-0.625em}
    \begin{tabularx}{\linewidth}{@{}>{\raggedleft\arraybackslash}m{15mm}|S[table-format=1.2]|X}
    \toprule
        \multicolumn{1}{>{\centering\arraybackslash}m{15mm}|}{$\Delta$ \textbf{mIoU}} & \textbf{mIoU} & \textbf{Configuration} \\
        \midrule
        -1.61 \cbarm{16.1}{5}{0} & 6.39 & No distillation \\
        -1.35 \cbarm{13.5}{5}{0} & 6.65 & No $k$NN-correlation loss ($\lambda_{k\text{NN}}=0$) \\
        -0.97 \cbarm{9.7}{5}{0} & 7.03 & No neighborhood sampling (\cf \cref{fig:feature_sampling}) \\ %
        -0.47 \cbarm{4.7}{5}{0} & 7.53 & 5-crop sampling~\cite{Hamilton:2022:USS} (instead 3D sampling) \\ %
        \rowcolor{tud0c!20} \multicolumn{1}{>{\raggedleft\arraybackslash}m{15mm}|}{---$\;\;\;$} & 8.00 & Full framework (\MethodName) \\
        \bottomrule
    \end{tabularx}
    \vspace{-0.6em}%
\end{table}

\begin{table}[t]\centering
    \caption{\textbf{Probing analysis.} We analyze linear and unsupervised probing of our distilled \MethodName features on SSCBench-KITTI-360 test using mean IoU (in \%, $\uparrow$). For reference, we also report S4C (2D supervised). Linear probing uses 2D annotations.}
    \vspace{-0.5em}
    \sisetup{table-number-alignment=center}
\setlength{\tabcolsep}{6.5525pt}
\footnotesize
\renewcommand{\arraystretch}{0.865}
\begin{tabularx}{\columnwidth}{>{\hspace{-\tabcolsep}\raggedright\columncolor{white}[\tabcolsep][\tabcolsep]}llS[table-format=2.2]S[table-format=2.2]S[table-format=2.2]}
	\toprule
     & & \multicolumn{3}{c}{\textbf{mIoU}} \\
    \cmidrule(l{0.2em}r{0.0em}){3-5}
    \multirow{-2}{*}{\vspace{0.5em}\textbf{Probing approach}} & \multirow{-2}{*}{\vspace{0.5em}\textbf{Target features}} & {\SI{12.8}{m}} & {\SI{25.6}{m}} & {\SI{51.2}{m}} \\
    \midrule
     & DINO~\cite{Caron:2021:EPS} & 10.76 & 10.01 & 8.00 \\
    \multirow{-2}{*}{Unsupervised} & DINOv2~\cite{Oquab:2023:DLR} & 13.76 & 11.78 & 9.08 \\
    \specialrule{0.5pt}{1pt}{2pt}
     & DINO~\cite{Caron:2021:EPS} & 13.63 & 12.07 & 9.34 \\
    \multirow{-2}{*}{Linear} & DINOv2~\cite{Oquab:2023:DLR} & 15.85 & 13.70 & 10.57 \\
    \specialrule{0.5pt}{1pt}{2pt}
    \color{tud0c!95} S4C (full training) & {\color{tud0c!95}\;\;\;\;\;\; n/a} & \color{tud0c!95} 16.94 & \color{tud0c!95} 13.94 & \color{tud0c!95} 10.19 \\
	\bottomrule
\end{tabularx}%

    \vspace{-0.6em}
    \label{tab:probing}
\end{table}

In \cref{tab:distillation}, we analyze our 3D distillation. Performing no distillation at all, just feature clustering, decreases mIoU by \SI{1.61}{\percent}. Omitting the $k$NN-correlation loss leads to an mIoU drop of \SI{1.35}{\percent}. Distilling only with center points, \ie, not performing neighborhood sampling (\cf \cref{fig:feature_sampling}), reduces mIoU by \SI{0.97}{\percent}. Using 5-crop feature sampling~\cite{Hamilton:2022:USS}, instead of our proposed 3D sampling, leads to a reduced mIoU of \SI{0.47}{\percent}. This demonstrates the effectiveness of performing distillation in 3D using our novel approach.

While focusing on unsupervised SSC, we can also linearly probe our distilled feature field (\cf \cref{tab:probing}). In particular, we train \MethodName using different target features (DINO~\cite{Caron:2021:EPS} and DINOv2~\cite{Caron:2021:EPS}), perform distillation, and probe the resulting distilled features. Using linear probing, \ie, training a \emph{single} linear layer using 2D semantic labels, leads to a consistent mIoU increase over unsupervised probing. \MethodName trained using DINOv2 target features even closes the gap to S4C, trained using 2D ground-truth semantic labels. We even surpass 2D supervised S4C slightly on the full range (\SI{51.2}{m}), suggesting the effectiveness of \MethodName also for weakly-supervised tasks.

\section{Conclusion\label{sec:conclusion}}
We presented \MethodName, to our knowledge, the first approach for unsupervised semantic scene completion. Trained using multi-view images and 2D DINO features without human supervision, \MethodName is able to predict an expressive 3D feature field using a single input image during inference. Our novel 3D distillation approach yields state-of-the-art results in unsupervised SSC. While we focus on unsupervised SSC, our multi-view feature consistency, linear probing, and domain generalization results highlight the potential of \MethodName as a strong foundation for various 3D scene-understanding tasks.

{\small \inparagraph{Acknowledgments.} This project was partially supported by the European Research Council (ERC) Advanced Grant SIMULACRON, DFG project CR 250/26-1 ``4D-YouTube'', and GNI Project ``AICC''. This project was also partially supported by the ERC under the European Union’s Horizon 2020 research and innovation programme (grant agreement No.\ 866008). Additionally, this work has been co-funded by the LOEWE initiative (Hesse, Germany) within the emergenCITY center [LOEWE/1/12/519/03/05.001(0016)/72] and by the Deutsche Forschungsgemeinschaft (German Research Foundation, DFG) under Germany's Excellence Strategy (EXC 3066/1 ``The Adaptive Mind'', Project No.\ 533717223). Christoph Reich is supported by the Konrad Zuse School of Excellence in Learning and Intelligent Systems (ELIZA) through the DAAD programme Konrad Zuse Schools of Excellence in Artificial Intelligence, sponsored by the Federal Ministry of Education and Research. Christian Rupprecht is supported by an Amazon Research Award. Finally, we acknowledge the support of the European Laboratory for Learning and Intelligent Systems (ELLIS) and thank Mateo de Mayo as well as Igor Cvišić for help with estimating camera poses.}

{
    \small
    \bibliographystyle{ieeenat_fullname}
    \bibliography{bibtex/short, references}

\begin{thebibliography}{120}
\providecommand{\natexlab}[1]{#1}
\providecommand{\url}[1]{\texttt{#1}}
\expandafter\ifx\csname urlstyle\endcsname\relax
  \providecommand{\doi}[1]{doi: #1}\else
  \providecommand{\doi}{doi: \begingroup \urlstyle{rm}\Url}\fi

\bibitem[Araslanov and Roth(2021)]{Araslanov:2021:SSA}
Nikita Araslanov and Stefan Roth.
\newblock Self-supervised augmentation consistency for adapting semantic
  segmentation.
\newblock In \emph{CVPR}, pages 15384--15394, 2021.

\bibitem[Asano et~al.(2020)Asano, Rupprecht, and Vedaldi]{Asano:2020:SLC}
Yuki~Markus Asano, Christian Rupprecht, and Andrea Vedaldi.
\newblock Self-labelling via simultaneous clustering and representation
  learning.
\newblock In \emph{ICLR}, 2020.

\bibitem[Bachman et~al.()Bachman, Hjelm, and Buchwalter]{Bachman:2019:LRM}
Philip Bachman, R.~Devon Hjelm, and William Buchwalter.
\newblock Learning representations by maximizing mutual information across
  views.
\newblock In \emph{NeurIPS*2019}, pages 15509--15519.

\bibitem[Bardes et~al.()Bardes, Ponce, and LeCun]{Bardes:2022:VIL}
Adrien Bardes, Jean Ponce, and Yann LeCun.
\newblock {VICRegL}: {S}elf-supervised learning of local visual features.
\newblock In \emph{NeurIPS*2022}, pages 8799--8810.

\bibitem[Bardes et~al.(2022)Bardes, Ponce, and LeCun]{Bardes:2022:VIC}
Adrien Bardes, Jean Ponce, and Yann LeCun.
\newblock {VICR}eg: {V}ariance-invariance-covariance regularization for
  self-supervised learning.
\newblock In \emph{ICLR}, 2022.

\bibitem[Cai et~al.(2021)Cai, Chen, Zhang, Lin, Wang, and Li]{Cai:2021:SSC}
Yingjie Cai, Xuesong Chen, Chao Zhang, Kwan{-}Yee Lin, Xiaogang Wang, and
  Hongsheng Li.
\newblock Semantic scene completion via integrating instances and scene
  in-the-loop.
\newblock In \emph{CVPR}, pages 324--333, 2021.

\bibitem[Campos et~al.(2021)Campos, Elvira, Rodr{\'\i}guez, Montiel, and
  Tard{\'o}s]{Campos:2021:ORB}
Carlos Campos, Richard Elvira, Juan J.~G{\'o}mez Rodr{\'\i}guez, Jos{\'e} M.~M.
  Montiel, and Juan~D Tard{\'o}s.
\newblock {ORB-SLAM3}: {A}n accurate open-source library for visual,
  visual-inertial and multi-map {SLAM}.
\newblock \emph{{IEEE Trans. Robot.}}, 37\penalty0 (6):\penalty0 1874--1890,
  2021.

\bibitem[Cao and de~Charette(2022)]{Cao:2022:M3S}
Anh{-}Quan Cao and Raoul de Charette.
\newblock Monoscene: Monocular {3D} semantic scene completion.
\newblock In \emph{CVPR}, pages 3981--3991, 2022.

\bibitem[Caron et~al.()Caron, Misra, Mairal, Goyal, Bojanowski, and
  Joulin]{Caron:2020:ULV}
Mathilde Caron, Ishan Misra, Julien Mairal, Priya Goyal, Piotr Bojanowski, and
  Armand Joulin.
\newblock Unsupervised learning of visual features by contrasting cluster
  assignments.
\newblock In \emph{NeurIPS*2020}, pages 9912--9924.

\bibitem[Caron et~al.(2018)Caron, Bojanowski, Joulin, and
  Douze]{Caron:2018:DCL}
Mathilde Caron, Piotr Bojanowski, Armand Joulin, and Matthijs Douze.
\newblock Deep clustering for unsupervised learning of visual features.
\newblock In \emph{ECCV}, pages 132--149, 2018.

\bibitem[Caron et~al.(2021)Caron, Touvron, Misra, J{\'e}gou, Mairal,
  Bojanowski, and Joulin]{Caron:2021:EPS}
Mathilde Caron, Hugo Touvron, Ishan Misra, Herv{\'e} J{\'e}gou, Julien Mairal,
  Piotr Bojanowski, and Armand Joulin.
\newblock Emerging properties in self-supervised vision transformers.
\newblock In \emph{ICCV}, pages 9650--9660, 2021.

\bibitem[Chen et~al.(2020{\natexlab{a}})Chen, Fan, Girshick, and
  He]{Chen:2020:ISL}
Xinlei Chen, Haoqi Fan, Ross Girshick, and Kaiming He.
\newblock Improved baselines with momentum contrastive learning.
\newblock \emph{arXiv:2003.04297 [cs.CV]}, 2020{\natexlab{a}}.

\bibitem[Chen et~al.(2020{\natexlab{b}})Chen, Lin, Qian, Zeng, and
  Li]{Chen:2020:3SA}
Xiaokang Chen, Kwan{-}Yee Lin, Chen Qian, Gang Zeng, and Hongsheng Li.
\newblock 3d sketch-aware semantic scene completion via semi-supervised
  structure prior.
\newblock In \emph{CVPR}, pages 4192--4201, 2020{\natexlab{b}}.

\bibitem[Chen et~al.(2021)Chen, Xie, and He]{Chen:2021:SSL}
Xinlei Chen, Saining Xie, and Kaiming He.
\newblock An empirical study of training self-supervised vision transformers.
\newblock In \emph{CVPR}, pages 9640--9649, 2021.

\bibitem[Chen et~al.()Chen, Zheng, Xu, Zhuang, Vedaldi, Cham, and
  Cai]{Chen:2024:MVS}
Yuedong Chen, Chuanxia Zheng, Haofei Xu, Bohan Zhuang, Andrea Vedaldi, Tat-Jen
  Cham, and Jianfei Cai.
\newblock {MVSplat360}: {F}eed-forward 360 scene synthesis from sparse views.
\newblock In \emph{NeurIPS*2024}, pages 107064--107086.

\bibitem[Cheng et~al.(2020)Cheng, Agia, Ren, Li, and Liu]{Cheng:2020:SSS}
Ran Cheng, Christopher Agia, Yuan Ren, Xinhai Li, and Bingbing Liu.
\newblock {S3CNet}: {A} sparse semantic scene completion network for {LiDAR}
  point clouds.
\newblock In \emph{CoRL}, pages 2148--2161, 2020.

\bibitem[Cho et~al.(2021)Cho, Mall, Bala, and Hariharan]{Cho:2021:PUS}
Jang~Hyun Cho, Utkarsh Mall, Kavita Bala, and Bharath Hariharan.
\newblock Pi{C}{I}{E}: {U}nsupervised semantic segmentation using invariance
  and equivariance in clustering.
\newblock In \emph{CVPR}, pages 16794--16804, 2021.

\bibitem[{\c{C}}i{\c{c}}ek et~al.(2016){\c{C}}i{\c{c}}ek, Abdulkadir, Lienkamp,
  Brox, and Ronneberger]{Cicek:2016:U3D}
{\"O}zg{\"u}n {\c{C}}i{\c{c}}ek, Ahmed Abdulkadir, Soeren~S Lienkamp, Thomas
  Brox, and Olaf Ronneberger.
\newblock {3D U-Net}: {L}earning dense volumetric segmentation from sparse
  annotation.
\newblock In \emph{{MICCAI}}, pages 424--432, 2016.

\bibitem[Cordts et~al.(2016)Cordts, Omran, Ramos, Rehfeld, Enzweiler, Benenson,
  Franke, Roth, and Schiele]{Cordts:2016:TCD}
Marius Cordts, Mohamed Omran, Sebastian Ramos, Timo Rehfeld, Markus Enzweiler,
  Rodrigo Benenson, Uwe Franke, Stefan Roth, and Bernt Schiele.
\newblock The {Cityscapes} dataset for semantic urban scene understanding.
\newblock In \emph{CVPR}, pages 3213--3223, 2016.

\bibitem[Darcet et~al.(2025)Darcet, Baldassarre, Oquab, Mairal, and
  Bojanowski]{Darcet:2025:CPL}
Timoth{\'e}e Darcet, Federico Baldassarre, Maxime Oquab, Julien Mairal, and
  Piotr Bojanowski.
\newblock Cluster and predict latent patches for improved masked image
  modeling.
\newblock \emph{{arXiv:2502.08769 [cs.CV]}}, 2025.

\bibitem[Deng et~al.(2009)Deng, Dong, Socher, Li, Li, and
  Fei-Fei]{Deng:2009:IMN}
Jia Deng, Wei Dong, Richard Socher, Li-Jia Li, Kai Li, and Li Fei-Fei.
\newblock {ImageNet}: {A} large-scale hierarchical image database.
\newblock In \emph{CVPR}, pages 248--255, 2009.

\bibitem[Doersch et~al.(2015)Doersch, Gupta, and Efros]{Doersch:2015:UVR}
Carl Doersch, Abhinav Gupta, and Alexei~A. Efros.
\newblock Unsupervised visual representation learning by context prediction.
\newblock In \emph{ICCV}, pages 1422--1430, 2015.

\bibitem[Dosovitskiy et~al.(2021)Dosovitskiy, Beyer, Kolesnikov, Weissenborn,
  Zhai, Unterthiner, Dehghani, Minderer, Heigold, Gelly, Uszkoreit, and
  Houlsby]{Dosovitskiy:2021:AIW}
Alexey Dosovitskiy, Lucas Beyer, Alexander Kolesnikov, Dirk Weissenborn,
  Xiaohua Zhai, Thomas Unterthiner, Mostafa Dehghani, Matthias Minderer, Georg
  Heigold, Sylvain Gelly, Jakob Uszkoreit, and Neil Houlsby.
\newblock An image is worth 16$\times$16 words: {T}ransformers for image
  recognition at scale.
\newblock In \emph{ICLR}, 2021.

\bibitem[Ericsson et~al.(2022)Ericsson, Gouk, Loy, and
  Hospedales]{Ericsson:2022:SSL}
Linus Ericsson, Henry Gouk, Chen~Change Loy, and Timothy~M. Hospedales.
\newblock Self-supervised representation learning: Introduction, advances, and
  challenges.
\newblock \emph{{IEEE Trans. Signal Process.}}, 39\penalty0 (3):\penalty0
  42--62, 2022.

\bibitem[Fu et~al.(2024)Fu, Hamilton, Brandt, Feldmann, Zhang, and
  Freeman]{Fu:2024:FUP}
Stephanie Fu, Mark Hamilton, Laura~E. Brandt, Axel Feldmann, Zhoutong Zhang,
  and William~T. Freeman.
\newblock {FeatUp}: {A} model-agnostic framework for features at any
  resolution.
\newblock In \emph{ICLR}, 2024.

\bibitem[Geiger et~al.(2013)Geiger, Lenz, Stiller, and
  Urtasun]{Geiger:2013:KIT}
Andreas Geiger, Philip Lenz, Christoph Stiller, and Raquel Urtasun.
\newblock Vision meets robotics: {T}he {KITTI} dataset.
\newblock \emph{{Int. J. Robot. Res.}}, 32\penalty0 (11):\penalty0 1231--1237,
  2013.

\bibitem[Godard et~al.(2017)Godard, Mac~Aodha, and Brostow]{Godard:2017:UDE}
Cl{\'e}ment Godard, Oisin Mac~Aodha, and Gabriel~J. Brostow.
\newblock Unsupervised monocular depth estimation with left-right consistency.
\newblock In \emph{CVPR}, pages 270--279, 2017.

\bibitem[Grill et~al.()Grill, Strub, Altché, Tallec, Richemond, Buchatskaya,
  Doersch, Pires, et~al.]{Grill:2020:BYL}
Jean-Bastien Grill, Florian Strub, Florent Altché, Corentin Tallec, Pierre~H.
  Richemond, Elena Buchatskaya, Carl Doersch, Bernardo~Avila Pires, et~al.
\newblock Bootstrap your own latent: {A} new approach to self-supervised
  learning.
\newblock In \emph{NeurIPS*2020}, pages 21271--21284.

\bibitem[Gupta et~al.()Gupta, Wu, Deng, and Fei-Fei]{Gupta:2023:SMA}
Agrim Gupta, Jiajun Wu, Jia Deng, and Li Fei-Fei.
\newblock {Siamese Masked Autoencoders}.
\newblock In \emph{NeurIPS*2023}, pages 40676--40693.

\bibitem[Ha and Song(2023)]{Ha2023:SEA}
Huy Ha and Shuran Song.
\newblock Semantic abstraction: {O}pen-world {3D} scene understanding from {2D}
  vision-language models.
\newblock In \emph{{CoRL}}, pages 643--653, 2023.

\bibitem[Hahn et~al.(2024)Hahn, Araslanov, Schaub-Meyer, and
  Roth]{Hahn:2024:BUS}
Oliver Hahn, Nikita Araslanov, Simone Schaub-Meyer, and Stefan Roth.
\newblock Boosting unsupervised semantic segmentation with principal mask
  proposals.
\newblock \emph{Trans. Mach. Learn. Res.}, 2024.

\bibitem[Hahn et~al.(2025)Hahn, Reich, Araslanov, Cremers, Rupprecht, and
  Roth]{Hahn:2025:UPS}
Oliver Hahn, Christoph Reich, Nikita Araslanov, Daniel Cremers, Christian
  Rupprecht, and Stefan Roth.
\newblock Scene-centric unsupervised panoptic segmentation.
\newblock In \emph{{CVPR}}, pages 24485--24495, 2025.

\bibitem[Hamilton et~al.(2022)Hamilton, Zhang, Hariharan, Snavely, and
  Freeman]{Hamilton:2022:USS}
Mark Hamilton, Zhoutong Zhang, Bharath Hariharan, Noah Snavely, and William~T.
  Freeman.
\newblock Unsupervised semantic segmentation by distilling feature
  correspondences.
\newblock In \emph{ICLR}, 2022.

\bibitem[Han et~al.(2024)Han, Muhle, Wimbauer, and Cremers]{Han:2024:BSV}
Keonhee Han, Dominik Muhle, Felix Wimbauer, and Daniel Cremers.
\newblock Boosting self-supervision for single-view scene completion via
  knowledge distillation.
\newblock In \emph{CVPR}, pages 9837--9847, 2024.

\bibitem[Han et~al.(2019)Han, Laga, and Bennamoun]{Han:2019:S3D}
Xian-Feng Han, Hamid Laga, and Mohammed Bennamoun.
\newblock Image-based {3D} object reconstruction: {S}tate-of-the-art and trends
  in the deep learning era.
\newblock \emph{IEEE Trans. Pattern Anal. Mach. Intell.}, 43\penalty0
  (5):\penalty0 1578--1604, 2019.

\bibitem[Harb and Kn{\"o}belreiter(2021)]{Harb:2021:IFS}
Robert Harb and Patrick Kn{\"o}belreiter.
\newblock {InfoSeg}: {U}nsupervised semantic image segmentation with mutual
  information maximization.
\newblock In \emph{{GCPR}}, pages 18--32, 2021.

\bibitem[Hartley and Zisserman(2003)]{Hartley:2003:MVG}
Richard Hartley and Andrew Zisserman.
\newblock \emph{Multiple view geometry in computer vision}.
\newblock Cambridge University Press, 2003.

\bibitem[Hayler et~al.(2024)Hayler, Wimbauer, Muhle, Rupprecht, and
  Cremers]{Hayler:2024:S4C}
Adrian Hayler, Felix Wimbauer, Dominik Muhle, Christian Rupprecht, and Daniel
  Cremers.
\newblock {S4C}: {S}elf-supervised semantic scene completion with neural
  fields.
\newblock In \emph{3DV}, pages 409--420, 2024.

\bibitem[He et~al.(2020)He, Fan, Wu, Xie, and Girshick]{He:2020:MCU}
Kaiming He, Haoqi Fan, Yuxin Wu, Saining Xie, and Ross Girshick.
\newblock Momentum contrast for unsupervised visual representation learning.
\newblock In \emph{CVPR}, pages 9729--9738, 2020.

\bibitem[He et~al.(2022)He, Chen, Xie, Li, Doll{\'a}r, and
  Girshick]{He:2022:MAE}
Kaiming He, Xinlei Chen, Saining Xie, Yanghao Li, Piotr Doll{\'a}r, and Ross
  Girshick.
\newblock Masked autoencoders are scalable vision learners.
\newblock In \emph{CVPR}, pages 16000--16009, 2022.

\bibitem[Henaff(2020)]{Henaff:2020:CPP}
Olivier Henaff.
\newblock Data-efficient image pecognition with contrastive predictive coding.
\newblock In \emph{ICML}, pages 4182--4192, 2020.

\bibitem[Hjelm et~al.(2019)Hjelm, Fedorov, Lavoie-Marchildon, Grewal, Bachman,
  Trischler, and Bengio]{Hjelm:2019:LDR}
R.~Devon Hjelm, Alex Fedorov, Samuel Lavoie-Marchildon, Karan Grewal, Phil
  Bachman, Adam Trischler, and Yoshua Bengio.
\newblock Learning deep representations by mutual information estimation and
  maximization.
\newblock In \emph{ICLR}, 2019.

\bibitem[Hu et~al.(2024)Hu, Yin, Zhang, Cai, Long, Chen, Wang, Yu, Shen, and
  Shen]{Hu:2024:M3D}
Mu Hu, Wei Yin, Chi Zhang, Zhipeng Cai, Xiaoxiao Long, Hao Chen, Kaixuan Wang,
  Gang Yu, Chunhua Shen, and Shaojie Shen.
\newblock {Metric3D v2}: {A} versatile monocular geometric foundation model for
  zero-shot metric depth and surface normal estimation.
\newblock \emph{IEEE Trans. Pattern Anal. Mach. Intell.}, 46\penalty0
  (12):\penalty0 10579--10596, 2024.

\bibitem[Huang et~al.(2024{\natexlab{a}})Huang, Peng, Takmaz, Tombari,
  Pollefeys, Song, Huang, and Engelmann]{Huang:2024:S3D}
Rui Huang, Songyou Peng, Ayca Takmaz, Federico Tombari, Marc Pollefeys, Shiji
  Song, Gao Huang, and Francis Engelmann.
\newblock {Segment3D}: {L}earning fine-grained class-agnostic {3D} segmentation
  without manual labels.
\newblock In \emph{ECCV}, pages 278--295, 2024{\natexlab{a}}.

\bibitem[Huang et~al.(2023)Huang, Zheng, Zhang, Zhou, and Lu]{Huang:2023:TPV}
Yuanhui Huang, Wenzhao Zheng, Yunpeng Zhang, Jie Zhou, and Jiwen Lu.
\newblock Tri-perspective view for vision-based {3D} semantic occupancy
  prediction.
\newblock In \emph{CVPR}, pages 9223--9232, 2023.

\bibitem[Huang et~al.(2024{\natexlab{b}})Huang, Zheng, Zhang, Zhou, and
  Lu]{Huang:2024:SES}
Yuanhui Huang, Wenzhao Zheng, Borui Zhang, Jie Zhou, and Jiwen Lu.
\newblock {SelfOcc}: {S}elf-supervised vision-based {3D} occupancy prediction.
\newblock In \emph{CVPR}, pages 19946--19956, 2024{\natexlab{b}}.

\bibitem[Janai et~al.(2020)Janai, G{\"u}ney, Behl, and Geiger]{Janai:2020:AVS}
Joel Janai, Fatma G{\"u}ney, Aseem Behl, and Andreas Geiger.
\newblock Computer vision for autonomous vehicles: {P}roblems, datasets and
  state of the art.
\newblock \emph{{Found. Trends Comput. Graph. Vis.}}, 12\penalty0
  (1--3):\penalty0 1--308, 2020.

\bibitem[Ji et~al.(2019)Ji, Henriques, and Vedaldi]{Ji:2019:IIC}
Xu Ji, Joao~F. Henriques, and Andrea Vedaldi.
\newblock Invariant information clustering for unsupervised image
  classification and segmentation.
\newblock In \emph{ICCV}, pages 9865--9874, 2019.

\bibitem[Jiang et~al.(2024)Jiang, Liu, Cheng, Wang, Lin, Su, Liu, and
  Wang]{Jiang:2024:GTR}
Haoyi Jiang, Liu Liu, Tianheng Cheng, Xinjie Wang, Tianwei Lin, Zhizhong Su,
  Wenyu Liu, and Xinggang Wang.
\newblock {GaussTR}: {F}oundation model-aligned {G}aussian transformer for
  self-supervised {3D} spatial understanding.
\newblock \emph{{arXiv:2412.13193 [cs.CV]}}, 2024.

\bibitem[Kerr et~al.(2023)Kerr, Kim, Goldberg, Kanazawa, and
  Tancik]{Kerr:2023:LER}
Justin Kerr, Chung~Min Kim, Ken Goldberg, Angjoo Kanazawa, and Matthew Tancik.
\newblock {LERF}: {L}anguage embedded radiance fields.
\newblock In \emph{ICCV}, pages 19729--19739, 2023.

\bibitem[Kim et~al.(2024)Kim, Han, Ju, and Hwang]{Kim:2024:EAL}
Chanyoung Kim, Woojung Han, Dayun Ju, and Seong~Jae Hwang.
\newblock {EAGLE:} {E}igen aggregation learning for object-centric unsupervised
  semantic segmentation.
\newblock In \emph{CVPR}, pages 3523--3533, 2024.

\bibitem[Kingma and Ba(2015)]{Kingma:2015:AMS}
Diederik~P. Kingma and Jimmy~Lei Ba.
\newblock Adam: {A} method for stochastic optimization.
\newblock In \emph{ICLR}, 2015.

\bibitem[Kirillov et~al.(2023)Kirillov, Mintun, Ravi, Mao, Rolland, Gustafson,
  Xiao, Whitehead, Berg, Lo, Doll{\'a}r, and Girshick]{Kirillov:2023:SAM}
Alexander Kirillov, Eric Mintun, Nikhila Ravi, Hanzi Mao, Chloe Rolland, Laura
  Gustafson, Tete Xiao, Spencer Whitehead, Alexander~C. Berg, Wan-Yen Lo, Piotr
  Doll{\'a}r, and Ross Girshick.
\newblock {Segment Anything}.
\newblock In \emph{ICCV}, pages 4015--4026, 2023.

\bibitem[Kobayashi et~al.()Kobayashi, Matsumoto, and
  Sitzmann]{Kobayashi:2022:DNE}
Sosuke Kobayashi, Eiichi Matsumoto, and Vincent Sitzmann.
\newblock Decomposing {NeRF} for editing via feature field distillation.
\newblock In \emph{NeurIPS*2022}, pages 23311--23330.

\bibitem[Koenig et~al.(2023)Koenig, Schambach, and Otterbach]{Koenig:2023:IWS}
Alexander Koenig, Maximilian Schambach, and Johannes Otterbach.
\newblock Uncovering the inner workings of {STEGO} for safe unsupervised
  semantic segmentation.
\newblock In \emph{{CVPRW}}, pages 3789--3798, 2023.

\bibitem[Kr{\"{a}}henb{\"{u}}hl and Koltun()]{Krahenbuhl:2011:EFF}
Philipp Kr{\"{a}}henb{\"{u}}hl and Vladlen Koltun.
\newblock Efficient inference in fully connected {CRFs} with {G}aussian edge
  potentials.
\newblock In \emph{NIPS*2011}, pages 109--117.

\bibitem[Kuhn(1955)]{Kuhn:1955:THM}
Harold~W. Kuhn.
\newblock The {H}ungarian method for the assignment problem.
\newblock \emph{Nav. Res. Logist. Q.}, 2:\penalty0 83--97, 1955.

\bibitem[Li et~al.(2019)Li, Liu, Gong, Shi, Yuan, Zhao, and Reid]{Li:2019:RBD}
Jie Li, Yu Liu, Dong Gong, Qinfeng Shi, Xia Yuan, Chunxia Zhao, and Ian~D.
  Reid.
\newblock {RGBD} based dimensional decomposition residual network for {3D}
  semantic scene completion.
\newblock In \emph{CVPR}, pages 7693--7702, 2019.

\bibitem[Li et~al.(2020{\natexlab{a}})Li, Han, Wang, Liu, and
  Yuan]{Li:2020:ACN}
Jie Li, Kai Han, Peng Wang, Yu Liu, and Xia Yuan.
\newblock Anisotropic convolutional networks for {3D} semantic scene
  completion.
\newblock In \emph{CVPR}, pages 3348--3356, 2020{\natexlab{a}}.

\bibitem[Li et~al.(2020{\natexlab{b}})Li, Liu, Yuan, Zhao, Siegwart, Reid, and
  Cadena]{Li:2020:DBS}
Jie Li, Yu Liu, Xia Yuan, Chunxia Zhao, Roland Siegwart, Ian Reid, and Cesar
  Cadena.
\newblock Depth based semantic scene completion with position importance aware
  loss.
\newblock \emph{{IEEE} Robotics Autom. Lett.}, 5\penalty0 (1):\penalty0
  219--226, 2020{\natexlab{b}}.

\bibitem[Li et~al.(2021{\natexlab{a}})Li, Zhou, Xiong, and Hoi]{Li:2021:PCL}
Junnan Li, Pan Zhou, Caiming Xiong, and Steven Hoi.
\newblock Prototypical contrastive learning of unsupervised representations.
\newblock In \emph{ICLR}, 2021{\natexlab{a}}.

\bibitem[Li et~al.(2021{\natexlab{b}})Li, Shi, Liu, Zhao, Zhou, and
  Zhang]{Li:2021:SIS}
Pengfei Li, Yongliang Shi, Tianyu Liu, Hao Zhao, Guyue Zhou, and Ya{-}Qin
  Zhang.
\newblock Semi-supervised implicit scene completion from sparse {LiDAR}.
\newblock \emph{{arXiv:2111.14798 [cs.CV]}}, 2021{\natexlab{b}}.

\bibitem[Li et~al.(2023{\natexlab{a}})Li, Yu, Choy, Xiao, {\'{A}}lvarez,
  Fidler, Feng, and Anandkumar]{Li:2023:VSV}
Yiming Li, Zhiding Yu, Christopher~B. Choy, Chaowei Xiao, Jos{\'{e}}~M.
  {\'{A}}lvarez, Sanja Fidler, Chen Feng, and Anima Anandkumar.
\newblock {VoxFormer}: {S}parse voxel transformer for camera-based {3D}
  semantic scene completion.
\newblock In \emph{CVPR}, pages 9087--9098, 2023{\natexlab{a}}.

\bibitem[Li et~al.(2024)Li, Li, Liu, Gong, Li, Chen, Wang, Li, Jiang, Yu, Wang,
  Zhao, Yu, and Feng]{Li:2024:SSC}
Yiming Li, Sihang Li, Xinhao Liu, Moonjun Gong, Kenan Li, Nuo Chen, Zijun Wang,
  Zhiheng Li, Tao Jiang, Fisher Yu, Yue Wang, Hang Zhao, Zhiding Yu, and Chen
  Feng.
\newblock {SSCBench}: {A} large-scale {3D} semantic scene completion benchmark
  for autonomous driving.
\newblock In \emph{{IROS}}, pages 13333--13340, 2024.

\bibitem[Li et~al.(2023{\natexlab{b}})Li, Yu, Austin, Fang, Lan, Kautz, and
  {\'{A}}lvarez]{Li:2023:F3O}
Zhiqi Li, Zhiding Yu, David Austin, Mingsheng Fang, Shiyi Lan, Jan Kautz, and
  Jos{\'{e}}~M. {\'{A}}lvarez.
\newblock {FB-OCC:} {3D} occupancy prediction based on forward-backward view
  transformation.
\newblock \emph{arXiv:2307.01492 [cs.CV]}, 2023{\natexlab{b}}.

\bibitem[Liao et~al.(2023)Liao, Xie, and Geiger]{Liao:2023:KND}
Yiyi Liao, Jun Xie, and Andreas Geiger.
\newblock {KITTI-360:} {A} novel dataset and benchmarks for urban scene
  understanding in {2D} and {3D}.
\newblock \emph{IEEE Trans. Pattern Anal. Mach. Intell.}, 45\penalty0
  (3):\penalty0 3292--3310, 2023.

\bibitem[Lin et~al.(2024)Lin, Maire, Belongie, Hays, Perona, Ramanan,
  Doll{\'a}r, and Zitnick]{Lin:2014:MSC}
Tsung-Yi Lin, Michael Maire, Serge Belongie, James Hays, Pietro Perona, Deva
  Ramanan, Piotr Doll{\'a}r, and C.~Lawrence Zitnick.
\newblock {Microsoft COCO}: {C}ommon objects in context.
\newblock In \emph{ECCV}, pages 740--755, 2024.

\bibitem[Liu et~al.()Liu, Hu, Zeng, Tang, Jin, Han, and Li]{Liu:2028:STD}
Shice Liu, Yu Hu, Yiming Zeng, Qiankun Tang, Beibei Jin, Yinhe Han, and Xiaowei
  Li.
\newblock See and think: Disentangling semantic scene completion.
\newblock In \emph{NeurIPS*2018}, pages 261--272.

\bibitem[Lloyd(1982)]{Lloyd:1982:PCM}
Stuart Lloyd.
\newblock Least squares quantization in {PCM}.
\newblock \emph{{IEEE Trans. Inf. Theory}}, 28\penalty0 (2):\penalty0 129--137,
  1982.

\bibitem[Ma and Liu(2018)]{Ma:2018:R3D}
Zhiliang Ma and Shilong Liu.
\newblock A review of {3D} reconstruction techniques in civil engineering and
  their applications.
\newblock \emph{{Adv. Eng. Inform.}}, 37:\penalty0 163--174, 2018.

\bibitem[MacQueen(1967)]{MacQueen:1967:KME}
James MacQueen.
\newblock Some methods for classification and analysis of multivariate
  observations.
\newblock In \emph{{Berkeley Symp. on Math. Statist. and Prob.}}, pages
  281--298, 1967.

\bibitem[Max(1995)]{Max:1995:VOL}
Nelson Max.
\newblock Optical models for direct volume rendering.
\newblock \emph{{IEEE Trans. Vis. Comput. Graph.}}, 1\penalty0 (2):\penalty0
  99--108, 1995.

\bibitem[Mazur et~al.(2023)Mazur, Sucar, and Davison]{Mazur:2023:FRN}
Kirill Mazur, Edgar Sucar, and Andrew~J Davison.
\newblock Feature-realistic neural fusion for real-time, open set scene
  understanding.
\newblock In \emph{{ICRA}}, pages 8201--8207, 2023.

\bibitem[Miao et~al.(2023)Miao, Liu, Chen, Gong, Xu, Hu, and
  Zhou]{Miao:2023:ODM}
Ruihang Miao, Weizhou Liu, Mingrui Chen, Zheng Gong, Weixin Xu, Chen Hu, and
  Shuchang Zhou.
\newblock Occdepth: {A} depth-aware method for {3D} semantic scene completion.
\newblock \emph{arXiv.2302.13540 [cs.CV]}, 2023.

\bibitem[Mildenhall et~al.(2021)Mildenhall, Srinivasan, Tancik, Barron,
  Ramamoorthi, and Ng]{Mildenhall:2021:NER}
Ben Mildenhall, Pratul~P. Srinivasan, Matthew Tancik, Jonathan~T. Barron, Ravi
  Ramamoorthi, and Ren Ng.
\newblock {NeRF}: Representing scenes as neural radiance fields for view
  synthesis.
\newblock \emph{{Commun. ACM}}, 65\penalty0 (1):\penalty0 99--106, 2021.

\bibitem[Ming et~al.(2021)Ming, Meng, Fan, and Yu]{Ming:2021:MDE}
Yue Ming, Xuyang Meng, Chunxiao Fan, and Hui Yu.
\newblock Deep learning for monocular depth estimation: {A} review.
\newblock \emph{{Neurocomputing}}, 438:\penalty0 14--33, 2021.

\bibitem[Nguyen et~al.(2024)Nguyen, Li, Aggarwal, Oswald, Kirillov, Snoek, and
  Chen]{Nguyen:2024:RMA}
Duy~Kien Nguyen, Yanghao Li, Vaibhav Aggarwal, Martin~R. Oswald, Alexander
  Kirillov, Cees G.~M. Snoek, and Xinlei Chen.
\newblock R-{MAE}: Regions meet masked autoencoders.
\newblock In \emph{ICLR}, 2024.

\bibitem[Niu et~al.(2024)Niu, Wang, Han, Lian, Herzig, and
  Darrell]{Niu:2024:U2S}
Dantong Niu, Xudong Wang, Xinyang Han, Long Lian, Roei Herzig, and Trevor
  Darrell.
\newblock Unsupervised universal image segmentation.
\newblock In \emph{CVPR}, pages 22744--22754, 2024.

\bibitem[Noroozi and Favaro(2016)]{Noroozi:2016:JIG}
Mehdi Noroozi and Paolo Favaro.
\newblock Unsupervised learning of visual representations by solving jigsaw
  puzzles.
\newblock In \emph{ECCV}, pages 69--84, 2016.

\bibitem[Oquab et~al.(2024)Oquab, Darcet, Moutakanni, Vo, Szafraniec, Khalidov,
  Fernandez, Haziza, Massa, El-Nouby, et~al.]{Oquab:2023:DLR}
Maxime Oquab, Timoth{\'e}e Darcet, Th{\'e}o Moutakanni, Huy~V. Vo, Marc
  Szafraniec, Vasil Khalidov, Pierre Fernandez, Daniel Haziza, Francisco Massa,
  Alaaeldin El-Nouby, et~al.
\newblock {DINO}v2: {L}earning robust visual features without supervision.
\newblock \emph{Trans. Mach. Learn. Res.}, 2024.

\bibitem[Oswald et~al.(2013)Oswald, T{\"o}ppe, Nieuwenhuis, and
  Cremers]{Oswald:2013:SSR}
Martin~R Oswald, Eno T{\"o}ppe, Claudia Nieuwenhuis, and Daniel Cremers.
\newblock A review of geometry recovery from a single image focusing on curved
  object reconstruction.
\newblock \emph{{Innovations for Shape Analysis: Models and Algorithms}}, pages
  343--378, 2013.

\bibitem[Pan et~al.(2024)Pan, Liu, Zhang, Huang, Li, Xie, Wang, Liu, and
  Zhang]{Pan:2024:RenderOcc}
Mingjie Pan, Jiaming Liu, Renrui Zhang, Peixiang Huang, Xiaoqi Li, Hongwei Xie,
  Bing Wang, Li Liu, and Shanghang Zhang.
\newblock {RenderOcc}: {V}ision-centric {3D} occupancy prediction with {2D}
  rendering supervision.
\newblock In \emph{{ICRA}}, pages 12404--12411, 2024.

\bibitem[Peng et~al.(2023)Peng, Genova, Jiang, Tagliasacchi, Pollefeys, and
  Funkhouser]{Peng:2023:OPS}
Songyou Peng, Kyle Genova, Chiyu~``Max'' Jiang, Andrea Tagliasacchi, Marc
  Pollefeys, and Thomas Funkhouser.
\newblock {OpenScene}: {3D} scene understanding with open vocabularies.
\newblock In \emph{CVPR}, pages 815--824, 2023.

\bibitem[Ranftl et~al.(2021)Ranftl, Bochkovskiy, and Koltun]{Ranftl:2021:DPT}
Ren{\'e} Ranftl, Alexey Bochkovskiy, and Vladlen Koltun.
\newblock Vision transformers for dense prediction.
\newblock In \emph{ICCV}, pages 12179--12188, 2021.

\bibitem[Ravi et~al.(2024)Ravi, Gabeur, Hu, Hu, Ryali, Ma, Khedr, R{\"a}dle,
  Rolland, Gustafson, Mintun, Pan, et~al.]{Ravi:2024:SAM}
Nikhila Ravi, Valentin Gabeur, Yuan-Ting Hu, Ronghang Hu, Chaitanya Ryali,
  Tengyu Ma, Haitham Khedr, Roman R{\"a}dle, Chloe Rolland, Laura Gustafson,
  Eric Mintun, Junting Pan, et~al.
\newblock {SAM 2}: {S}egment anything in images and videos.
\newblock \emph{{arXiv:2408.00714 [cs.CV]}}, 2024.

\bibitem[Richter and Roth(2018)]{Richter:2018:MNP}
Stephan~R. Richter and Stefan Roth.
\newblock Matryoshka networks: {P}redicting {3D} geometry via nested shape
  layers.
\newblock In \emph{CVPR}, pages 1936--1944, 2018.

\bibitem[Rist et~al.(2022)Rist, Emmerichs, Enzweiler, and
  Gavrila]{Rist:2022:SSC}
Christoph~B. Rist, David Emmerichs, Markus Enzweiler, and Dariu~M. Gavrila.
\newblock Semantic scene completion using local deep implicit functions on
  {LiDAR} data.
\newblock \emph{IEEE Trans. Pattern Anal. Mach. Intell.}, 44\penalty0
  (10):\penalty0 7205--7218, 2022.

\bibitem[Rold{\~{a}}o et~al.(2020)Rold{\~{a}}o, de~Charette, and
  Verroust{-}Blondet]{Roldao:2020:LM3}
Luis Rold{\~{a}}o, Raoul de Charette, and Anne Verroust{-}Blondet.
\newblock {LMSCNet}: {L}ightweight multiscale {3D} semantic completion.
\newblock In \emph{3DV}, pages 111--119, 2020.

\bibitem[Roldao et~al.(2022)Roldao, De~Charette, and
  Verroust-Blondet]{Roldao:2022:SSC}
Luis Roldao, Raoul De~Charette, and Anne Verroust-Blondet.
\newblock {3D} semantic scene completion: {A} survey.
\newblock \emph{Int. J. Comput. Vis.}, 130\penalty0 (8):\penalty0 1978--2005,
  2022.

\bibitem[Schönberger and Frahm(2016)]{Schonberger:2016:SFM}
Johannes~L. Schönberger and Jan-Michael Frahm.
\newblock Structure-from-motion revisited.
\newblock In \emph{CVPR}, pages 4104--4113, 2016.

\bibitem[Sculley(2010)]{Sculley:2010:MBK}
David Sculley.
\newblock Web-scale $k$-means clustering.
\newblock In \emph{{WWW}}, page 1177–1178, 2010.

\bibitem[Seong et~al.(2023)Seong, Moon, Lee, and Heo]{Seong:2023:LHP}
Hyun~Seok Seong, WonJun Moon, SuBeen Lee, and Jae-Pil Heo.
\newblock Leveraging hidden positives for unsupervised semantic segmentation.
\newblock In \emph{CVPR}, pages 19540--19549, 2023.

\bibitem[Shafiullah et~al.(2023)Shafiullah, Paxton, Pinto, Chintala, and
  Szlam]{Shafiullah:2023:CFI}
Nur Muhammad~Mahi Shafiullah, Chris Paxton, Lerrel Pinto, Soumith Chintala, and
  Arthur Szlam.
\newblock {CLIP-Fields}: {W}eakly supervised semantic fields for robotic
  memory.
\newblock In \emph{{ICRA Workshop on Pretraining for Robotics}}, 2023.

\bibitem[Shen et~al.(2023)Shen, Yang, Yu, Wong, Kaelbling, and
  Isola]{Shen:2023:DFF}
William Shen, Ge Yang, Alan Yu, Jansen Wong, Leslie~Pack Kaelbling, and Phillip
  Isola.
\newblock Distilled feature fields enable few-shot language-guided
  manipulation.
\newblock In \emph{{CoRL}}, pages 405--424, 2023.

\bibitem[Sick et~al.(2024)Sick, Engel, Hermosilla, and Ropinski]{Sick:2024:USS}
Leon Sick, Dominik Engel, Pedro Hermosilla, and Timo Ropinski.
\newblock Unsupervised semantic segmentation through depth-guided feature
  correlation and sampling.
\newblock In \emph{CVPR}, pages 3637--3646, 2024.

\bibitem[Song et~al.(2017)Song, Yu, Zeng, Chang, Savva, and
  Funkhouser]{Song:2017:SSC}
Shuran Song, Fisher Yu, Andy Zeng, Angel~X. Chang, Manolis Savva, and Thomas~A.
  Funkhouser.
\newblock Semantic scene completion from a single depth image.
\newblock In \emph{CVPR}, pages 190--198, 2017.

\bibitem[Szymanowicz et~al.(2024)Szymanowicz, Insafutdinov, Zheng, Campbell,
  Henriques, Rupprecht, and Vedaldi]{Szymanowicz:2024:F3D}
Stanislaw Szymanowicz, Eldar Insafutdinov, Chuanxia Zheng, Dylan Campbell,
  Joao~F. Henriques, Christian Rupprecht, and Andrea Vedaldi.
\newblock {Flash3D}: {F}eed-forward generalisable {3D} scene reconstruction
  from a single image.
\newblock \emph{{arXiv:2406.04343 [cs.CV]}}, 2024.

\bibitem[Takmaz et~al.()Takmaz, Fedele, Sumner, Pollefeys, Tombari, and
  Engelmann]{Takmaz:2023:OM3}
Ay{\c{c}}a Takmaz, Elisabetta Fedele, Robert~W. Sumner, Marc Pollefeys,
  Federico Tombari, and Francis Engelmann.
\newblock {OpenMask3D}: {O}pen-vocabulary {3D} instance segmentation.
\newblock In \emph{NeurIPS*2023}, pages 68367--68390.

\bibitem[Teed and Deng(2020)]{Teed:2020:RAF}
Zachary Teed and Jia Deng.
\newblock {RAFT}: {R}ecurrent all-pairs field transforms for optical flow.
\newblock In \emph{ECCV}, pages 402--419, 2020.

\bibitem[Tong et~al.(2023)Tong, Sima, Wang, Chen, Wu, Deng, Gu, Lu, Luo, Lin,
  and Li]{Tong:2023:SAO}
Wenwen Tong, Chonghao Sima, Tai Wang, Li Chen, Silei Wu, Hanming Deng, Yi Gu,
  Lewei Lu, Ping Luo, Dahua Lin, and Hongyang Li.
\newblock Scene as occupancy.
\newblock In \emph{ICCV}, pages 8372--8381, 2023.

\bibitem[Tsagkas et~al.(2023)Tsagkas, Mac~Aodha, and Lu]{Tsagkas:2023:VLF}
Nikolaos Tsagkas, Oisin Mac~Aodha, and Chris~Xiaoxuan Lu.
\newblock {VL-Fields}: {T}owards language-grounded neural implicit spatial
  representations.
\newblock In \emph{ICRA Workshop on Representations, Abstractions, and Priors
  for Robot Learning}, 2023.

\bibitem[Tschernezki et~al.(2022)Tschernezki, Laina, Larlus, and
  Vedaldi]{Tschernezki:2022:NFF}
Vadim Tschernezki, Iro Laina, Diane Larlus, and Andrea Vedaldi.
\newblock Neural feature fusion fields: {3D} distillation of self-supervised
  {2D} image representations.
\newblock In \emph{{3DV}}, pages 443--453, 2022.

\bibitem[Tulsiani et~al.(2017)Tulsiani, Zhou, Efros, and
  Malik]{Tulsiani:2017:CVPR}
Shubham Tulsiani, Tinghui Zhou, Alexei~A. Efros, and Jitendra Malik.
\newblock Multi-view supervision for single-view reconstruction via
  differentiable ray consistency.
\newblock In \emph{CVPR}, 2017.

\bibitem[Wang et~al.(2023)Wang, Girdhar, Yu, and Misra]{Wang:2023:CAL}
Xudong Wang, Rohit Girdhar, Stella~X. Yu, and Ishan Misra.
\newblock Cut and learn for unsupervised object detection and instance
  segmentation.
\newblock In \emph{CVPR}, pages 3124--3134, 2023.

\bibitem[Wang et~al.(2004)Wang, Bovik, Sheikh, and Simoncelli]{Wang:2004:SSI}
Zhou Wang, Alan~C. Bovik, Hamid~R. Sheikh, and Eero~P. Simoncelli.
\newblock Image quality assessment: {F}rom error visibility to structural
  similarity.
\newblock \emph{IEEE Trans. Image Process.}, 13\penalty0 (4):\penalty0
  600--612, 2004.

\bibitem[Weder et~al.(2024)Weder, Blum, Engelmann, and
  Pollefeys]{Weder:2024:LAM}
Silvan Weder, Hermann Blum, Francis Engelmann, and Marc Pollefeys.
\newblock {LabelMaker}: Automatic semantic label generation from {RGB-D}
  trajectories.
\newblock In \emph{{3DV}}, pages 334--343, 2024.

\bibitem[Wei et~al.(2022)Wei, Fan, Xie, Wu, Yuille, and
  Feichtenhofer]{Wei:2022:MFP}
Chen Wei, Haoqi Fan, Saining Xie, Chao-Yuan Wu, Alan Yuille, and Christoph
  Feichtenhofer.
\newblock Masked feature prediction for self-supervised visual pre-training.
\newblock In \emph{CVPR}, pages 14668--14678, 2022.

\bibitem[Wimbauer et~al.(2023)Wimbauer, Yang, Rupprecht, and
  Cremers]{Wimbauer:2023:BTS}
Felix Wimbauer, Nan Yang, Christian Rupprecht, and Daniel Cremers.
\newblock Behind the scenes: {D}ensity fields for single view reconstruction.
\newblock In \emph{CVPR}, pages 9076--9086, 2023.

\bibitem[Xie et~al.(2022)Xie, Takikawa, Saito, Litany, Yan, Khan, Tombari,
  Tompkin, Sitzmann, and Sridhar]{Xie:2022:NEF}
Yiheng Xie, Towaki Takikawa, Shunsuke Saito, Or Litany, Shiqin Yan, Numair
  Khan, Federico Tombari, James Tompkin, Vincent Sitzmann, and Srinath Sridhar.
\newblock Neural fields in visual computing and beyond.
\newblock In \emph{{Comput. Graph. Forum}}, pages 641--676, 2022.

\bibitem[Yan et~al.(2021)Yan, Gao, Li, Zhang, Li, Huang, and Cui]{Yan:2021:SSS}
Xu Yan, Jiantao Gao, Jie Li, Ruimao Zhang, Zhen Li, Rui Huang, and Shuguang
  Cui.
\newblock Sparse single sweep {LiDAR} point cloud segmentation via learning
  contextual shape priors from scene completion.
\newblock In \emph{AAAI}, pages 3101--3109, 2021.

\bibitem[Yang et~al.(2024{\natexlab{a}})Yang, Ivanovic, Litany, Weng, Kim, Li,
  Che, Xu, Fidler, Pavone, and Wang]{Yang:2024:EME}
Jiawei Yang, Boris Ivanovic, Or Litany, Xinshuo Weng, Seung~Wook Kim, Boyi Li,
  Tong Che, Danfei Xu, Sanja Fidler, Marco Pavone, and Yue Wang.
\newblock Emerne{RF}: {E}mergent spatial-temporal scene decomposition via
  self-supervision.
\newblock In \emph{ICLR}, 2024{\natexlab{a}}.

\bibitem[Yang et~al.(2024{\natexlab{b}})Yang, Luo, Li, Deng, Guibas, Krishnan,
  Weinberger, Tian, and Wang]{Yang:2024:DEN}
Jiawei Yang, Katie~Z Luo, Jiefeng Li, Congyue Deng, Leonidas Guibas, Dilip
  Krishnan, Kilian~Q Weinberger, Yonglong Tian, and Yue Wang.
\newblock Denoising vision transformers.
\newblock In \emph{ECCV}, pages 453--469, 2024{\natexlab{b}}.

\bibitem[Yang et~al.(2024{\natexlab{c}})Yang, Dai, and Pan]{Yang:2024:M3D}
Zhuoyue Yang, Ju Dai, and Junjun Pan.
\newblock {3D} reconstruction from endoscopy images: {A} survey.
\newblock \emph{{Comput. Biol. Med.}}, 175:\penalty0 108546,
  2024{\natexlab{c}}.

\bibitem[Yu et~al.(2021)Yu, Ye, Tancik, and Kanazawa]{Yu:2021:PNE}
Alex Yu, Vickie Ye, Matthew Tancik, and Angjoo Kanazawa.
\newblock {pixelNeRF}: {N}eural radiance fields from one or few images.
\newblock In \emph{CVPR}, pages 4578--4587, 2021.

\bibitem[Yu et~al.(2020)Yu, Chen, Wang, Xian, Chen, Liu, Madhavan, and
  Darrell]{Yu:2020:BDD}
Fisher Yu, Haofeng Chen, Xin Wang, Wenqi Xian, Yingying Chen, Fangchen Liu,
  Vashisht Madhavan, and Trevor Darrell.
\newblock {BDD100K}: {A} diverse driving dataset for heterogeneous multitask
  learning.
\newblock In \emph{CVPR}, pages 2633--2642, 2020.

\bibitem[Yue et~al.(2024)Yue, Das, Engelmann, Tang, and Lenssen]{Yue:2024:FIT}
Yuanwen Yue, Anurag Das, Francis Engelmann, Siyu Tang, and Jan~Eric Lenssen.
\newblock Improving {2D} feature representations by {3D}-aware fine-tuning.
\newblock In \emph{ECCV}, pages 57--74, 2024.

\bibitem[Zhang et~al.(2019)Zhang, Liu, Lei, Lu, and Yang]{Zhang:2019:CCP}
Pingping Zhang, Wei Liu, Yinjie Lei, Huchuan Lu, and Xiaoyun Yang.
\newblock Cascaded context pyramid for full-resolution {3D} semantic scene
  completion.
\newblock In \emph{ICCV}, pages 7800--7809, 2019.

\bibitem[Zhang et~al.(2023)Zhang, Zhu, and Du]{Zhang:2023:OCC}
Yunpeng Zhang, Zheng Zhu, and Dalong Du.
\newblock {OccFormer}: {D}ual-path transformer for vision-based {3D} semantic
  occupancy prediction.
\newblock In \emph{ICCV}, pages 9433--9443, 2023.

\bibitem[Zhou et~al.(2018)Zhou, Tucker, Flynn, Fyffe, and
  Snavely]{Zhou:2018:SML}
Tinghui Zhou, Richard Tucker, John Flynn, Graham Fyffe, and Noah Snavely.
\newblock Stereo magnification: {L}earning view synthesis using multiplane
  images.
\newblock \emph{{ACM Trans. Graph.}}, 37\penalty0 (4):\penalty0 65, 2018.

\bibitem[Özyeşil et~al.(2017)Özyeşil, Voroninski, Basri, and
  Singer]{Ozyesil:2017:SFM}
Onur Özyeşil, Vladislav Voroninski, Ronen Basri, and Amit Singer.
\newblock A survey of structure from motion.
\newblock \emph{{Acta Numer.}}, 26:\penalty0 305–364, 2017.

\end{thebibliography}


{
\small
\begin{thebibliography}{114}

\bibitem[121]{slam}Lucas R. Agostinho, Nuno M. Ricardo, Maria I. Pereira, Pinto Antoine, and Andry M. Pinto. A practical survey on visual odometry for autonomous driving in challenging scenarios and conditions. \textit{IEEE Access}, 10:72182-72205, 2022.

\bibitem[122]{soft2}Igor Cvišić, Ivan Marković, and Ivan Petrović. SOFT2: Stereo visual odometry for road vehicles based on a point-to-epipolar-line metric. \textit{IEEE Trans. Robot.}, 39(1):273-288, 2023.

\bibitem[123]{pytorch} Adam Paszke, Sam Gross, Francisco Massa, Adam Lerer, James Bradbury, Gregory Chanan, Trevor Killeen, Zeming Lin, Natalia Gimelshein, Luca Antiga, Alban Desmaison, Andreas K{\"{o}}pf, Edward Z. Yang, Zachary DeVito, Martin Raison, Alykhan Tejani, Sasank Chilamkurthy, Benoit Steiner, Lu Fang, Junjie Bai, and Soumith Chintala. PyTorch: An imperative style, high-performance deep learning library. In \textit{NeurIPS*2019}, pages 8024--8035.

\bibitem[124]{dynamodepth}Yihong Sun and Bharath Hariharan. Dynamo-Depth: Fixing unsupervised depth estimation for dynamical scenes. In \textit{NeurIPS*2023}, pages 54987--55005.

\bibitem[125]{occ}Narayanan Sundaram, Thomas Brox, and Kurt Keutzer. Dense point trajectories by GPU-accelerated large displacement optical flow. In \textit{ECCV}, pages 438–-451, 2010.

\bibitem[126]{prodepth}Sungmin Woo, Wonjoon Lee, Woo Woo Jin, Dogyoon Lee, and Sangyoun Lee. ProDepth: Boosting self-supervised multi-frame monocular depth with probabilistic fusion. In \textit{ECCV}, pages 201--217, 2024.

\bibitem[127]{ade20k}Bolei Zhou, Hang Zhao, Xavier Puig, Sanja Fidler, Adela Barriuso, and Antonio Torralba. Scene parsing through ADE20K dataset. In \textit{CVPR}, pages 5122--5130, 2017.

\end{thebibliography}
}
}

\clearpage
\setcounter{section}{0}
\renewcommand\thesection{\Alph{section}}
\setcounter{page}{1}
\pagenumbering{roman}
\twocolumn[{%
\renewcommand\twocolumn[1][]{#1}%
\maketitlesupplementary
{\large Aleksandar Jevti\'{c}\textsuperscript{\normalfont{}* 1}
\authorstep Christoph Reich\textsuperscript{\normalfont{}* 1,2,4,5}
\authorstep Felix Wimbauer\textsuperscript{\normalfont{} 1,4}\\[0pt]
Oliver Hahn\textsuperscript{\normalfont{} 2}
\authorstep Christian Rupprecht\textsuperscript{\normalfont{} 3}
\authorstep Stefan Roth\textsuperscript{\normalfont{} 2,5,6}
\authorstep Daniel Cremers\textsuperscript{\normalfont{} 1,4,5}\\[4pt]
\small{\textsuperscript{1}TU Munich \affiliationstep \textsuperscript{2}TU Darmstadt \affiliationstep \textsuperscript{3}University of Oxford \affiliationstep \textsuperscript{4}MCML\affiliationstep \textsuperscript{5}ELIZA\affiliationstep \textsuperscript{6}hessian.AI\affiliationstep
\textsuperscript{*}equal contribution}\\[-5.5pt]\small {\url{https://visinf.github.io/scenedino}}}
\vspace{0.75cm}
}]

In this appendix, we provide further implementation details, including dataset properties and an overview of \MethodName's computational complexity (\cf \cref{supp:reproducibility}). We discuss our multi-view feature consistency evaluation approach (\cf \cref{supp:mvfc}). Next, we provide additional qualitative and quantitative results (\cf \cref{supp:results}), including failure cases. Finally, we discuss the limitations of \MethodName and suggest future research directions (\cf \cref{supp:limitations}).

\section{Reproducibility\label{supp:reproducibility}}

Here, we provide further implementation details, information about the utilized dataset, and computational complexity details to ensure reproducibility. Note that our code is available at\, \url{https://github.com/tum-vision/scenedino}. 

\subsection{Implementation details}
We implement \MethodName in PyTorch~\cite{pytorch} and build on the code of BTS~\cite{Wimbauer:2023:BTS}, STEGO~\cite{Hamilton:2022:USS}, and S4C~\cite{Hayler:2024:S4C}. Our encoder-decoder (pre-trained DINO-B/8 and randomly initialized dense prediction decoder) produces per-pixel embeddings of dimensionality ${\rm D}_{\mathbf{E}}=256$. Based on these embeddings, the two-layer MLP $\phi$ (hidden dimensionality \num{128}) predicts \num{64} features. As rendering features is expensive, requiring multiple forward passes through the MLP, $\phi$ predicts \num{64} features. We employ another MLP to up-project again to the full dimensionality ${\rm D}=768$; this MLP is learned with \MethodName and can up-project both 3D features and 2D rendered features. We train for \SI{100}{k} steps with a base learning rate of $10^{-4}$, dropping to $10^{-5}$ after \SI{50}{k} steps. We train using a batch size of \num{4}, extracting \num{32} patches of size $8 \times 8$ per image. These patches align with the per-patch DINO target features. For our feature field loss formulation (\cf \cref{subsec:training}), we use the loss weights $\lambda_{\text{p}}=1, \lambda_{\text{s}}=0.001, \lambda_{\text{f}}=0.2, \lambda_{\text{fs}}=0.25$.

The MLP head $h$ (hidden dimensionality $768$) produces \num{64} distilled features $\rm K$. We perform distillation for \num{1000} steps with a learning rate of $5 \cdot 10^{-4}$. We train using a batch size of \num{4}, \num{5} center points, a feature batch of size \num{576}, and cluster with ${\rm C}=19$. For $k$NN sampling, we use $k=4$. The feature buffer holds \num{256} feature batches. The loss term in \cref{eq:distill_loss} is parameterized with $\lambda_{\text{self}}=0.08$, $\lambda_{k\text{NN}}=0.43$, and  $\lambda_{\text{rand}}=0.67$. For the similarity thresholds, we use $b_{\text{self}}=0.44$, $b_{k\text{NN}}=0.18$, and $b_{\text{rand}}=0.87$.

We follow standard practice in 2D unsupervised semantic segmentation \cite{Cho:2021:PUS, Hamilton:2022:USS, Seong:2023:LHP, Kim:2024:EAL, Sick:2024:USS, Hahn:2024:BUS, Niu:2024:U2S} by applying Hungarian matching~\cite{Kuhn:1955:THM} to align our pseudo semantics. For SSC validation, we map down to 15 semantic classes while following existing work \cite{Hamilton:2022:USS, Hahn:2024:BUS} for 2D validation and map to 19 semantic classes.

\begin{figure*}[t]
    \centering
    \newcommand{\imgwidth}{0.27}
\newcommand{\dddviswidth}{0.182}

\scriptsize
\sffamily
\setlength{\tabcolsep}{0pt}
\renewcommand{\arraystretch}{0.66}
\begin{tabular}{>{\centering\arraybackslash} m{\imgwidth\textwidth} 
                >{\centering\arraybackslash} m{\dddviswidth\textwidth} 
                >{\centering\arraybackslash} m{\dddviswidth\textwidth} 
                >{\centering\arraybackslash} m{\dddviswidth\textwidth}
                >{\centering\arraybackslash} m{\dddviswidth\textwidth}}

\multirow{2}{*}{\vspace{-0.5em}\textbf{Input Image}} & \multicolumn{2}{c}{\textbf{\MethodName \textit{(Ours)}}} & \textbf{S4C + STEGO} & \multirow{2}{*}{\vspace{-0.5em}\textbf{Ground Truth}} \\
\cmidrule(l{0.5em}r{0.5em}){2-3} \cmidrule(l{0.5em}r{0.5em}){4-4}
& Feature Field & SSC Prediction & SSC Prediction & \\[-1pt]

\includegraphics[width=\linewidth]{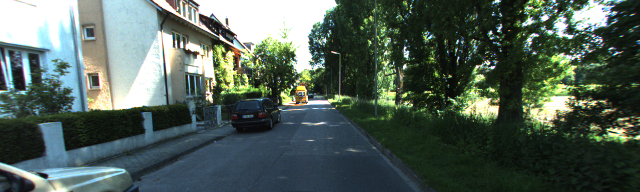} &
\includegraphics[width=\linewidth]{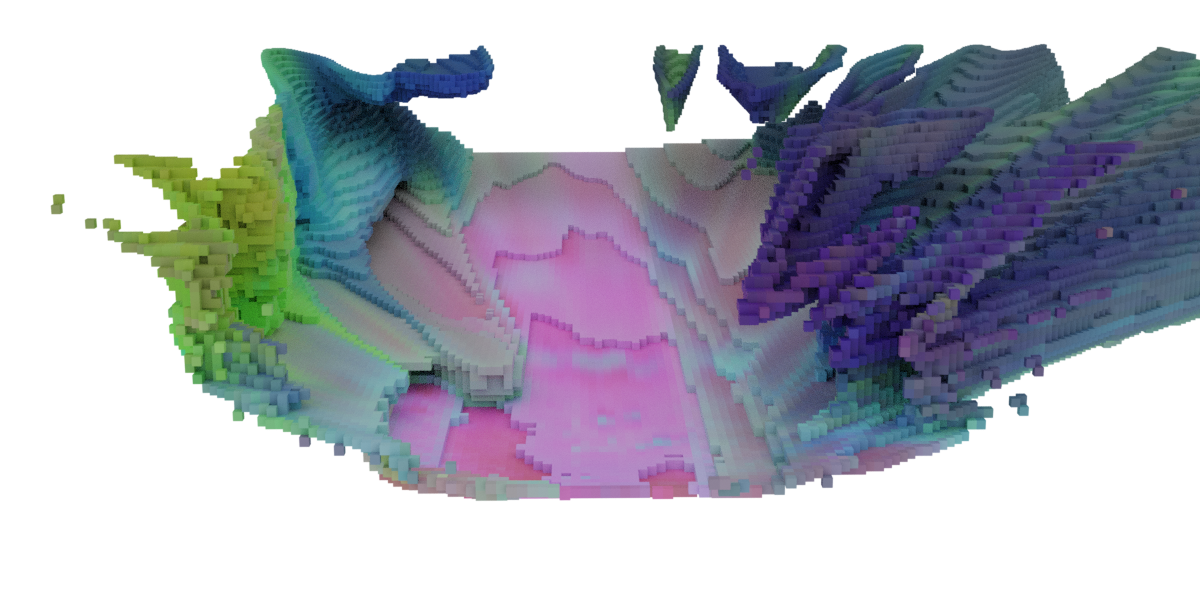} &
\includegraphics[width=\linewidth]{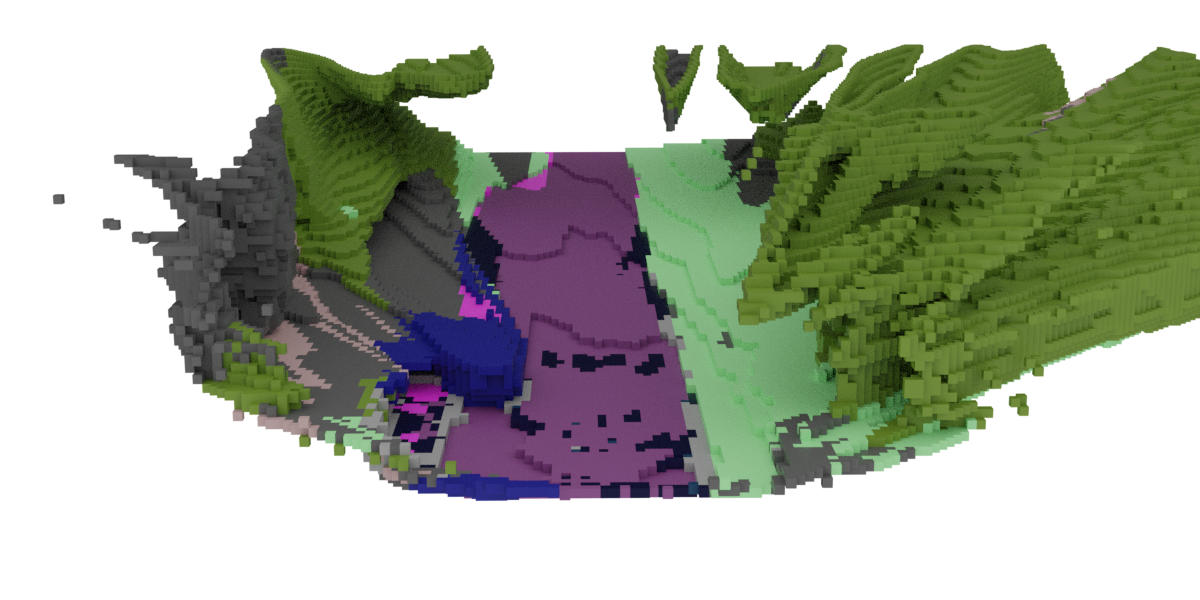} &
\includegraphics[width=\linewidth]{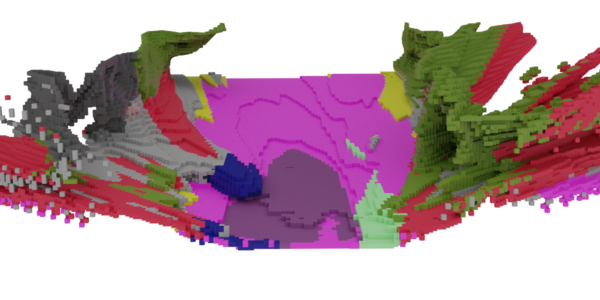} &
\includegraphics[width=\linewidth]{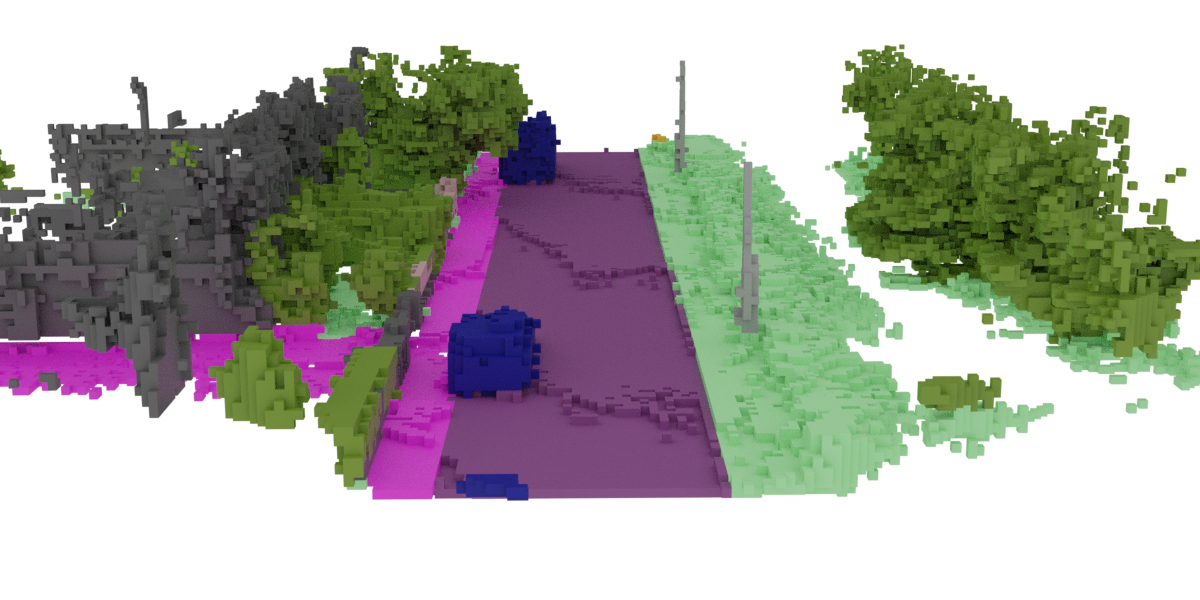} \\[-6.5pt]

\includegraphics[width=\linewidth]{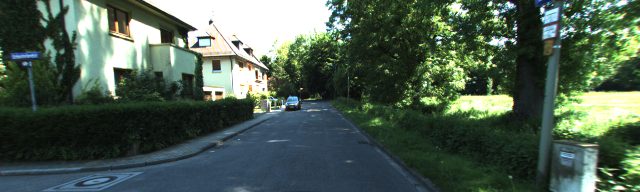} &
\includegraphics[width=\linewidth]{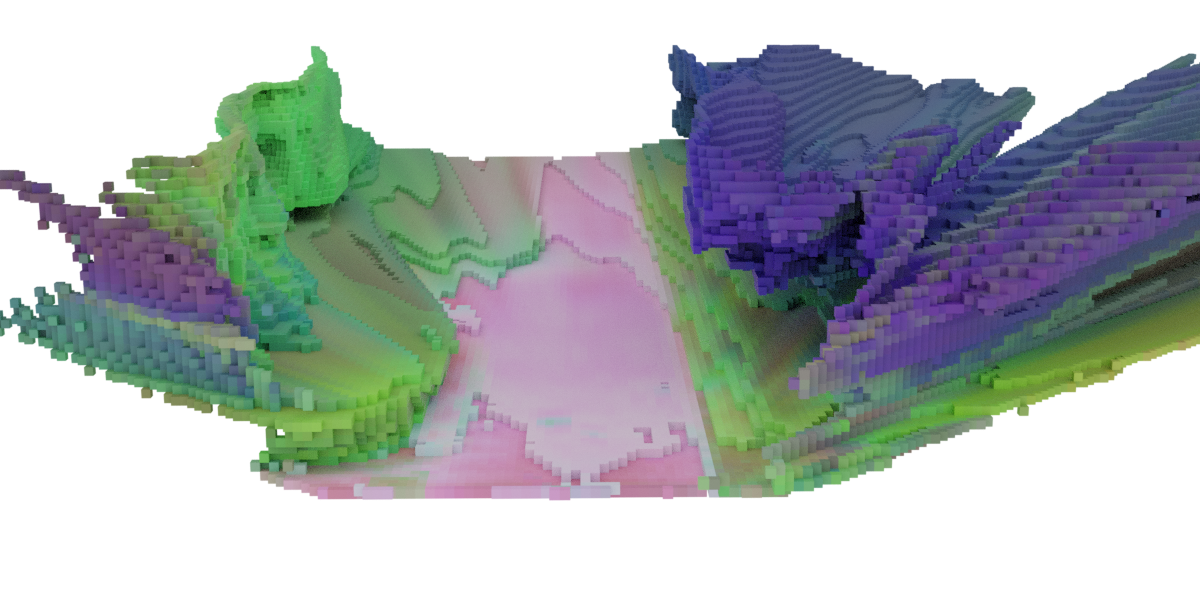} &
\includegraphics[width=\linewidth]{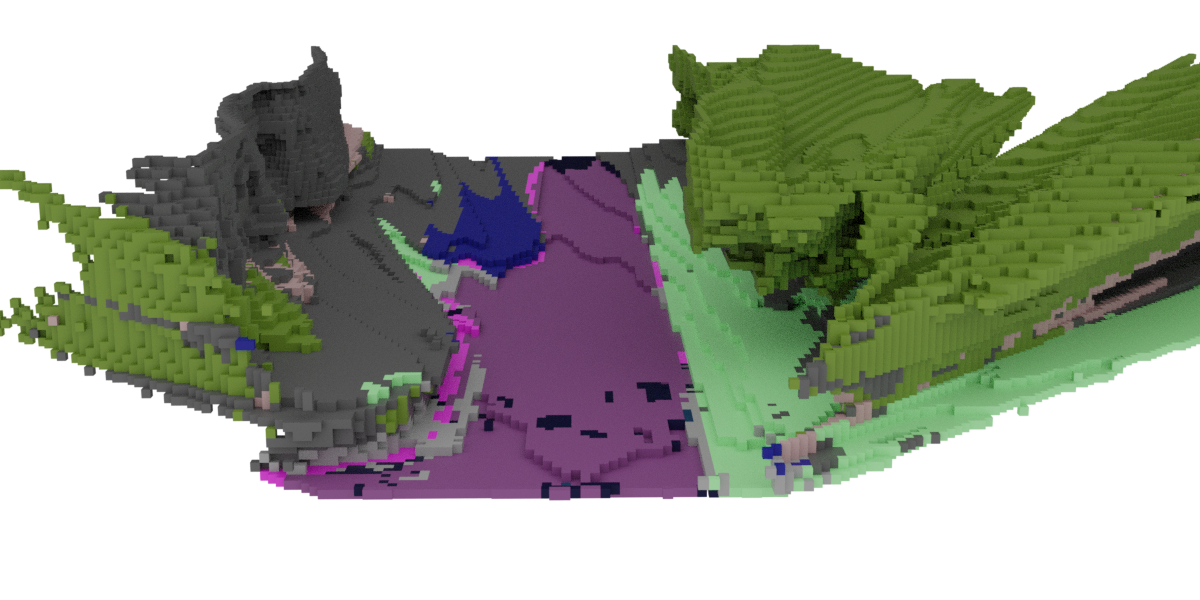} &
\includegraphics[width=\linewidth]{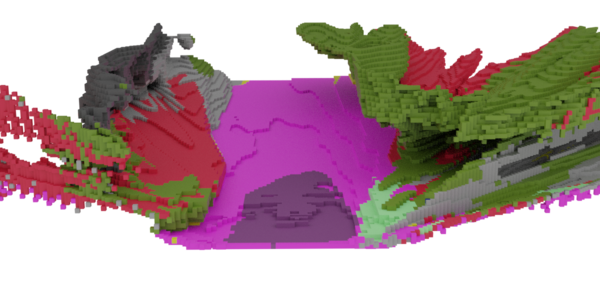} &
\includegraphics[width=\linewidth]{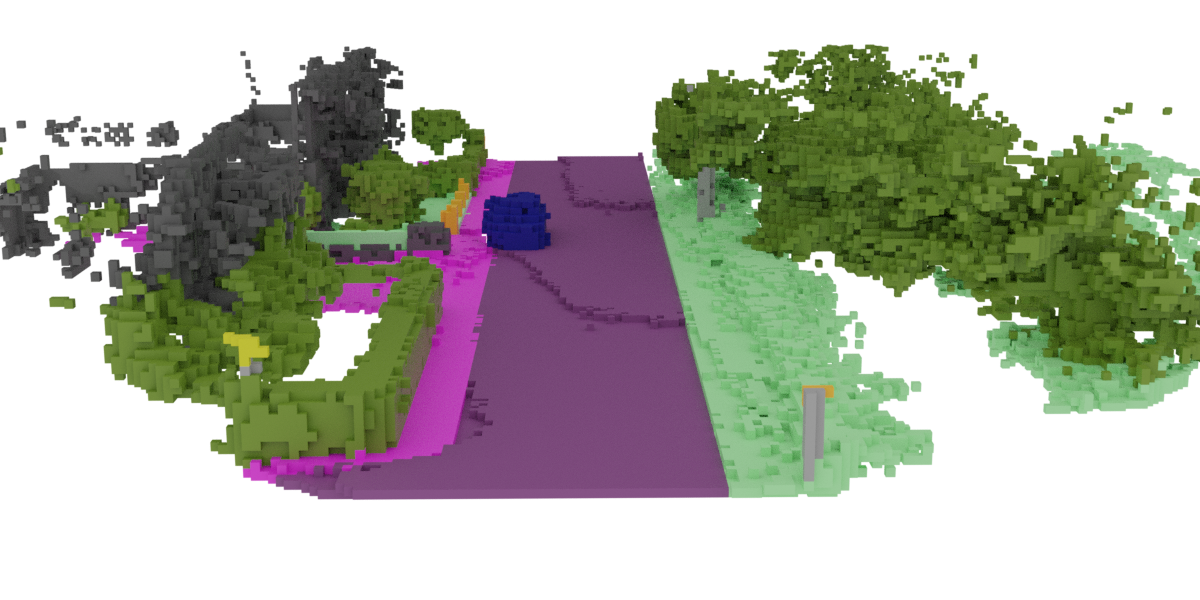} \\[-6.5pt]

\includegraphics[width=\linewidth]{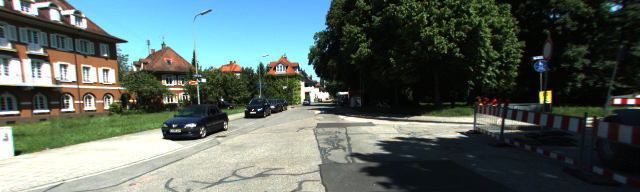} &
\includegraphics[width=\linewidth]{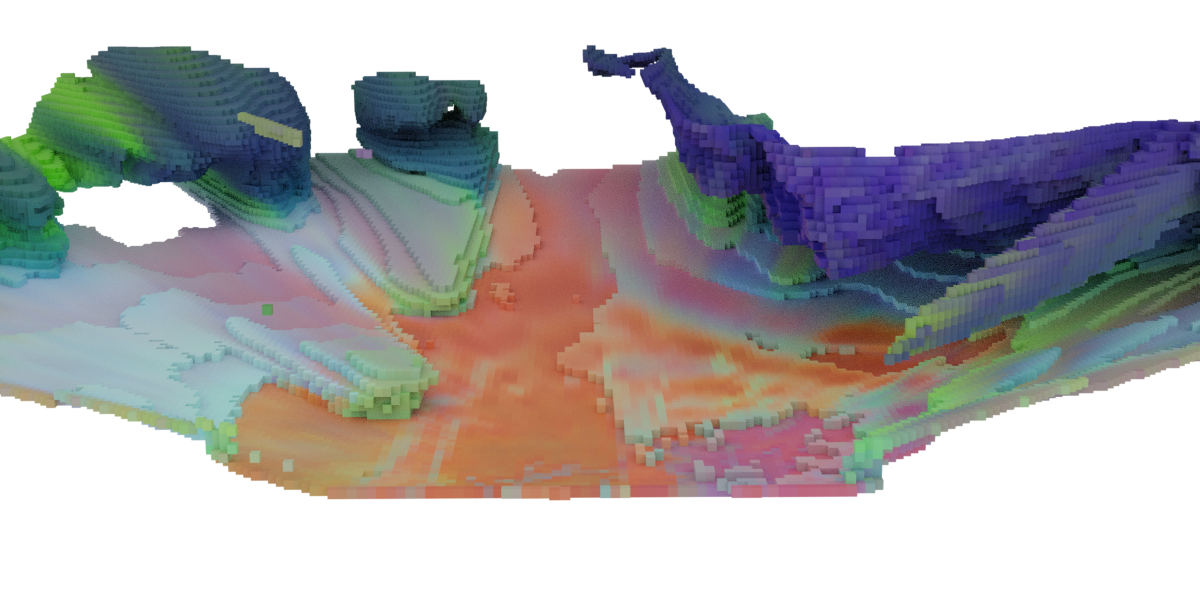} &
\includegraphics[width=\linewidth]{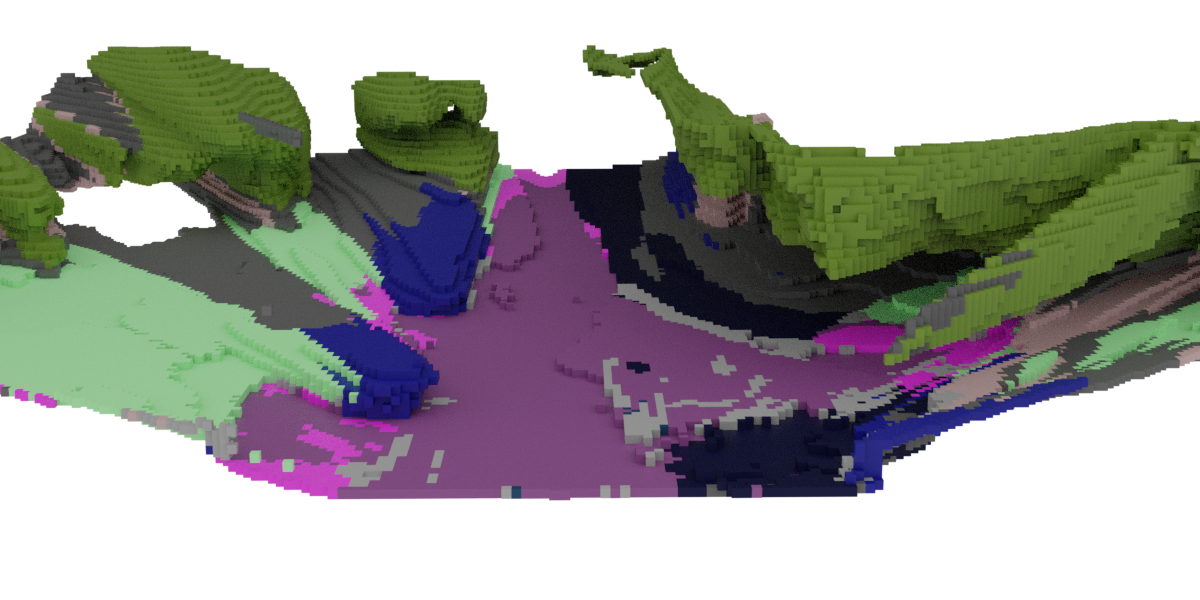} &
\includegraphics[width=\linewidth]{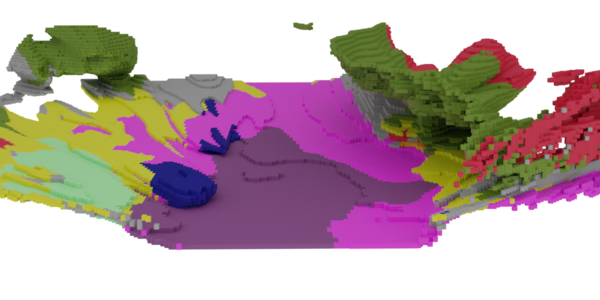} &
\includegraphics[width=\linewidth]{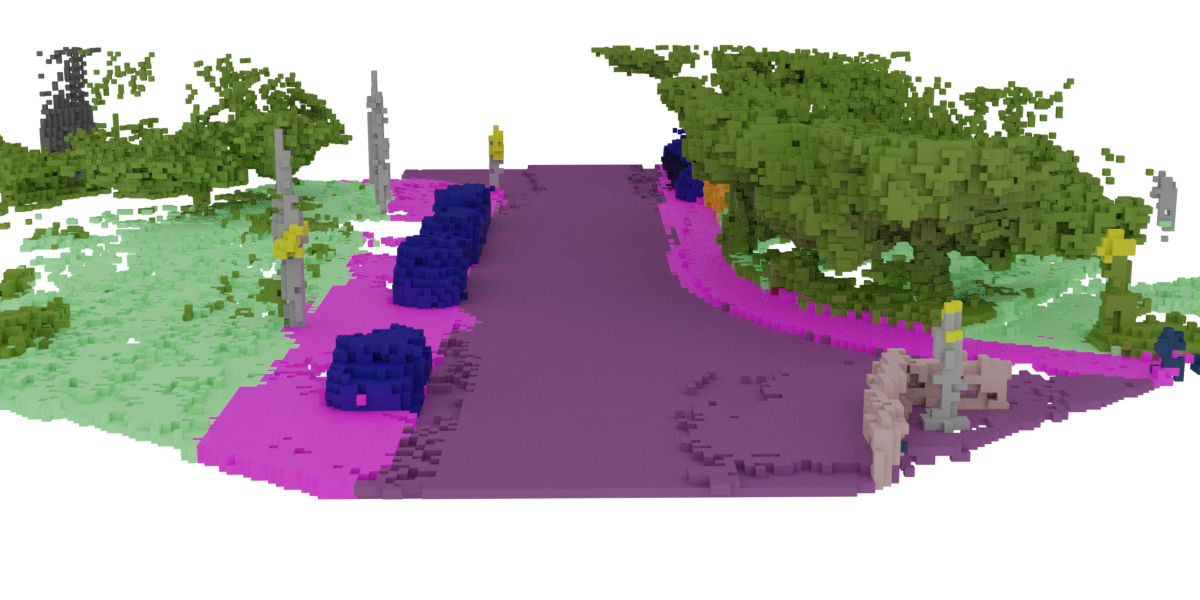} \\[-6.5pt]

\includegraphics[width=\linewidth]{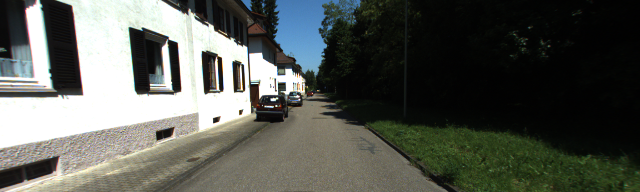} &
\includegraphics[width=\linewidth]{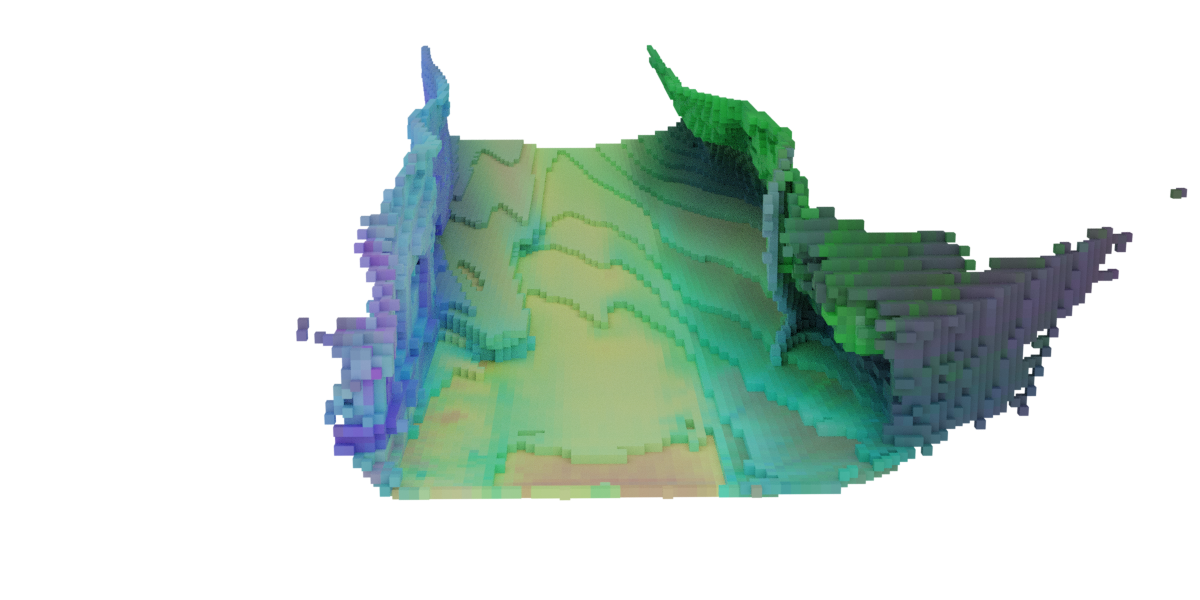} &
\includegraphics[width=\linewidth]{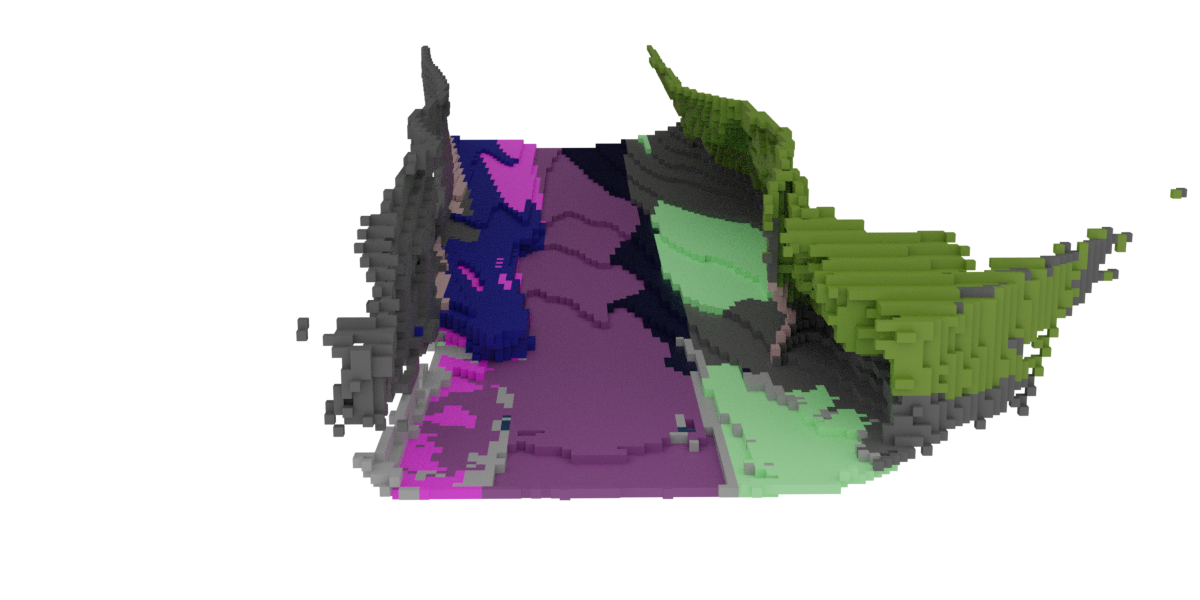} &
\includegraphics[width=\linewidth]{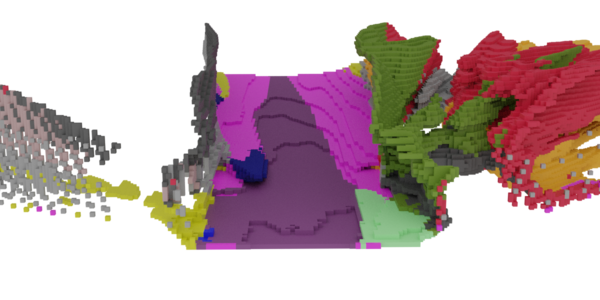} &
\includegraphics[width=\linewidth]{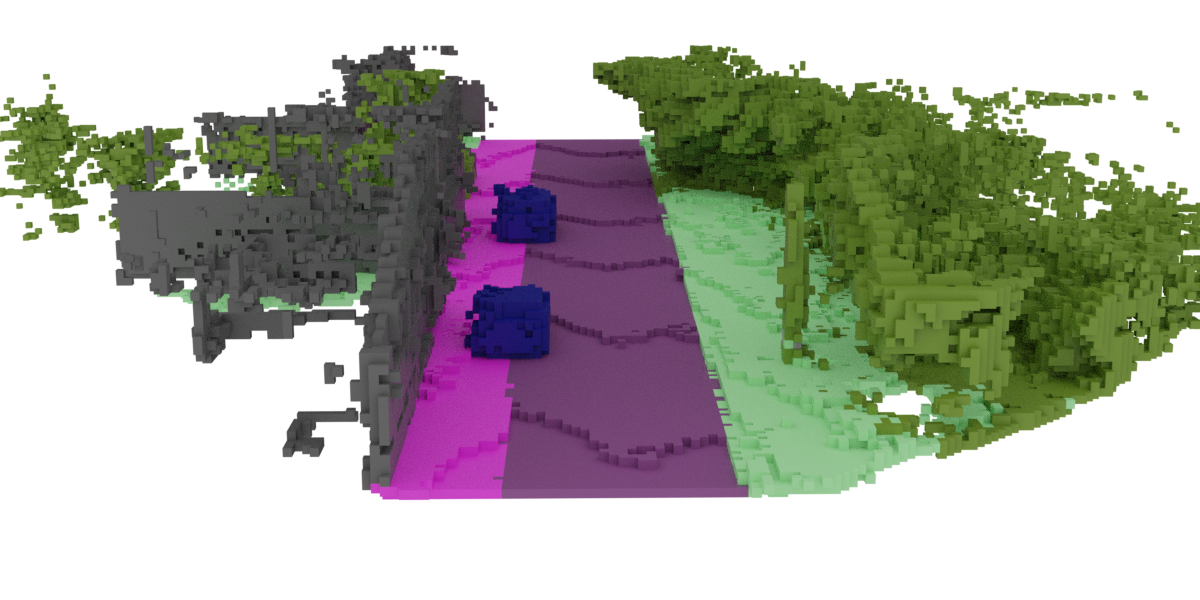} \\[-6.5pt]

\includegraphics[width=\linewidth]{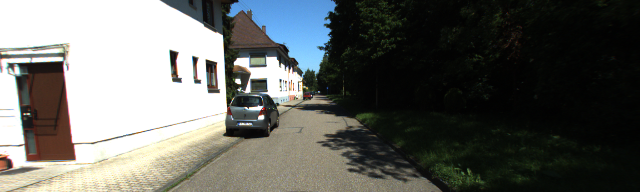} &
\includegraphics[width=\linewidth]{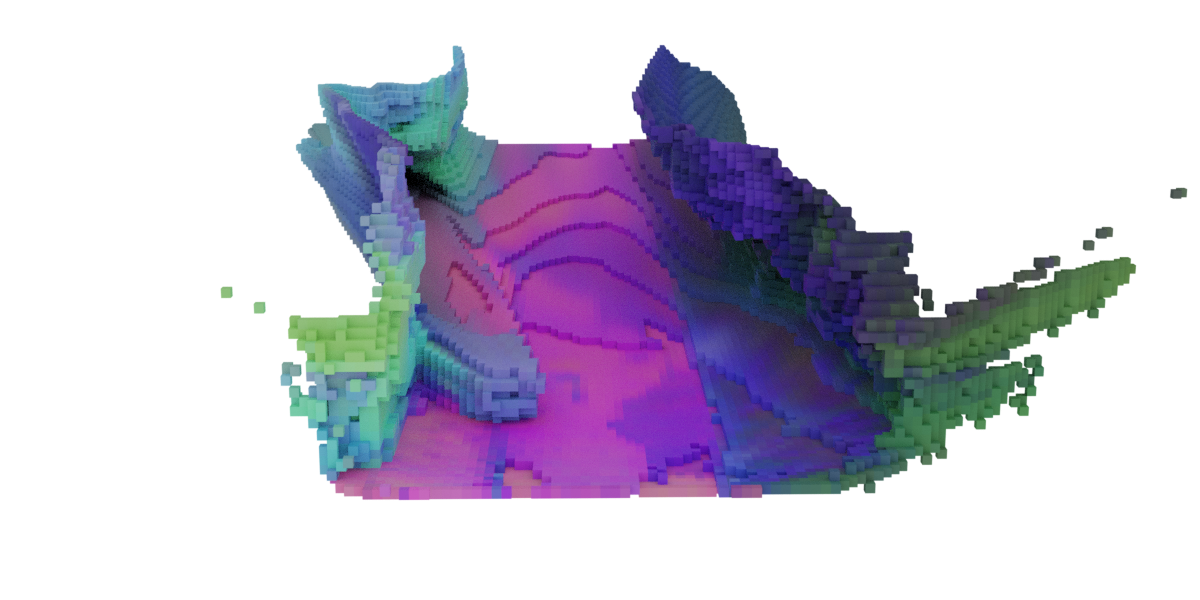} &
\includegraphics[width=\linewidth]{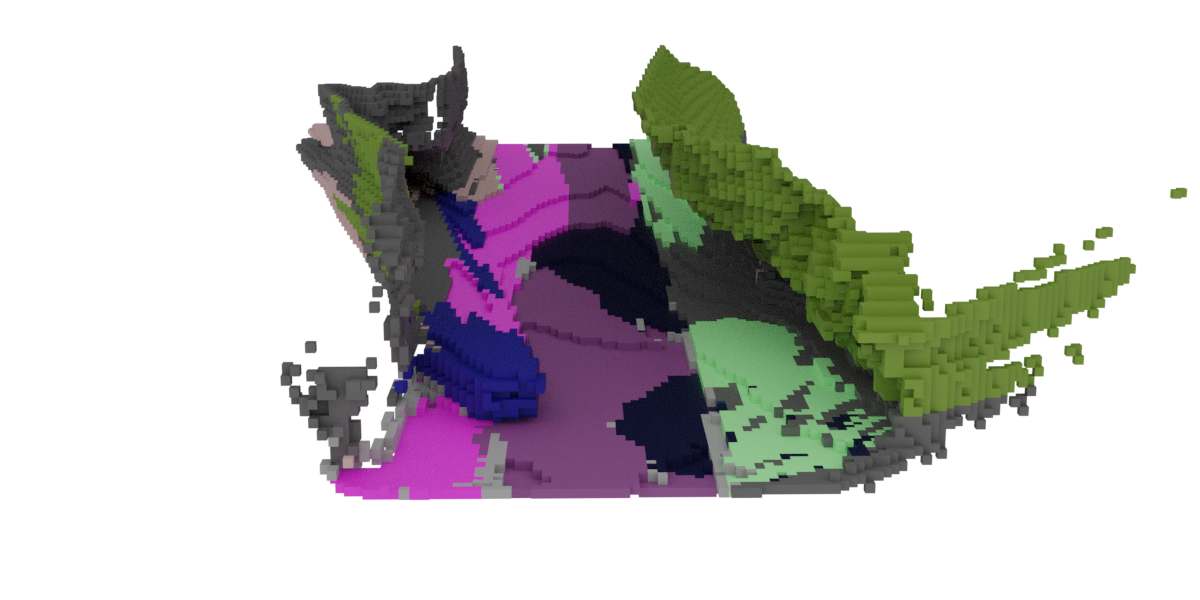} &
\includegraphics[width=\linewidth]{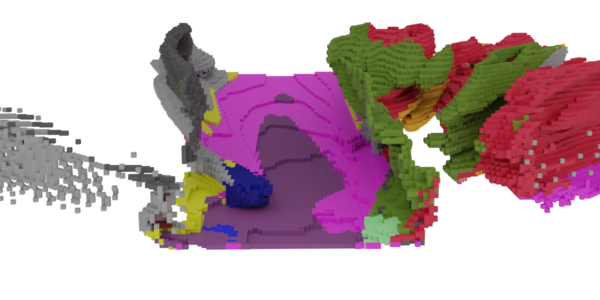} &
\includegraphics[width=\linewidth]{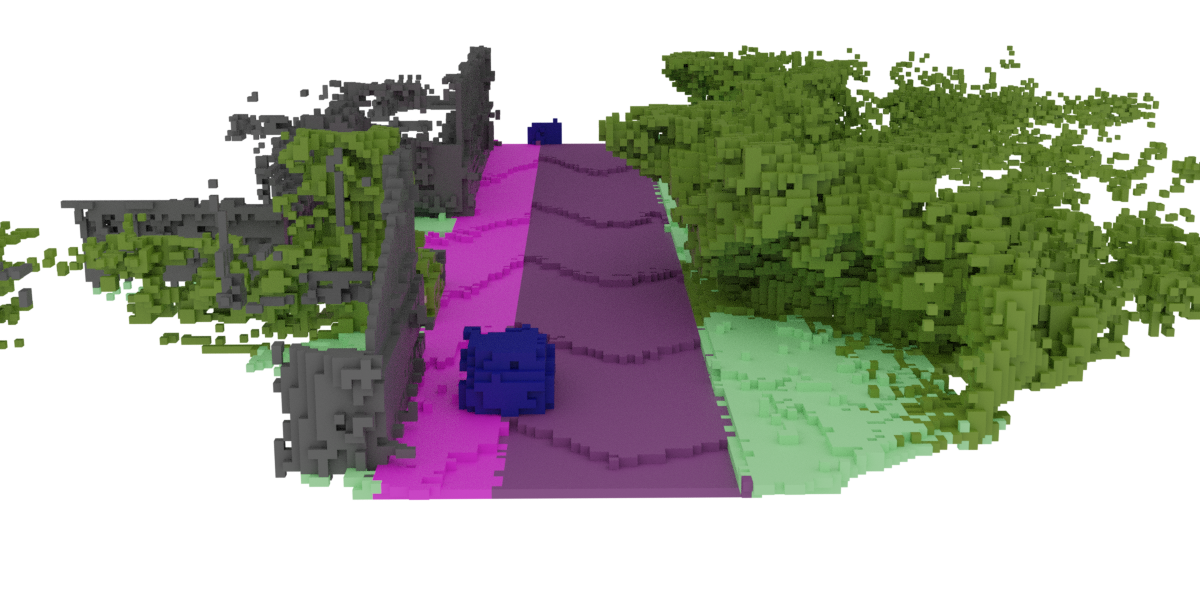} \\[-6.5pt]

\includegraphics[width=\linewidth]{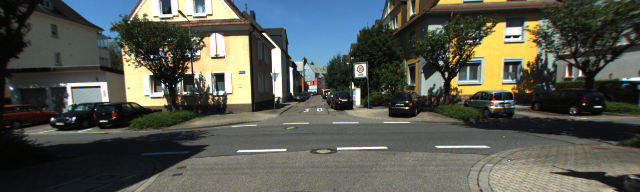} &
\includegraphics[width=\linewidth]{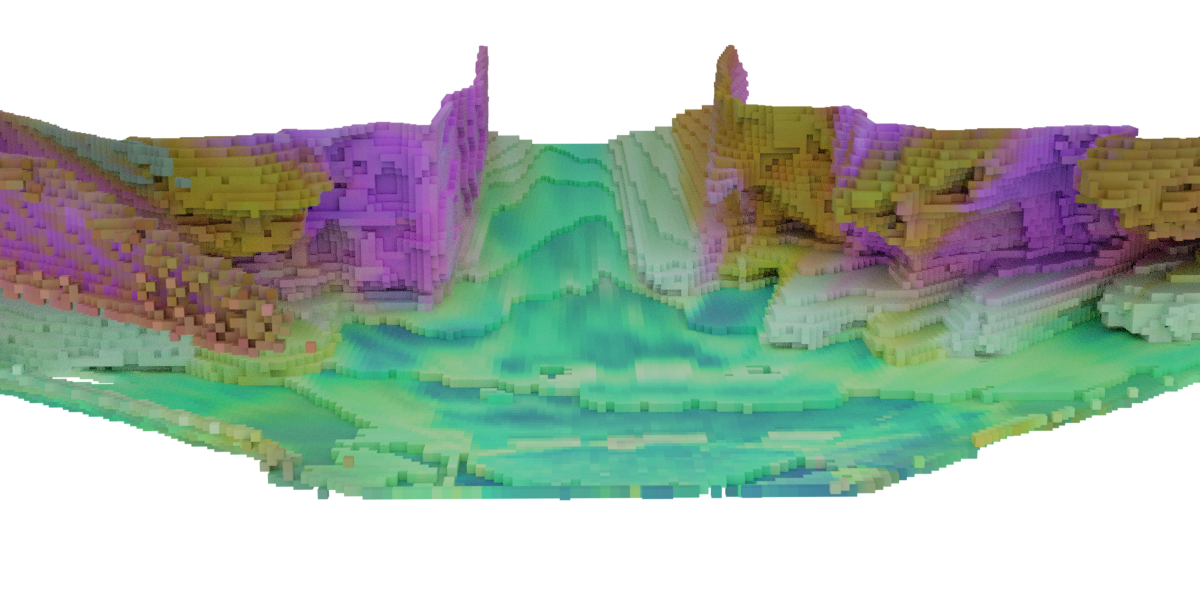} &
\includegraphics[width=\linewidth]{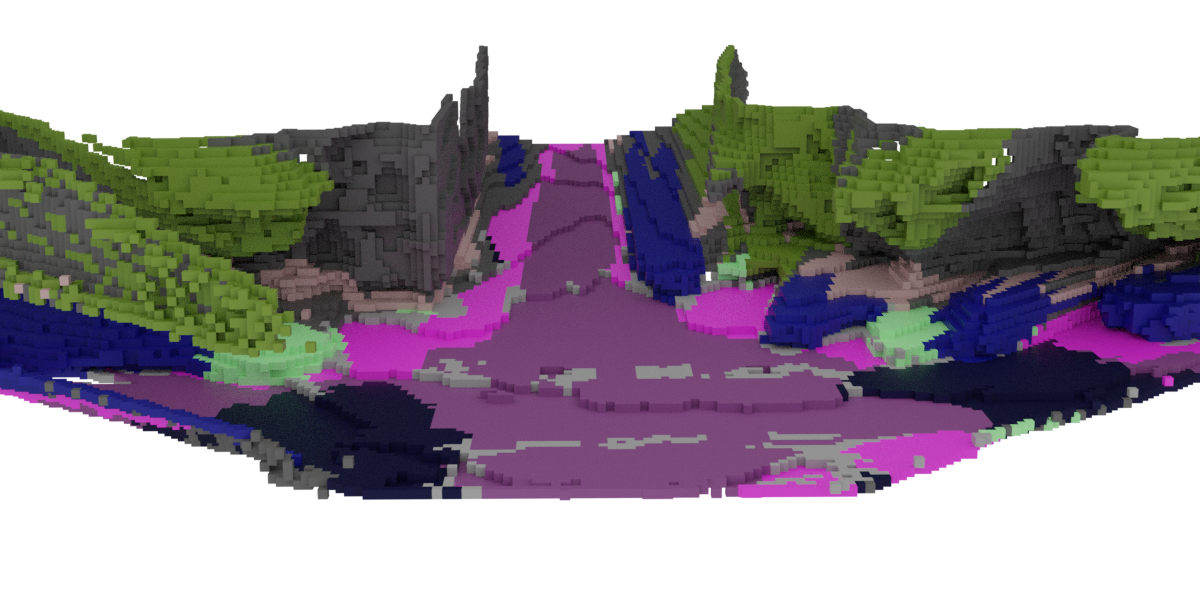} &
\includegraphics[width=\linewidth]{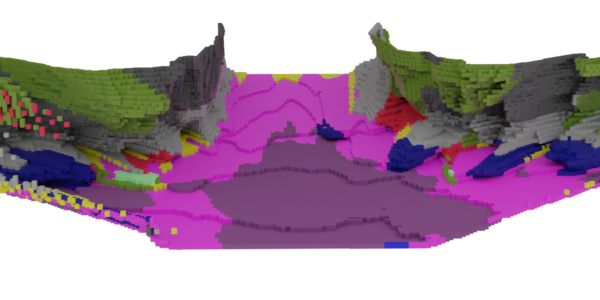} &
\includegraphics[width=\linewidth]{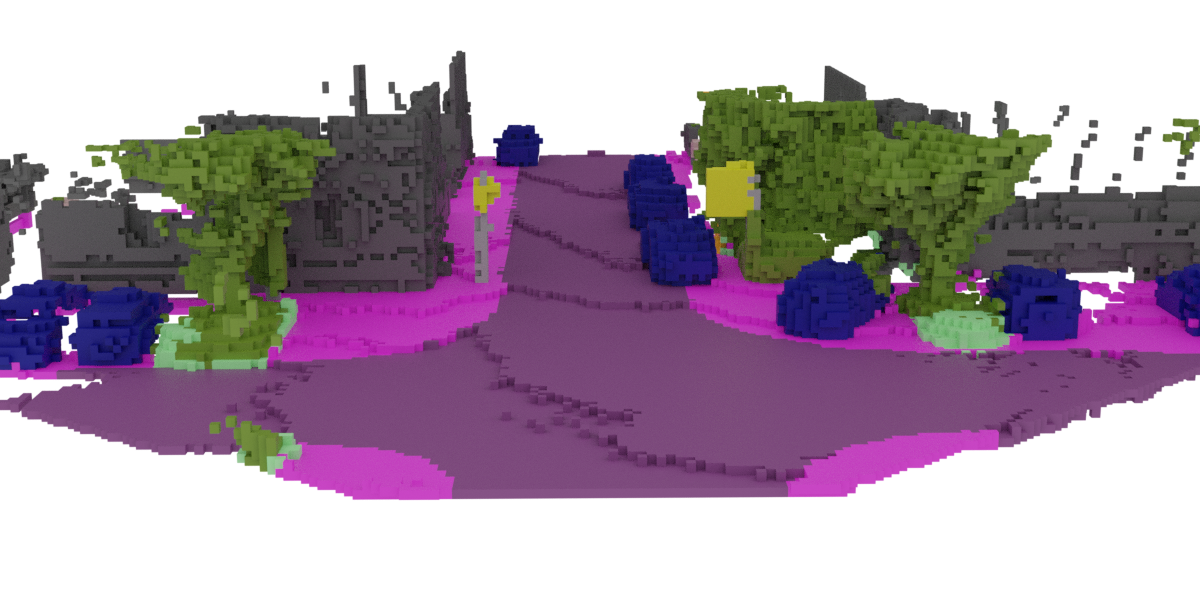} \\

\end{tabular}\\[-4.975pt]%
\tiny%
\renewcommand{\arraystretch}{1.3}%
\begin{tabularx}{0.9982\textwidth}{*{16}{>{\centering\arraybackslash}X}}  
    \cellcolor{road}\textcolor{white}{Road}
    & \cellcolor{sidewalk}\textcolor{white}{Sidewalk}
    & \cellcolor{building}\textcolor{white}{Building}
    & \cellcolor{fence}\textcolor{white}{Fence}
    & \cellcolor{pole}\textcolor{white}{Pole}
    & \cellcolor{trafficlight}\textcolor{white}{Other~Object}
    & \cellcolor{trafficsign}\textcolor{white}{Traffic~Sign}
    & \cellcolor{vegetation}\textcolor{white}{Vegetation}
    & \cellcolor{terrain}\textcolor{white}{Terrain}
    & \cellcolor{person}\textcolor{white}{Person}
    & \cellcolor{car}\textcolor{white}{Car}
    & \cellcolor{truck}\textcolor{white}{Other Vehicle}
    & \cellcolor{motorcycle}\textcolor{white}{Motorcycle}
    & \cellcolor{bicycle}\textcolor{white}{Bicycle}
\end{tabularx}

    \vspace{-0.3em}
    \caption{\textbf{3D qualitative SSC comparison on KITTI-360.} We provide additional qualitative results, visualizing the input image, \MethodName's predicted feature field using the first three principal components, and SSC prediction, the SSC prediction of our baseline S4C+STEGO, and the SSC ground truth. We only visualize surface voxels within the field of view for the sake of clarity.\label{fig:ssbench_supp}}
    \vspace{-0.6em}
\end{figure*}

\subsection{Datasets}

We provide additional details about the datasets utilized to train and evaluate \MethodName.

\inparagraphnohspace{KITTI-360~\cite{Liao:2023:KND, Li:2024:SSC}} 
provides video sequences from a moving vehicle equipped with a forward-facing stereo camera pair and two side-facing fisheye cameras. In future frames, the fisheye views capture additional geometric and semantic cues of regions occluded in the forward-facing view. For training, we resample the fisheye images into perspective projection. We focus on an area approximately 50 meters ahead of the ego vehicle. Assuming an average velocity of $30-50$\,km/h, side views are randomly sampled 1 – 4 seconds into the future. Given a frame rate of $10$\,Hz, this translates to 10 -- 40 time steps. Each training sample consists of eight images: four forward-facing views (including the input image) and four side-facing views.

To evaluate our predicted field in SSCBench-KITTI-360, we follow the evaluation procedure of S4C~\cite{Hayler:2024:S4C}. The voxel predictions are evaluated in three different ranges: $12.8\,\text{m} \times 12.8\,\text{m} \times 6.4\,\text{m}$, $25.6\,\text{m} \times 25.6\,\text{m} \times 6.4\,\text{m}$, and the full range $51.2\,\text{m} \times 51.2\,\text{m} \times 6.4\,\text{m}$. For each voxel, multiple evenly distributed points are sampled from the semantic field. The predictions are aggregated per voxel by taking the maximum occupancy and weighting the class predictions accordingly.

\inparagraphnohspace{Cityscapes~\cite{Cordts:2016:TCD}} consists of \num{500} high-resolution and densely annotated validation images of ego-centric driving scenes. For validation, Cityscapes uses a 19-class taxonomy. We leverage the Cityscapes validation samples at a resolution of $640\times 192$ for our domain generalization experiments (2D semantic segmentation).

\inparagraphnohspace{BDD-100K~\cite{Yu:2020:BDD}} is a driving scene dataset obtained from urban areas in the US. BDD-100K contains \num{1000} semantic segmentation validation images. The semantic taxonomy follows the 19-class Cityscapes definition. For domain generalization experiments, we utilize BDD-100K images at a resolution of $640\times 192$.

\inparagraphnohspace{RealEstate10K~\cite{Zhou:2018:SML}} is a large-scale dataset containing videos of real-world indoor and outdoor scenes, primarily sourced from YouTube. For our experiments, we train with a resolution of $512\times 288$. Each training sample consists of three frames, separated by a randomly sampled time offset. There are no semantic annotations provided with the dataset. We evaluate the multi-view consistency of our model in this setting.

\begin{figure*}[t]
    \centering
    \newcommand{\imgwidth}{0.452}
\newcommand{\dddviswidth}{0.182}

\scriptsize
\sffamily
\setlength{\tabcolsep}{0pt}
\renewcommand{\arraystretch}{0.66}
\begin{tabular}{>{\centering\arraybackslash} m{\imgwidth\textwidth} 
                >{\centering\arraybackslash} m{\dddviswidth\textwidth} 
                >{\centering\arraybackslash} m{\dddviswidth\textwidth} 
                >{\centering\arraybackslash} m{\dddviswidth\textwidth}}

\multirow{2}{*}{\vspace{-0.5em}\textbf{Input Image}} & \multicolumn{2}{c}{\textbf{\MethodName \textit{(Ours)}}} & \multirow{2}{*}{\vspace{-0.5em}\textbf{Ground Truth}} \\
\cmidrule(l{0.5em}r{0.5em}){2-3}
& Feature Field & SSC Prediction & \\[-1pt]

\includegraphics[width=\linewidth]{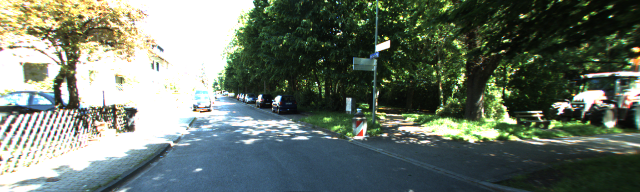} &
\includegraphics[width=\linewidth]{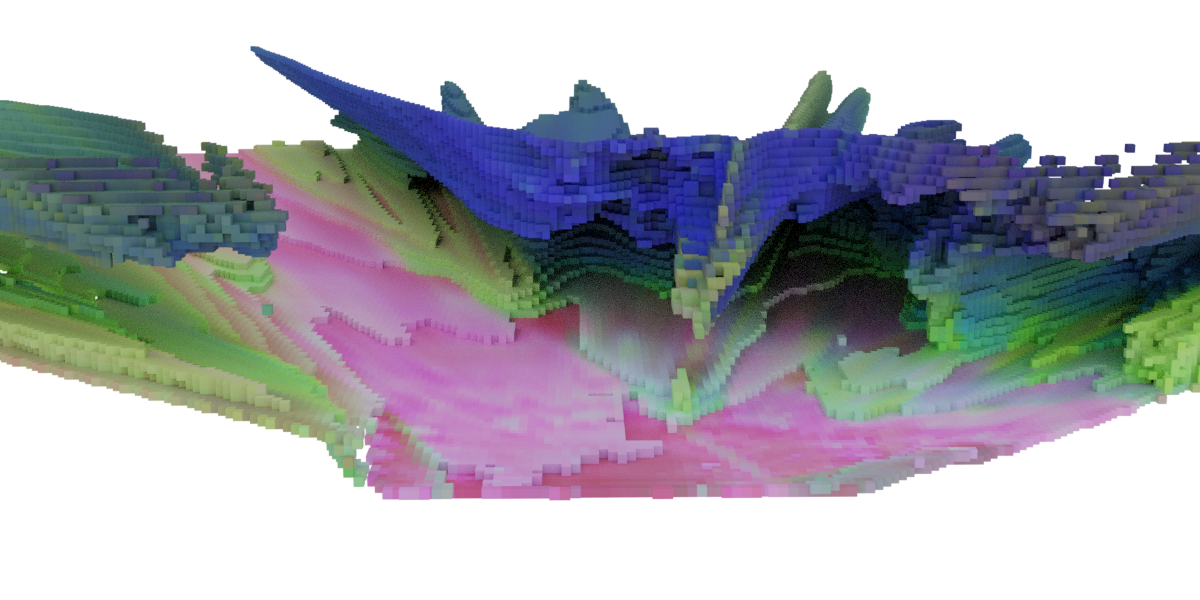} &
\includegraphics[width=\linewidth]{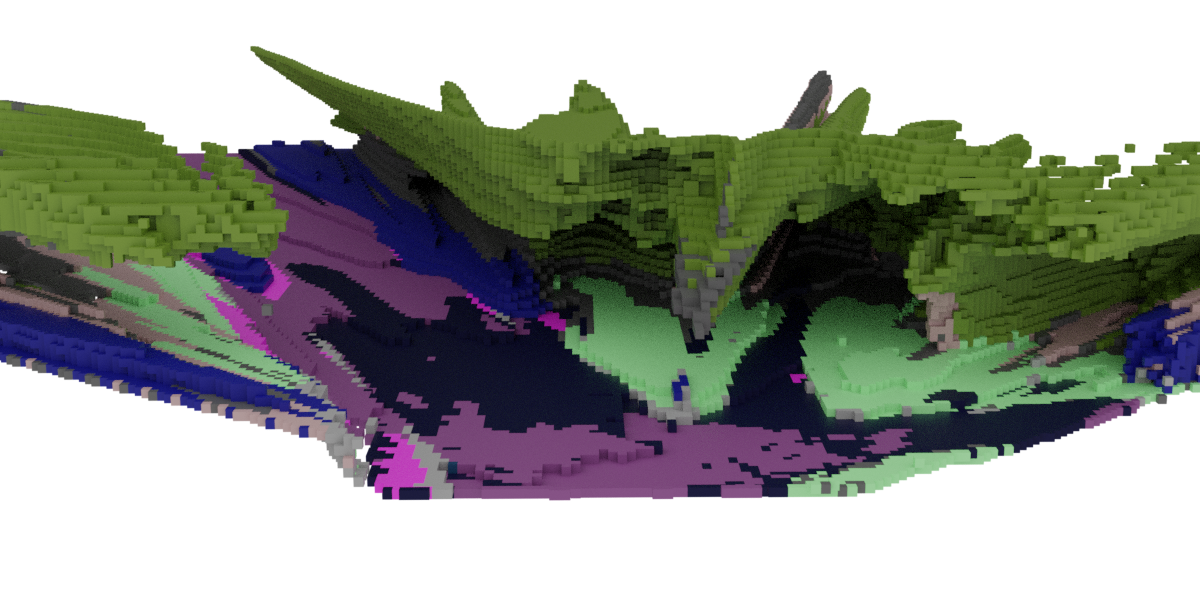} &
\includegraphics[width=\linewidth]{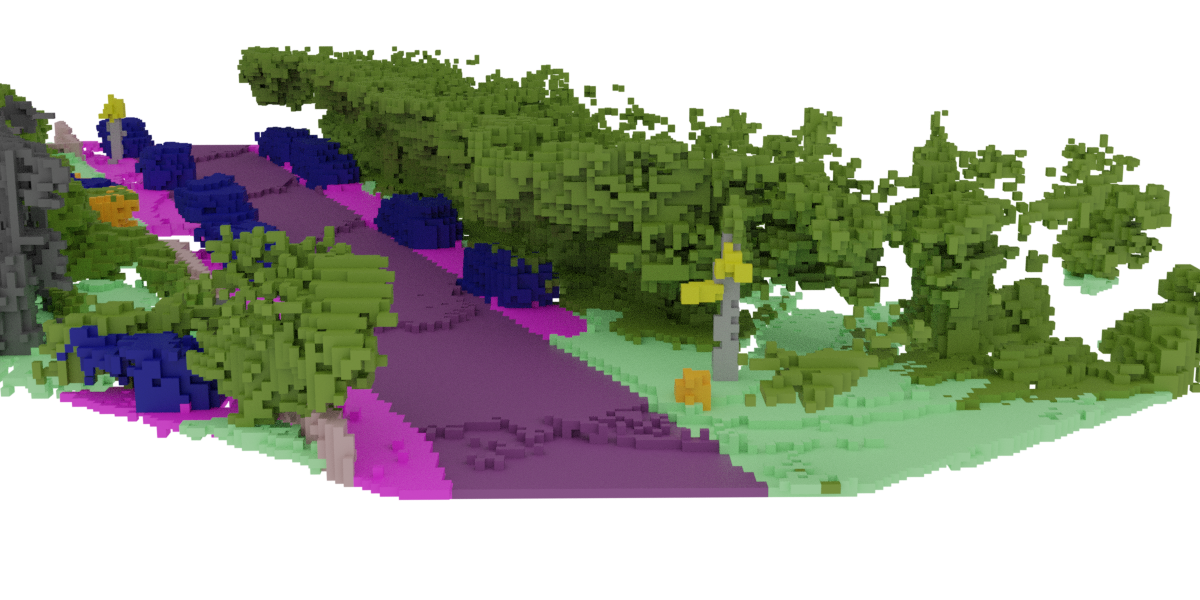} \\[-1.6pt]

\includegraphics[width=\linewidth]{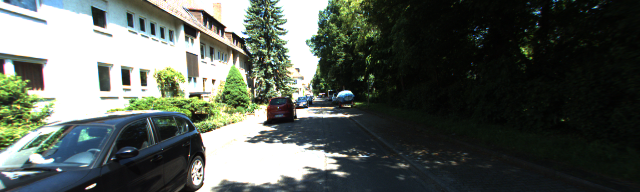} &
\includegraphics[width=\linewidth]{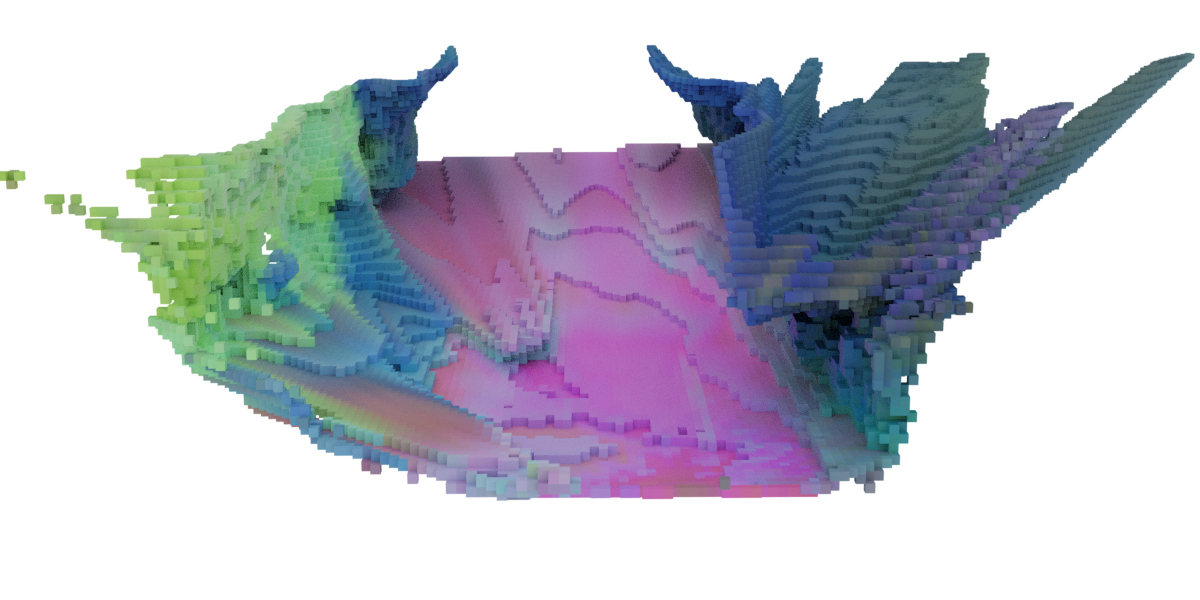} &
\includegraphics[width=\linewidth]{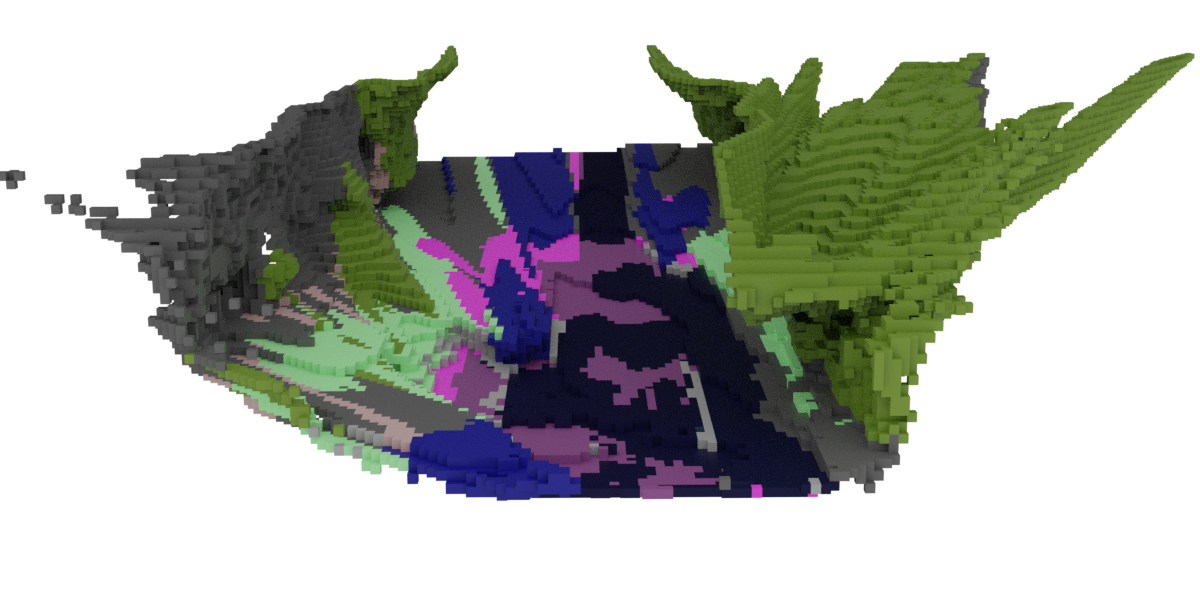} &
\includegraphics[width=\linewidth]{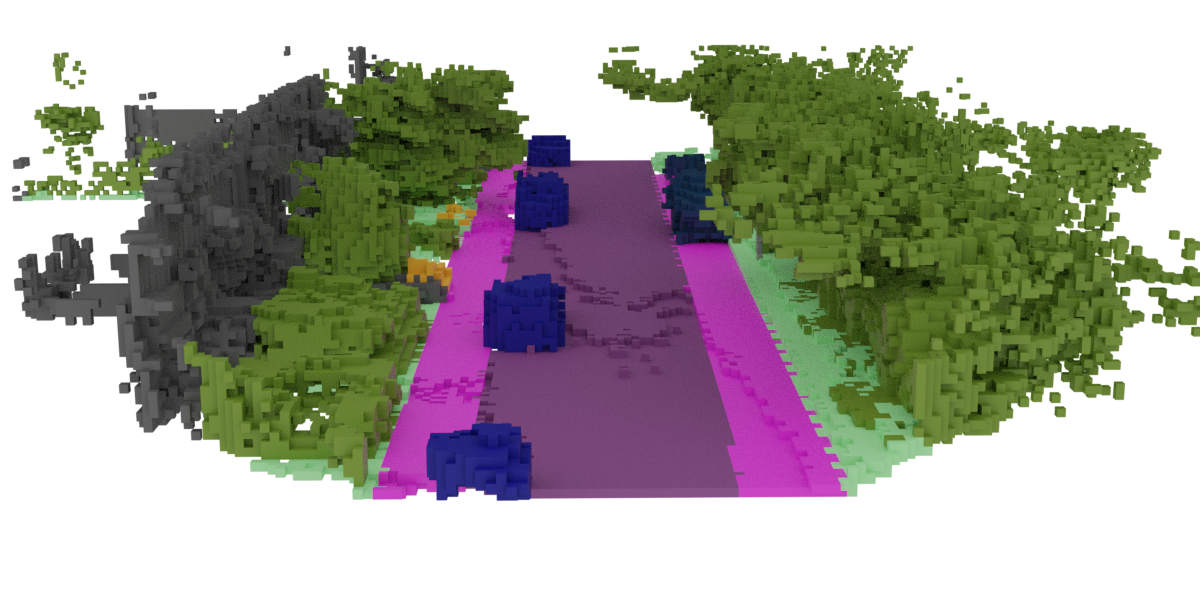} \\[-1.6pt]

\includegraphics[width=\linewidth]{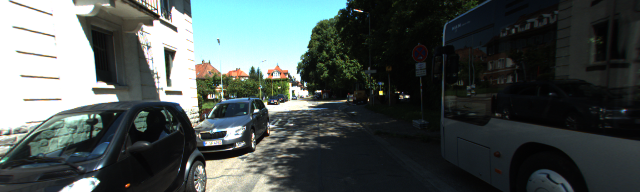} &
\includegraphics[width=\linewidth]{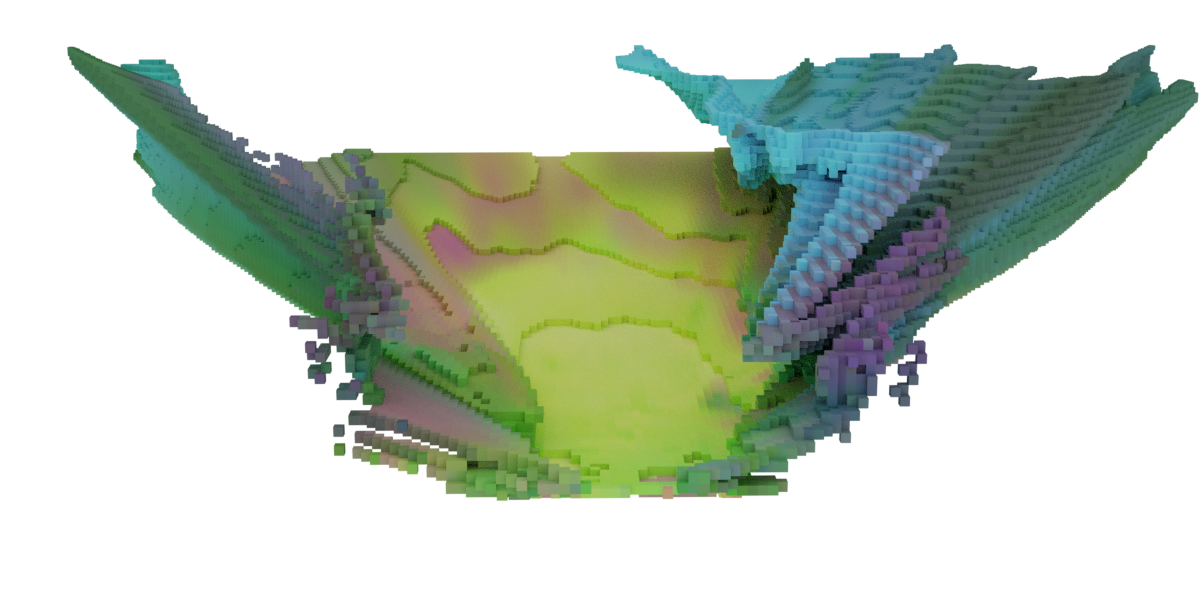} &
\includegraphics[width=\linewidth]{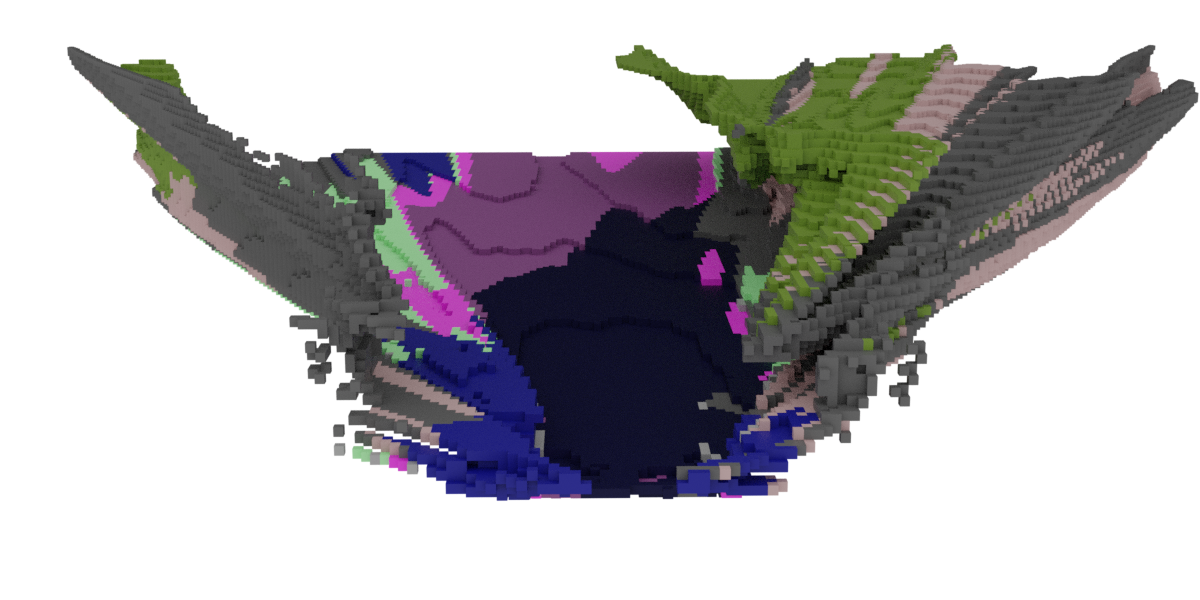} &
\includegraphics[width=\linewidth]{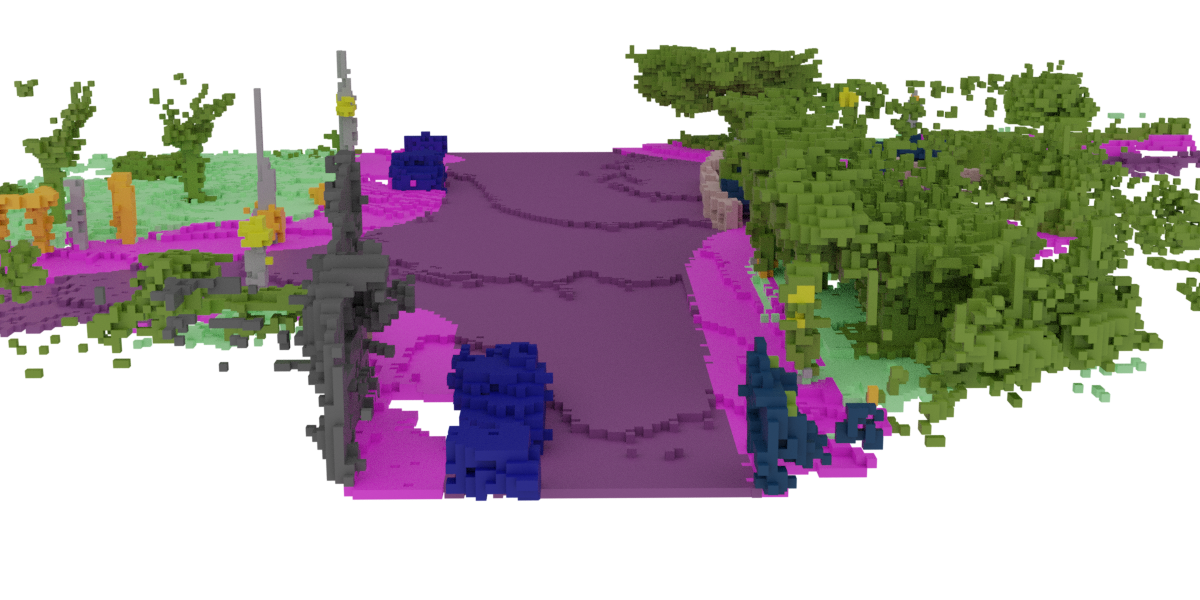} \\

\end{tabular}\\[-2.6pt]%
\tiny%
\renewcommand{\arraystretch}{1.3}%
\begin{tabularx}{0.9982\textwidth}{*{16}{>{\centering\arraybackslash}X}}  
    \cellcolor{road}\textcolor{white}{Road}
    & \cellcolor{sidewalk}\textcolor{white}{Sidewalk}
    & \cellcolor{building}\textcolor{white}{Building}
    & \cellcolor{fence}\textcolor{white}{Fence}
    & \cellcolor{pole}\textcolor{white}{Pole}
    & \cellcolor{trafficlight}\textcolor{white}{Other~Object}
    & \cellcolor{trafficsign}\textcolor{white}{Traffic~Sign}
    & \cellcolor{vegetation}\textcolor{white}{Vegetation}
    & \cellcolor{terrain}\textcolor{white}{Terrain}
    & \cellcolor{person}\textcolor{white}{Person}
    & \cellcolor{car}\textcolor{white}{Car}
    & \cellcolor{truck}\textcolor{white}{Other Vehicle}
    & \cellcolor{motorcycle}\textcolor{white}{Motorcycle}
    & \cellcolor{bicycle}\textcolor{white}{Bicycle}
\end{tabularx}

    \vspace{-0.3em}
    \caption{\textbf{Failure cases of \MethodName on KITTI-360.} We provide failure cases of \MethodName. We visualize the input image, the predicted feature field using the first three principal components, the SSC prediction, and the SSC ground truth. We observe that our semantic predictions struggle in shaded regions. We only visualize surface voxels within the field of view for the sake of clarity.\label{fig:ssbench_supp_fail}}
    \vspace{-0.6em}
\end{figure*}%
\begin{figure}[t]
    \centering
    \scriptsize
\sffamily
\setlength{\tabcolsep}{0pt}
\renewcommand{\arraystretch}{0.66}
\begin{tabular}{>{\centering\arraybackslash} m{0.33333333\columnwidth} 
                >{\centering\arraybackslash} m{0.33333333\columnwidth} 
                >{\centering\arraybackslash} m{0.33333333\columnwidth}}

\textbf{Input Image} & \textbf{\MethodName} & \textbf{DINO} \\[2pt]

\includegraphics[width=0.33333333\columnwidth]{artwork/experiments/3D_qualitative_kitti/good/2d_images/000120_image_0.png} &
\includegraphics[width=0.33333333\columnwidth]{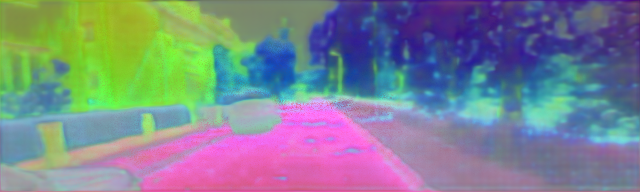} &
\includegraphics[width=0.33333333\columnwidth]{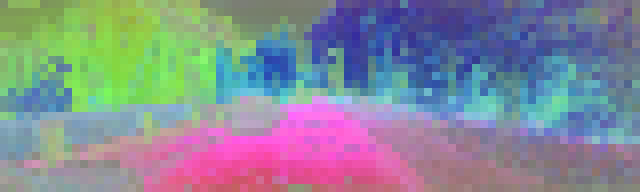} \\[-1.7pt]

\includegraphics[width=0.33333333\columnwidth]{artwork/experiments/3D_qualitative_kitti/good/2d_images/000185_image_0.png} &
\includegraphics[width=0.33333333\columnwidth]{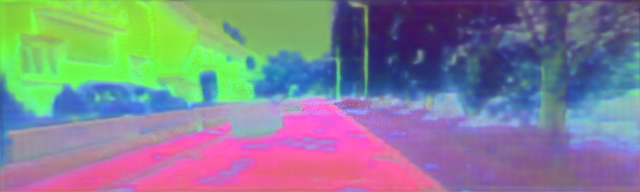} &
\includegraphics[width=0.33333333\columnwidth]{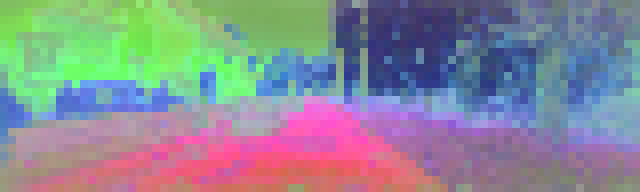} \\[-1.7pt]

\includegraphics[width=0.33333333\columnwidth]{artwork/experiments/3D_qualitative_kitti/good/2d_images/000240_image_0.png} &
\includegraphics[width=0.33333333\columnwidth]{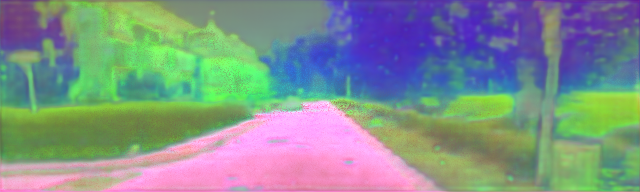} &
\includegraphics[width=0.33333333\columnwidth]{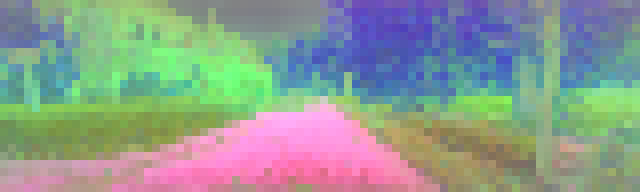} \\[-1.7pt]

\includegraphics[width=0.33333333\columnwidth]{artwork/experiments/3D_qualitative_kitti/good/2d_images/000500_image_0.png} &
\includegraphics[width=0.33333333\columnwidth]{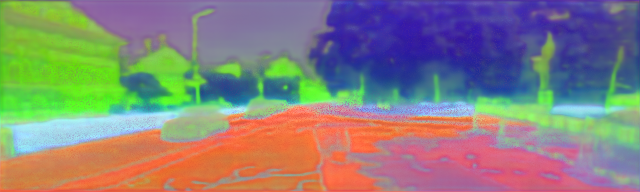} &
\includegraphics[width=0.33333333\columnwidth]{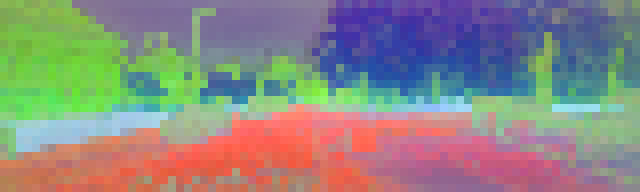} \\[-1.7pt]

\includegraphics[width=0.33333333\columnwidth]{artwork/experiments/3D_qualitative_kitti/good/2d_images/000640_image_0.png} &
\includegraphics[width=0.33333333\columnwidth]{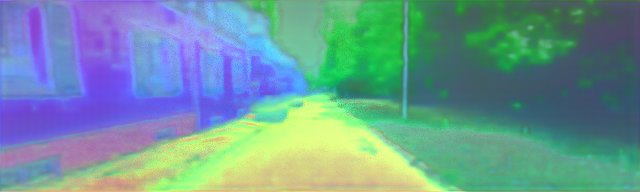} &
\includegraphics[width=0.33333333\columnwidth]{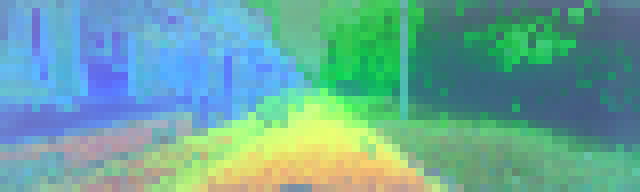} \\[-1.7pt]

\includegraphics[width=0.33333333\columnwidth]{artwork/experiments/3D_qualitative_kitti/good/2d_images/000655_image_0.png} &
\includegraphics[width=0.33333333\columnwidth]{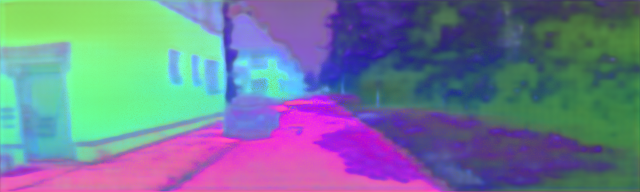} &
\includegraphics[width=0.33333333\columnwidth]{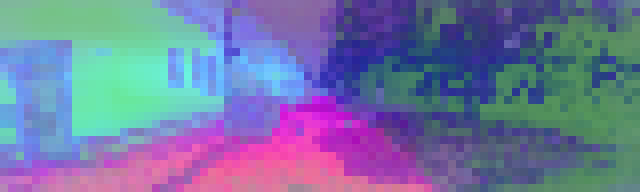} \\[-1.7pt]

\end{tabular}

    \vspace{-0.3em}
    \caption{\textbf{2D \MethodName features on KITTI-360.} We visualize our 2D rendered features and DINO features for a given input image \emph{(left)}. We use the first three principal components for feature visualization. Notably, \MethodName's features \emph{(middle)} are smoother and capture finer structures than DINO \emph{(right)}. Additionally, \MethodName's features are high-resolution, while DINO generates features with a stride of $8$.}
    \label{fig:features_2d}
    \vspace{-0.6em}
\end{figure}

\subsection{Computational complexity}

\MethodName requires only a \emph{single} GPU for training and inference. In SSCBench (\SI{51.2}{m} range), SceneDINO requires 0.76$\pm$0.1\,s to infer a full scene on a V100 GPU. The peak VRAM usage during inference is \SI{11}{GB}. For reference, S4C requires 0.32$\pm$0.13\,s. Considering our expressive and high-dimensional feature field and ViT encoder, this is a moderate runtime increase. SceneDINO has \SI{100}{M} parameters and is trained for approximately \num{2} days on a \emph{single} V100 \SI{32}{GB} GPU. All results are reported using automatic mixed precision.

\section{Multi-View Feature Consistency Evaluation\label{supp:mvfc}}

We aim to assess the multi-view consistency of 2D and 3D features in \cref{tab:mvc}. Note, we are not aware of any standardized approach for evaluating multi-view feature consistency. To this end, we employ a straightforward approach. Given two video frames with a temporal stride of \num{3}, forward optical flow is computed using RAFT large~\cite{Teed:2020:RAF}. We estimate occlusion by forward-backward consistency~\cite{occ}; for this, we also compute backward optical flow. 2D feature maps obtained using the second frame are backward warped to the 2D features of the first frame. We compute different similarity metrics between the aligned features (L\textsubscript{1}, L\textsubscript{2}, and cos-sim), ignoring occlusions. While features from DINO, DINOv2, and FiT3D possess a lower resolution than our 2D rendered \MethodName features, we upscale these features to the image resolution before warping. This evaluation approach utilizes optical flow correspondences and captures both ego motion as well as object motion, offering a simple way to evaluate multi-view feature consistency.

\section{Additional Results \label{supp:results}}

Here we provide additional qualitative and quantitative results, extending our results reported in the main paper.

\inparagraph{Qualitative results.} In \cref{fig:ssbench_supp}, we present additional qualitative results of \MethodName using our 3D feature distillation approach on unsupervised semantic scene completion. We also provide visualizations of our unsupervised SSC baseline, S4C + STEGO. Qualitatively, our approach obtains more accurate SSC results and is able to segment far-away objects, such as cars, better than the S4C + STEGO baseline. This observation aligns with the quantitative results presented in \cref{tab:sscbench} of the main paper.

\Cref{fig:features_2d} qualitatively analyzes our 2D rendered features against DINO. Our features exhibit a smooth appearance for uniform regions, such as sidewalks. Additionally, \MethodName's features better capture fine structures like poles than DINO features. 2D rendered \MethodName features are also high resolution in contrast to DINO features that exhibit a lower resolution.

\inparagraph{Failure cases.} In \cref{fig:ssbench_supp_fail}, we provide failure cases of \MethodName's SSC predictions. Our predictions exhibit two common failure cases. First, shadowed regions often lead to wrong semantic predictions. Regions affected by significant brightness changes are breaking the brightness consistency, subsequently offering a poor learning signal during training, thus impeding accurate predictions of shadowed regions. Second, objects such as cars can entail tail-like artifacts, not accurately capturing the geometry. As our multi-view image and feature reconstruction training cannot handle dynamic objects, tail-like artifacts could be caused by the poor learning signal for dynamic objects.

\begin{table*}[t]
    \centering
    \caption{\textbf{SSCBench-KITTI-360 results.} Semantic results using mIoU and per class IoU, and geometric results using IoU, Precision, and Recall (all in \%, $\uparrow$) on SSCBench-KITTI-360 test using three depth ranges. We extend \cref{tab:sscbench} and compare \MethodName against our baseline S4C~\cite{Hayler:2024:S4C} + STEGO~\cite{Hamilton:2022:USS}, 2D-supervised S4C~\cite{Hayler:2024:S4C}, and three 3D-supervised approaches (VoxFormer-S~\cite{Li:2023:VSV}, OccFormer~\cite{Zhang:2023:OCC}, and SSCNet~\cite{Song:2017:SSC}). Note that SSCNet uses depth as an additional input during inference, while all other approaches use a single input image.}
    \vspace{-0.3em}
    \renewcommand\tabcolsep{3.1525pt}
\scriptsize
\renewcommand{\arraystretch}{0.865}
\begin{tabularx}{\textwidth}{lS[table-format=2.2]S[table-format=2.2]S[table-format=2.2]|S[table-format=2.2]S[table-format=2.2]S[table-format=2.2]|S[table-format=2.2]S[table-format=2.2]S[table-format=2.2]|S[table-format=2.2]S[table-format=2.2]S[table-format=2.2]|S[table-format=2.2]S[table-format=2.2]S[table-format=2.2]|S[table-format=2.2]S[table-format=2.2]S[table-format=2.2]}
\toprule
\textbf{Method} & \multicolumn{3}{c|}{\textbf{S4C + STEGO}} & \multicolumn{3}{c|}{\textbf{\MethodName (Ours)}} & \multicolumn{3}{c|}{\textbf{\color{tud0c!95} S4C}} & \multicolumn{3}{c|}{\textbf{\color{tud0c!95} VoxFormer-S}} & \multicolumn{3}{c|}{\textbf{\color{tud0c!95} OccFormer}} & \multicolumn{3}{c}{\textbf{\color{tud0c!95} SSCNet}} \\
\midrule
\textbf{Supervision} & \multicolumn{6}{c}{\cellcolor{LimeGreen!70}\textbf{Unsupervised}} & \multicolumn{3}{c}{\cellcolor{RedOrange!70}\textbf{2D supervision}} & \multicolumn{6}{c}{\cellcolor{RedOrange!80}\textbf{3D supervision}} & \multicolumn{3}{c}{\cellcolor{RedOrange!90}\textbf{3D sup. + depth input}} \\
\midrule
\textbf{Range} & {\SI{12.8}{\meter}} & {\SI{25.6}{\meter}} & {\SI{51.2}{\meter}} & {\SI{12.8}{\meter}} & {\SI{25.6}{\meter}} & {\SI{51.2}{\meter}} & \color{tud0c!95} {\SI{12.8}{\meter}} & \color{tud0c!95} {\SI{25.6}{\meter}} & \color{tud0c!95} {\SI{51.2}{\meter}} & \color{tud0c!95} {\SI{12.8}{\meter}} & \color{tud0c!95} {\SI{25.6}{\meter}} & \color{tud0c!95} {\SI{51.2}{\meter}} & \color{tud0c!95} {\SI{12.8}{\meter}} & \color{tud0c!95} {\SI{25.6}{\meter}} & \color{tud0c!95} {\SI{51.2}{\meter}}& \color{tud0c!95} {\SI{12.8}{\meter}} & \color{tud0c!95} {\SI{25.6}{\meter}} & \color{tud0c!95} {\SI{51.2}{\meter}} \\
\specialrule{0.5pt}{1.5pt}{2pt}
\multicolumn{19}{c}{\textit{Semantic validation}} \\
\specialrule{0.5pt}{1.5pt}{2pt}
\cellcolor{tud0c!20}\textbf{mIoU} & \cellcolor{tud0c!20}10.53 & \cellcolor{tud0c!20}9.26 & \cellcolor{tud0c!20}6.60 & \cellcolor{tud0c!20}\bfseries 10.76 & \cellcolor{tud0c!20}\bfseries 10.01 & \cellcolor{tud0c!20}\bfseries 8.00 & \cellcolor{tud0c!20}\color{tud0c!95}16.94 & \cellcolor{tud0c!20}\color{tud0c!95}13.94 & \cellcolor{tud0c!20}\color{tud0c!95}10.19 & \cellcolor{tud0c!20}\color{tud0c!95}18.17 & \cellcolor{tud0c!20}\color{tud0c!95}15.40 & \cellcolor{tud0c!20}\color{tud0c!95}11.91 & \cellcolor{tud0c!20}\color{tud0c!95}23.04 & \cellcolor{tud0c!20}\color{tud0c!95}18.38 & \cellcolor{tud0c!20}\color{tud0c!95}13.81 & \cellcolor{tud0c!20}\color{tud0c!95}26.64 & \cellcolor{tud0c!20}\color{tud0c!95}24.33 & \cellcolor{tud0c!20}\color{tud0c!95}19.23 \\
car           & 18.57 & 14.09 & 9.22 & 21.24 & 15.94 & 11.21 & \color{tud0c!95}22.58 & \color{tud0c!95}18.64 & \color{tud0c!95}11.49 & \color{tud0c!95}29.41 & \color{tud0c!95}25.08 & \color{tud0c!95}17.84 & \color{tud0c!95}40.87 & \color{tud0c!95}33.10 & \color{tud0c!95}22.58 & \color{tud0c!95}52.72 & \color{tud0c!95}45.93 & \color{tud0c!95}31.89 \\
bicycle       & 0.01 & 0.01 & 0.01 &  0.00 & 0.00& 0.00  & \color{tud0c!95}0.00 & \color{tud0c!95}0.00 &  \color{tud0c!95}0.00 & \color{tud0c!95}2.73 & \color{tud0c!95}1.73 & \color{tud0c!95}1.16 & \color{tud0c!95}1.94 & \color{tud0c!95}1.04 & \color{tud0c!95}0.66 & \color{tud0c!95}0.00 & \color{tud0c!95}0.00 & \color{tud0c!95}0.00 \\
motorcycle    & 0.00 & 0.00 & 0.00 &  0.00 & 0.00& 0.00  & \color{tud0c!95}0.00 & \color{tud0c!95}0.00 & \color{tud0c!95}0.00 & \color{tud0c!95}1.97 & \color{tud0c!95}1.47 & \color{tud0c!95}0.89 & \color{tud0c!95}1.03 & \color{tud0c!95}0.43 & \color{tud0c!95}0.26 & \color{tud0c!95}1.41 & \color{tud0c!95}0.41 & \color{tud0c!95}0.19 \\
truck         & 0.11 & 0.04 & 0.02 &  0.00 & 0.00& 0.00  & \color{tud0c!95}7.51 &\color{tud0c!95} 4.37 & \color{tud0c!95}2.12 & \color{tud0c!95}6.08 & \color{tud0c!95}6.63 & \color{tud0c!95}4.56 & \color{tud0c!95}22.40 & \color{tud0c!95}15.21 & \color{tud0c!95}9.89 & \color{tud0c!95}16.91 & \color{tud0c!95}14.91 & \color{tud0c!95}10.78 \\
other-v. & 0.01 & 0.05 & 0.02 &  0.00 & 0.00& 0.00  & \color{tud0c!95}0.00 & \color{tud0c!95}0.01 & \color{tud0c!95}0.06 & \color{tud0c!95}3.71 & \color{tud0c!95}3.56 & \color{tud0c!95}2.06 & \color{tud0c!95}8.48 & \color{tud0c!95}6.12 & \color{tud0c!95}3.82 & \color{tud0c!95}1.45 & \color{tud0c!95}1.00 & \color{tud0c!95}0.60 \\
person        & 0.01 & 0.01 & 0.01 &  0.00 & 0.00& 0.00  & \color{tud0c!95}0.00 & \color{tud0c!95}0.00 & \color{tud0c!95}0.00 & \color{tud0c!95}2.06 & \color{tud0c!95}2.20 & \color{tud0c!95}1.63 & \color{tud0c!95}4.54 & \color{tud0c!95}3.79 & \color{tud0c!95}2.77 & \color{tud0c!95}0.36 & \color{tud0c!95}0.16 & \color{tud0c!95}0.09 \\
road          & 61.97 & 52.47 & 38.15 &  51.10 & 49.12 & 39.82  & \color{tud0c!95}69.38 & \color{tud0c!95}61.46 & \color{tud0c!95}48.23 & \color{tud0c!95}66.10 & \color{tud0c!95}58.58 & \color{tud0c!95}47.01 & \color{tud0c!95}73.34 & \color{tud0c!95}66.53 & \color{tud0c!95}54.30 & \color{tud0c!95}87.81 & \color{tud0c!95}85.42 & \color{tud0c!95}73.82 \\
sidewalk      & 18.74 & 20.95 & 18.21 &  20.26 & 22.31 & 18.97  & \color{tud0c!95}45.03 & \color{tud0c!95}37.12 & \color{tud0c!95}28.45 & \color{tud0c!95}38.00 & \color{tud0c!95}33.63 & \color{tud0c!95}27.20 & \color{tud0c!95}49.76 & \color{tud0c!95}41.30 & \color{tud0c!95}31.53 & \color{tud0c!95}67.19 & \color{tud0c!95}60.34 & \color{tud0c!95}46.96 \\
building      & 14.75 & 24.44 & 17.81 &  12.33 & 18.27 & 14.32  & \color{tud0c!95}26.34 & \color{tud0c!95}28.48 & \color{tud0c!95}21.36 & \color{tud0c!95}41.12 & \color{tud0c!95}38.24 & \color{tud0c!95}31.18 & \color{tud0c!95}53.65 & \color{tud0c!95}44.86 & \color{tud0c!95}36.42 & \color{tud0c!95}53.93 & \color{tud0c!95}54.55 & \color{tud0c!95}44.67 \\
fence         & 1.41 & 0.20 & 0.11 &  1.91 & 0.90 & 0.58  & \color{tud0c!95}9.70 & \color{tud0c!95}6.37 & \color{tud0c!95}3.64 & \color{tud0c!95}8.99 & \color{tud0c!95}7.43 & \color{tud0c!95}4.97 & \color{tud0c!95}10.64 & \color{tud0c!95}7.85 & \color{tud0c!95}4.80 & \color{tud0c!95}14.39 & \color{tud0c!95}10.73 & \color{tud0c!95}6.42 \\
vegetation    & 15.83 & 16.58 & 11.30 &  31.22 & 25.57 & 19.85  & \color{tud0c!95}35.78 & \color{tud0c!95}28.04 & \color{tud0c!95}21.43 & \color{tud0c!95}45.68 & \color{tud0c!95}35.16 & \color{tud0c!95}28.99 & \color{tud0c!95}49.91 & \color{tud0c!95}37.96 & \color{tud0c!95}31.00 & \color{tud0c!95}56.66 & \color{tud0c!95}51.77 & \color{tud0c!95}43.30 \\
terrain       & 26.49 &  9.95 & 4.17 &  23.26 & 18.02 & 15.22  & \color{tud0c!95}35.03 & \color{tud0c!95}22.88 & \color{tud0c!95}15.08 & \color{tud0c!95}24.70 & \color{tud0c!95}18.53 & \color{tud0c!95}14.69 & \color{tud0c!95}34.63 & \color{tud0c!95}24.99 & \color{tud0c!95}19.51 & \color{tud0c!95}43.47 & \color{tud0c!95}36.44 & \color{tud0c!95}27.83 \\
pole          & 0.08 & 0.04 & 0.04 &  0.05 & 0.05 & 0.05  & \color{tud0c!95}1.23 & \color{tud0c!95}0.94 & \color{tud0c!95}0.65 & \color{tud0c!95}8.84 & \color{tud0c!95}8.16 & \color{tud0c!95}6.51 & \color{tud0c!95}12.93 & \color{tud0c!95}10.25 & \color{tud0c!95}7.77 & \color{tud0c!95}1.03 & \color{tud0c!95}1.05 & \color{tud0c!95}0.62 \\
traffic-sign  & 0.00 & 0.00 & 0.00 &  0.00 & 0.00 & 0.00  & \color{tud0c!95}1.57 & \color{tud0c!95}0.83 & \color{tud0c!95}0.36 & \color{tud0c!95}9.15 & \color{tud0c!95}9.02 & \color{tud0c!95}6.92 & \color{tud0c!95}14.25 & \color{tud0c!95}12.37 & \color{tud0c!95}8.51 & \color{tud0c!95}1.01 & \color{tud0c!95}1.22 & \color{tud0c!95}0.70 \\
other-obj. & 0.05 & 0.04 & 0.02 &  0.00 & 0.00 & 0.00  & \color{tud0c!95}0.00 & \color{tud0c!95}0.00 & \color{tud0c!95}0.00 & \color{tud0c!95}4.40 & \color{tud0c!95}3.27 & \color{tud0c!95}2.43 & \color{tud0c!95}8.96 & \color{tud0c!95}6.71 & \color{tud0c!95}4.60 & \color{tud0c!95}1.20 & \color{tud0c!95}0.97 & \color{tud0c!95}0.58 \\
\specialrule{0.5pt}{1.5pt}{2pt}
\multicolumn{19}{c}{\textit{Geometric validation}} \\
\specialrule{0.5pt}{1.5pt}{2pt}
\cellcolor{tud0c!20}\textbf{IoU}           & \cellcolor{tud0c!20}49.32 & \cellcolor{tud0c!20}41.08 & \cellcolor{tud0c!20}36.39 & \cellcolor{tud0c!20}\bfseries 49.54 & \cellcolor{tud0c!20}\bfseries 42.27 & \cellcolor{tud0c!20}\bfseries 37.60 & \cellcolor{tud0c!20}\color{tud0c!95}54.64 & \cellcolor{tud0c!20}\color{tud0c!95}45.57 & \cellcolor{tud0c!20}\color{tud0c!95}39.35 & \cellcolor{tud0c!20}\color{tud0c!95}55.45 & \cellcolor{tud0c!20}\color{tud0c!95}46.36 & \cellcolor{tud0c!20}\color{tud0c!95}38.76 & \cellcolor{tud0c!20}\color{tud0c!95}58.71 & \cellcolor{tud0c!20}\color{tud0c!95}47.96 & \cellcolor{tud0c!20}\color{tud0c!95}40.27 & \cellcolor{tud0c!20}\color{tud0c!95}74.93 & \cellcolor{tud0c!20}\color{tud0c!95}66.36 & \cellcolor{tud0c!20}\color{tud0c!95}55.81 \\
Precision     & 54.04 & 46.23 & 41.91 &  53.27 & 46.10 & 41.59  & \color{tud0c!95}59.75 & \color{tud0c!95}50.34 & \color{tud0c!95}43.59 & \color{tud0c!95}66.10 & \color{tud0c!95}61.34 & \color{tud0c!95}58.52 & \color{tud0c!95}69.47 & \color{tud0c!95}62.68 & \color{tud0c!95}59.70 & \color{tud0c!95}83.65 & \color{tud0c!95}77.85 & \color{tud0c!95}75.41 \\
Recall        & 84.95 & 78.69 & 73.43 & 87.61 & 83.59 & 79.67  & \color{tud0c!95}86.47 & \color{tud0c!95}82.79 & \color{tud0c!95}80.16  & \color{tud0c!95}77.48 & \color{tud0c!95}65.48 & \color{tud0c!95}53.44 & \color{tud0c!95}79.13 & \color{tud0c!95}67.12 & \color{tud0c!95}55.31 & \color{tud0c!95}87.79 & \color{tud0c!95}81.80 & \color{tud0c!95}68.22 \\
\bottomrule
\end{tabularx}

    \label{tab:sscbenchfull}
    \vspace{-0.6em}
\end{table*}

\begin{table}[t]
    \centering
    \caption{\textbf{SSCBench-KITTI-360 results (DINOv2).} Semantic results using mIoU and per class IoU, and geometric results using IoU, Precision, and Recall (all in \%, $\uparrow$) on SSCBench-KITTI-360 test using three depth ranges. We compare our baseline S4C + STEGO to \MethodName, both using DINOv2 features.}
    \vspace{-0.3em}
    \renewcommand\tabcolsep{6.22pt}
\scriptsize
\renewcommand{\arraystretch}{0.865}
\begin{tabularx}{\columnwidth}{lS[table-format=2.2]S[table-format=2.2]S[table-format=2.2]|S[table-format=2.2]S[table-format=2.2]S[table-format=2.2]}
\toprule
\textbf{Method} & \multicolumn{3}{c|}{\textbf{S4C + STEGO {\tiny{}w/ DINOv2}}} & \multicolumn{3}{c}{\textbf{\MethodName{} {\tiny{}w/ DINOv2} (Ours)}} \\
\midrule
\textbf{Supervision} & \multicolumn{6}{c}{\cellcolor{LimeGreen!70}\textbf{Unsupervised}} \\
\midrule
\textbf{Range} & {\SI{12.8}{\meter}} & {\SI{25.6}{\meter}} & {\SI{51.2}{\meter}} & {\SI{12.8}{\meter}} & {\SI{25.6}{\meter}} & {\SI{51.2}{\meter}} \\
\specialrule{0.5pt}{1.5pt}{2pt}
\multicolumn{7}{c}{\textit{Semantic validation}} \\
\specialrule{0.5pt}{1.5pt}{2pt}
\cellcolor{tud0c!20}\textbf{mIoU} & \cellcolor{tud0c!20}11.70 & \cellcolor{tud0c!20}9.27 & \cellcolor{tud0c!20}6.25 & \cellcolor{tud0c!20}\bfseries 13.76 & \cellcolor{tud0c!20}\bfseries 11.78 & \cellcolor{tud0c!20}\bfseries 9.08 \\
car           &  15.66 & 10.31 &  5.84  &  18.27 & 13.83 &  9.51  \\
bicycle       &   0.00 &  0.00 &  0.00  &   0.00 &  0.00 &  0.00  \\
motorcycle    &   0.00 &  0.00 &  0.00  &   0.00 &  0.00 &  0.00 \\
truck         &   0.00 &  0.00 &  0.00  &   0.00 &  0.00 &  0.00  \\
other-v.      &   0.01 &  0.01 &  0.01  &   0.00 &  0.00 &  0.00  \\
person        &   0.00 &  0.00 &  0.00  &   0.00 &  0.00 &  0.00  \\
road          &  65.81 & 55.73 & 35.00  &  68.04 & 61.35 & 46.70  \\
sidewalk      &  31.78 & 24.13 & 19.43  &  41.63 & 36.02 & 27.32  \\
building      &   0.83 &  0.41 &  0.23  &  15.97 & 20.87 & 16.81  \\
fence         &   0.89 &  0.57 &  0.41  &   0.00 &  0.00 &  0.00  \\
vegetation    &   9.92 & 11.42 &  9.24  &  25.37 & 17.86 & 14.82  \\
terrain       &  33.79 & 15.96 &  8.45  &  37.07 & 26.81 & 21.06  \\
pole          &  16.84 & 20.43 & 15.14  &   0.00 &  0.00 &  0.00  \\
traffic-sign  &   0.00 &  0.00 &  0.01  &   0.00 &  0.00 &  0.00  \\
other-obj.    &   0.01 &  0.01 &  0.02  &   0.00 &  0.00 &  0.00  \\

\specialrule{0.5pt}{1.5pt}{2pt}
\multicolumn{7}{c}{\textit{Geometric validation}} \\
\specialrule{0.5pt}{1.5pt}{2pt}
\cellcolor{tud0c!20}\textbf{IoU}           & \cellcolor{tud0c!20} 47.51 & \cellcolor{tud0c!20} 39.99 & \cellcolor{tud0c!20} 35.63 & \cellcolor{tud0c!20}\bfseries 48.12 & \cellcolor{tud0c!20}\bfseries 40.35 & \cellcolor{tud0c!20}\bfseries 36.21 \\
Precision     &  55.89 & 47.32 & 42.36  &  52.95 & 45.44 & 40.92  \\
Recall        &  76.02 & 72.06 & 69.14  &  84.07 & 78.29 & 75.89  \\
\bottomrule
\end{tabularx}

    \label{tab:sscbenchdinov2}
    \vspace{-0.6em}
\end{table}

\begin{table}[t]
    \centering
    \caption{\textbf{Class-wise 2D unsupervised semantic segmentation results on KITTI-360}. We compare the class-wise IoU scores (all in \%, $\uparrow$) of \MethodName against STEGO in 2D on the SSCBench-KITTI-360 test split.}
    \vspace{-0.3em}
    \renewcommand\tabcolsep{28.8pt}
\scriptsize
\renewcommand{\arraystretch}{0.865}
\begin{tabularx}{\columnwidth}{>{\hspace{-\tabcolsep}\raggedright\columncolor{white}[\tabcolsep][\tabcolsep]}lS[table-format=2.2]|S[table-format=2.2]}
\toprule
\;\textbf{Method} & \multicolumn{1}{c|}{\textbf{STEGO}} & \multicolumn{1}{c}{\textbf{\MethodName}} \\
\midrule
\cellcolor{tud0c!20}\;\textbf{mIoU} & \cellcolor{tud0c!20}\bfseries 23.57 & \cellcolor{tud0c!20}\bfseries 25.81 \\
\;{road}           & 63.81 & 77.73 \\
\;{sidewalk}       & 7.70 & 44.48 \\
\;{building}       & 65.24 & 77.67 \\
\;{wall}           & 11.94 & 3.68 \\
\;{fence}          & 15.36 & 18.13 \\
\;{pole}           & 11.43 & 0.93 \\
\;{traffic light}  & 0.00 & 0.00 \\
\;{traffic sign}   & 0.11 & 0.00 \\
\;{vegetation}     & 73.35 & 73.38 \\
\;{terrain}        & 49.31 & 41.29 \\
\;{sky}            & 69.18 & 71.72 \\
\;{person}         & 0.00 & 0.00 \\
\;{rider}          & 0.05 & 0.00 \\
\;{car}            & 77.72 & 81.31 \\
\;{truck}          & 2.09 & 0.04 \\
\;{bus}            & 0.02 & 0.00 \\
\;{train}          & 0.00 & 0.00 \\
\;{motorcycle}     & 0.08 & 0.00 \\
\;{bicycle}        & 0.00 & 0.00 \\

\bottomrule
\end{tabularx}

    \label{tab:semantic_segmentation_2d_results_classes}
    \vspace{-0.6em}
\end{table}

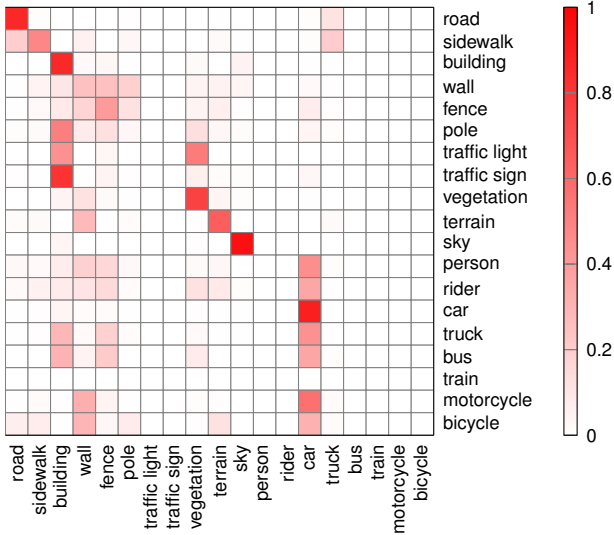
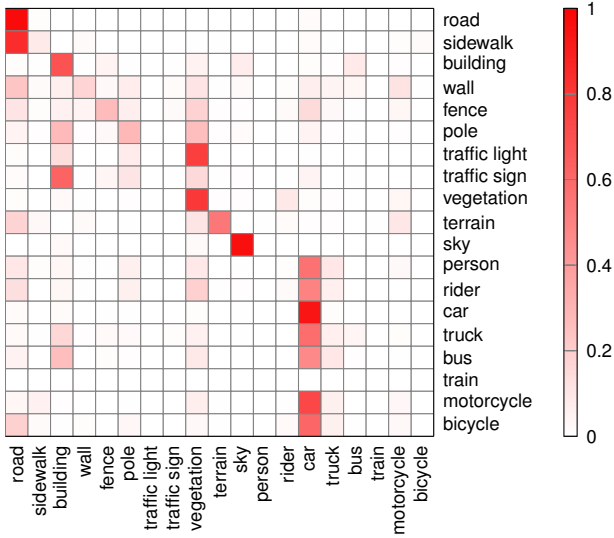
\begin{figure}[t]
    \centering
    \begin{subfigure}{\columnwidth}
        \begin{tikzpicture}[every node/.style={font=\sffamily\scriptsize},]
    \begin{axis}[
        axis equal image,
        colormap={bluewhite}{color=(white) rgb255=(90,96,191)},
        xticklabels={road, sidewalk, building, wall, fence, pole, traffic light, traffic sign, vegetation, terrain, sky, person, rider, car, truck, bus, train, motorcycle, bicycle}, %
        xtick={0,...,18}, %
        xtick style={draw=none},
        axis y line*=right, 
        yticklabel pos=right,
        yticklabels={road, sidewalk, building, wall, fence, pole, traffic light, traffic sign, vegetation, terrain, sky, person, rider, car, truck, bus, train, motorcycle, bicycle}, %
        ytick={0,...,18}, %
        ytick style={draw=none},
        tick label style={font=\sffamily\scriptsize},
        enlargelimits=false,
        colorbar,
        colorbar style={
            width=5pt,
            ytick={0, 0.2, 0.4, 0.6, 0.8, 1},
            yticklabels={0, 0.2, 0.4, 0.6, 0.8, 1},
        },
        xticklabel style={
          rotate=90
        },
        point meta max=1.0,
        point meta min=0,
        colormap={svbcolorbar}{[1pt]
          rgb(0pt)=(1,1,1);
          rgb(200pt)=(0.9803921568627451,0.01568627450980392,0.01568627450980392)
        },
    ]
    \addplot[
        matrix plot,
        mesh/cols=19, %
        point meta=explicit,draw=gray
    ] table [meta=C] {
    
x y C
0 0 0.856
1 0 0.013
2 0 0.000
3 0 0.001
4 0 0.000
5 0 0.006
6 0 0.000
7 0 0.000
8 0 0.000
9 0 0.003
10 0 0.000
11 0 0.000
12 0 0.000
13 0 0.010
14 0 0.111
15 0 0.000
16 0 0.000
17 0 0.000
18 0 0.000

0 1 0.197
1 1 0.479
2 1 0.001
3 1 0.056
4 1 0.001
5 1 0.032
6 1 0.000
7 1 0.000
8 1 0.000
9 1 0.020
10 1 0.002
11 1 0.000
12 1 0.000
13 1 0.010
14 1 0.203
15 1 0.000
16 1 0.000
17 1 0.000
18 1 0.000

0 2 0.000
1 2 0.002
2 2 0.857
3 2 0.022
4 2 0.035
5 2 0.004
6 2 0.000
7 2 0.000
8 2 0.021
9 2 0.001
10 2 0.053
11 2 0.000
12 2 0.000
13 2 0.003
14 2 0.002
15 2 0.000
16 2 0.000
17 2 0.000
18 2 0.000

0 3 0.004
1 3 0.049
2 3 0.098
3 3 0.248
4 3 0.249
5 3 0.187
6 3 0.000
7 3 0.000
8 3 0.043
9 3 0.050
10 3 0.039
11 3 0.000
12 3 0.000
13 3 0.025
14 3 0.009
15 3 0.000
16 3 0.000
17 3 0.000
18 3 0.000

0 4 0.005
1 4 0.022
2 4 0.087
3 4 0.173
4 4 0.401
5 4 0.116
6 4 0.000
7 4 0.000
8 4 0.042
9 4 0.065
10 4 0.008
11 4 0.000
12 4 0.000
13 4 0.073
14 4 0.009
15 4 0.000
16 4 0.000
17 4 0.000
18 4 0.000

0 5 0.013
1 5 0.019
2 5 0.510
3 5 0.078
4 5 0.122
5 5 0.032
6 5 0.000
7 5 0.000
8 5 0.128
9 5 0.035
10 5 0.014
11 5 0.000
12 5 0.000
13 5 0.039
14 5 0.010
15 5 0.000
16 5 0.000
17 5 0.000
18 5 0.000

0 6 0.000
1 6 0.000
2 6 0.444
3 6 0.000
4 6 0.034
5 6 0.000
6 6 0.000
7 6 0.000
8 6 0.522
9 6 0.000
10 6 0.000
11 6 0.000
12 6 0.000
13 6 0.000
14 6 0.000
15 6 0.000
16 6 0.000
17 6 0.000
18 6 0.000

0 7 0.009
1 7 0.005
2 7 0.812
3 7 0.005
4 7 0.044
5 7 0.008
6 7 0.000
7 7 0.000
8 7 0.060
9 7 0.014
10 7 0.010
11 7 0.000
12 7 0.000
13 7 0.033
14 7 0.000
15 7 0.000
16 7 0.000
17 7 0.000
18 7 0.000

0 8 0.000
1 8 0.001
2 8 0.045
3 8 0.120
4 8 0.020
5 8 0.002
6 8 0.000
7 8 0.000
8 8 0.754
9 8 0.039
10 8 0.012
11 8 0.000
12 8 0.000
13 8 0.004
14 8 0.001
15 8 0.000
16 8 0.000
17 8 0.000
18 8 0.000

0 9 0.015
1 9 0.021
2 9 0.000
3 9 0.276
4 9 0.002
5 9 0.015
6 9 0.000
7 9 0.000
8 9 0.007
9 9 0.644
10 9 0.000
11 9 0.000
12 9 0.000
13 9 0.001
14 9 0.019
15 9 0.000
16 9 0.000
17 9 0.000
18 9 0.000

0 10 0.000
1 10 0.000
2 10 0.038
3 10 0.000
4 10 0.000
5 10 0.000
6 10 0.000
7 10 0.000
8 10 0.008
9 10 0.000
10 10 0.954
11 10 0.000
12 10 0.000
13 10 0.000
14 10 0.000
15 10 0.000
16 10 0.000
17 10 0.000
18 10 0.000

0 11 0.026
1 11 0.023
2 11 0.073
3 11 0.183
4 11 0.155
5 11 0.028
6 11 0.000
7 11 0.000
8 11 0.019
9 11 0.032
10 11 0.001
11 11 0.000
12 11 0.000
13 11 0.449
14 11 0.010
15 11 0.000
16 11 0.000
17 11 0.000
18 11 0.000

0 12 0.025
1 12 0.055
2 12 0.081
3 12 0.108
4 12 0.144
5 12 0.017
6 12 0.000
7 12 0.000
8 12 0.114
9 12 0.087
10 12 0.012
11 12 0.000
12 12 0.000
13 12 0.354
14 12 0.002
15 12 0.000
16 12 0.000
17 12 0.000
18 12 0.000

0 13 0.005
1 13 0.007
2 13 0.040
3 13 0.015
4 13 0.017
5 13 0.002
6 13 0.000
7 13 0.000
8 13 0.008
9 13 0.005
10 13 0.003
11 13 0.000
12 13 0.000
13 13 0.893
14 13 0.005
15 13 0.000
16 13 0.000
17 13 0.000
18 13 0.000

0 14 0.002
1 14 0.006
2 14 0.287
3 14 0.025
4 14 0.187
5 14 0.016
6 14 0.000
7 14 0.000
8 14 0.025
9 14 0.002
10 14 0.004
11 14 0.000
12 14 0.000
13 14 0.442
14 14 0.004
15 14 0.000
16 14 0.000
17 14 0.000
18 14 0.000

0 15 0.001
1 15 0.000
2 15 0.301
3 15 0.047
4 15 0.208
5 15 0.000
6 15 0.000
7 15 0.000
8 15 0.079
9 15 0.000
10 15 0.000
11 15 0.000
12 15 0.000
13 15 0.353
14 15 0.011
15 15 0.000
16 15 0.000
17 15 0.000
18 15 0.000

0 16 0.000
1 16 0.000
2 16 0.000
3 16 0.000
4 16 0.000
5 16 0.000
6 16 0.000
7 16 0.000
8 16 0.000
9 16 0.000
10 16 0.000
11 16 0.000
12 16 0.000
13 16 0.000
14 16 0.000
15 16 0.000
16 16 0.000
17 16 0.000
18 16 0.000

0 17 0.004
1 17 0.017
2 17 0.008
3 17 0.321
4 17 0.045
5 17 0.003
6 17 0.000
7 17 0.000
8 17 0.010
9 17 0.006
10 17 0.000
11 17 0.000
12 17 0.000
13 17 0.571
14 17 0.014
15 17 0.000
16 17 0.000
17 17 0.000
18 17 0.000

0 18 0.066
1 18 0.074
2 18 0.001
3 18 0.293
4 18 0.032
5 18 0.079
6 18 0.000
7 18 0.000
8 18 0.007
9 18 0.121
10 18 0.000
11 18 0.000
12 18 0.000
13 18 0.310
14 18 0.017
15 18 0.000
16 18 0.000
17 18 0.000
18 18 0.000

    }; %
    \end{axis}
\end{tikzpicture}
        \caption{\MethodName} \label{tab:confusion_matrix_2d:a}
    \end{subfigure}\vspace{0.4em}
    \begin{subfigure}{\columnwidth}
        \begin{tikzpicture}[every node/.style={font=\sffamily\scriptsize},]
    \begin{axis}[
        axis equal image,
        colormap={bluewhite}{color=(white) rgb255=(90,96,191)},
        xticklabels={road, sidewalk, building, wall, fence, pole, traffic light, traffic sign, vegetation, terrain, sky, person, rider, car, truck, bus, train, motorcycle, bicycle}, %
        xtick={0,...,18}, %
        xtick style={draw=none},
        axis y line*=right, 
        yticklabel pos=right,
        yticklabels={road, sidewalk, building, wall, fence, pole, traffic light, traffic sign, vegetation, terrain, sky, person, rider, car, truck, bus, train, motorcycle, bicycle}, %
        ytick={0,...,18}, %
        ytick style={draw=none},
        tick label style={font=\sffamily\scriptsize},
        enlargelimits=false,
        colorbar,
        colorbar style={
            width=5pt,
            ytick={0, 0.2, 0.4, 0.6, 0.8, 1},
            yticklabels={0, 0.2, 0.4, 0.6, 0.8, 1},
        },
        xticklabel style={
          rotate=90
        },
        point meta max=1.0,
        point meta min=0,
        colormap={svbcolorbar}{[1pt]
          rgb(0pt)=(1,1,1);
          rgb(200pt)=(0.9803921568627451,0.01568627450980392,0.01568627450980392)
        },
    ]
    \addplot[
        matrix plot,
        mesh/cols=19, %
        point meta=explicit,draw=gray
    ] table [meta=C] {
    
x y C
0 0 0.972
1 0 0.011
2 0 0.000
3 0 0.001
4 0 0.000
5 0 0.000
6 0 0.000
7 0 0.000
8 0 0.000
9 0 0.000
10 0 0.000
11 0 0.000
12 0 0.000
13 0 0.014
14 0 0.001
15 0 0.000
16 0 0.000
17 0 0.000
18 0 0.000

0 1 0.842
1 1 0.082
2 1 0.000
3 1 0.020
4 1 0.000
5 1 0.001
6 1 0.000
7 1 0.000
8 1 0.004
9 1 0.002
10 1 0.000
11 1 0.000
12 1 0.000
13 1 0.014
14 1 0.002
15 1 0.000
16 1 0.000
17 1 0.013
18 1 0.019

0 2 0.011
1 2 0.003
2 2 0.689
3 2 0.004
4 2 0.046
5 2 0.005
6 2 0.000
7 2 0.002
8 2 0.052
9 2 0.000
10 2 0.070
11 2 0.000
12 2 0.001
13 2 0.013
14 2 0.008
15 2 0.087
16 2 0.000
17 2 0.009
18 2 0.000

0 3 0.231
1 3 0.018
2 3 0.065
3 3 0.173
4 3 0.028
5 3 0.070
6 3 0.000
7 3 0.015
8 3 0.107
9 3 0.002
10 3 0.020
11 3 0.000
12 3 0.013
13 3 0.066
14 3 0.042
15 3 0.034
16 3 0.000
17 3 0.112
18 3 0.003

0 4 0.109
1 4 0.010
2 4 0.055
3 4 0.048
4 4 0.270
5 4 0.066
6 4 0.000
7 4 0.018
8 4 0.179
9 4 0.007
10 4 0.007
11 4 0.000
12 4 0.021
13 4 0.144
14 4 0.027
15 4 0.002
16 4 0.000
17 4 0.037
18 4 0.000

0 5 0.047
1 5 0.006
2 5 0.275
3 5 0.009
4 5 0.029
5 5 0.281
6 5 0.000
7 5 0.001
8 5 0.259
9 5 0.008
10 5 0.017
11 5 0.000
12 5 0.003
13 5 0.046
14 5 0.006
15 5 0.004
16 5 0.000
17 5 0.007
18 5 0.000

0 6 0.013
1 6 0.000
2 6 0.137
3 6 0.000
4 6 0.000
5 6 0.081
6 6 0.000
7 6 0.000
8 6 0.766
9 6 0.000
10 6 0.000
11 6 0.000
12 6 0.000
13 6 0.003
14 6 0.000
15 6 0.000
16 6 0.000
17 6 0.000
18 6 0.000

0 7 0.016
1 7 0.000
2 7 0.622
3 7 0.002
4 7 0.041
5 7 0.103
6 7 0.000
7 7 0.002
8 7 0.152
9 7 0.002
10 7 0.009
11 7 0.000
12 7 0.002
13 7 0.044
14 7 0.001
15 7 0.003
16 7 0.000
17 7 0.001
18 7 0.000

0 8 0.013
1 8 0.001
2 8 0.024
3 8 0.002
4 8 0.001
5 8 0.006
6 8 0.000
7 8 0.000
8 8 0.792
9 8 0.008
10 8 0.007
11 8 0.000
12 8 0.090
13 8 0.012
14 8 0.007
15 8 0.000
16 8 0.001
17 8 0.034
18 8 0.000

0 9 0.176
1 9 0.028
2 9 0.000
3 9 0.019
4 9 0.000
5 9 0.002
6 9 0.000
7 9 0.000
8 9 0.105
9 9 0.540
10 9 0.000
11 9 0.002
12 9 0.017
13 9 0.006
14 9 0.006
15 9 0.000
16 9 0.002
17 9 0.096
18 9 0.000

0 10 0.000
1 10 0.000
2 10 0.024
3 10 0.000
4 10 0.000
5 10 0.001
6 10 0.000
7 10 0.000
8 10 0.020
9 10 0.000
10 10 0.956
11 10 0.000
12 10 0.000
13 10 0.000
14 10 0.000
15 10 0.000
16 10 0.000
17 10 0.000
18 10 0.000

0 11 0.096
1 11 0.011
2 11 0.035
3 11 0.003
4 11 0.004
5 11 0.059
6 11 0.000
7 11 0.000
8 11 0.090
9 11 0.001
10 11 0.002
11 11 0.000
12 11 0.010
13 11 0.565
14 11 0.096
15 11 0.001
16 11 0.000
17 11 0.026
18 11 0.000

0 12 0.127
1 12 0.001
2 12 0.035
3 12 0.003
4 12 0.000
5 12 0.058
6 12 0.000
7 12 0.000
8 12 0.188
9 12 0.004
10 12 0.007
11 12 0.000
12 12 0.020
13 12 0.494
14 12 0.063
15 12 0.000
16 12 0.000
17 12 0.000
18 12 0.000

0 13 0.016
1 13 0.001
2 13 0.014
3 13 0.000
4 13 0.002
5 13 0.002
6 13 0.000
7 13 0.000
8 13 0.013
9 13 0.000
10 13 0.004
11 13 0.000
12 13 0.001
13 13 0.932
14 13 0.013
15 13 0.001
16 13 0.000
17 13 0.001
18 13 0.000

0 14 0.022
1 14 0.002
2 14 0.160
3 14 0.009
4 14 0.022
5 14 0.022
6 14 0.000
7 14 0.011
8 14 0.054
9 14 0.000
10 14 0.004
11 14 0.000
12 14 0.001
13 14 0.580
14 14 0.062
15 14 0.040
16 14 0.000
17 14 0.010
18 14 0.000

0 15 0.050
1 15 0.001
2 15 0.259
3 15 0.000
4 15 0.010
5 15 0.008
6 15 0.000
7 15 0.000
8 15 0.093
9 15 0.000
10 15 0.000
11 15 0.000
12 15 0.001
13 15 0.469
14 15 0.095
15 15 0.009
16 15 0.000
17 15 0.005
18 15 0.000

0 16 0.000
1 16 0.000
2 16 0.000
3 16 0.000
4 16 0.000
5 16 0.000
6 16 0.000
7 16 0.000
8 16 0.000
9 16 0.000
10 16 0.000
11 16 0.000
12 16 0.000
13 16 0.000
14 16 0.000
15 16 0.000
16 16 0.000
17 16 0.000
18 16 0.000

0 17 0.035
1 17 0.057
2 17 0.007
3 17 0.001
4 17 0.000
5 17 0.004
6 17 0.000
7 17 0.000
8 17 0.068
9 17 0.002
10 17 0.000
11 17 0.000
12 17 0.002
13 17 0.734
14 17 0.060
15 17 0.000
16 17 0.000
17 17 0.030
18 17 0.000

0 18 0.182
1 18 0.019
2 18 0.002
3 18 0.013
4 18 0.000
5 18 0.031
6 18 0.000
7 18 0.000
8 18 0.026
9 18 0.004
10 18 0.000
11 18 0.000
12 18 0.023
13 18 0.611
14 18 0.058
15 18 0.004
16 18 0.000
17 18 0.027
18 18 0.000

    }; %
    \end{axis}
\end{tikzpicture}
        \caption{STEGO} \label{tab:confusion_matrix_2d:b}
    \end{subfigure}
    \vspace{-1.1em}
    \caption{\textbf{Confusion matrices for 2D unsupervised semantic segmentation on KITTI-360}. Rows represent ground-truth class labels (normalized to 1), while columns correspond to predicted class labels. We report results for \emph{\subref{tab:confusion_matrix_2d:a}} \MethodName and \emph{\subref{tab:confusion_matrix_2d:b}} STEGO on the SSCBench-KITTI-360 test split.}
    \label{tab:confusion_matrix_2d}
    \vspace{-0.6em}
\end{figure}

\inparagraph{Quantitative results.} In \cref{tab:sscbenchfull}, we provide additional semantic scene completion results of 3D-supervised approaches as an additional point of comparison. In particular, we report official SSCBench~\cite{Li:2024:SSC} results of VoxFormer-S~\cite{Li:2023:VSV} and OccFormer~\cite{Zhang:2023:OCC}. Both utilize 3D supervision, including both semantic and geometric annotations. We also report the results of SSCNet~\cite{Song:2017:SSC}. This approach trains using 3D supervision but utilizes a depth image during inference. While \MethodName achieves state-of-the-art segmentation accuracy in the unsupervised setting, supervised approaches are significantly more accurate.

\begin{table}[t]
    \centering
    \caption{\textbf{Linear probing results on SSCBench-KITTI-360.} We extend \cref{tab:probing} and report detailed results of \MethodName using 2D-supervised linear probing. Semantic results using mIoU and class IoU, and geometric results using IoU, Precision, and Recall, and (all in \%, $\uparrow$) on SSCBench-KITTI-360 test using three depth ranges.}
    \vspace{-0.3em}
    \renewcommand\tabcolsep{6.22pt}
\scriptsize
\renewcommand{\arraystretch}{0.865}
\begin{tabularx}{\columnwidth}{lS[table-format=2.2]S[table-format=2.2]S[table-format=2.2]|S[table-format=2.2]S[table-format=2.2]S[table-format=2.2]}
\toprule
\textbf{Method} & \multicolumn{3}{c|}{\textbf{\MethodName{} {\tiny{}w/ DINO} (Ours)}} & \multicolumn{3}{c}{\textbf{\MethodName{} {\tiny{}w/ DINOv2} (Ours)}} \\
\midrule
\textbf{Supervision} & \multicolumn{6}{c}{\cellcolor{LimeGreen!70}\textbf{Unsupervised}} \\
\midrule
\textbf{Range} & {\SI{12.8}{\meter}} & {\SI{25.6}{\meter}} & {\SI{51.2}{\meter}} & {\SI{12.8}{\meter}} & {\SI{25.6}{\meter}} & {\SI{51.2}{\meter}} \\
\specialrule{0.5pt}{1.5pt}{2pt}
\multicolumn{7}{c}{\textit{Semantic validation}} \\
\specialrule{0.5pt}{1.5pt}{2pt}
\cellcolor{tud0c!20}\textbf{mIoU} & \cellcolor{tud0c!20}13.63 & \cellcolor{tud0c!20}12.07 & \cellcolor{tud0c!20}9.34 & \cellcolor{tud0c!20}\bfseries 15.85 & \cellcolor{tud0c!20}\bfseries 13.70 & \cellcolor{tud0c!20}\bfseries 10.57 \\
car           &  16.77 & 12.37 &  8.42  &  20.35 & 15.04 & 10.16  \\
bicycle       &   1.10 &  0.70 &  0.47  &   0.00 &  0.00 &  0.00  \\
motorcycle    &   0.00 &  0.00 &  0.00  &   0.00 &  0.00 &  0.00  \\
truck         &   3.80 &  2.21 &  1.52  &  11.48 &  7.46 &  4.63  \\
other-v.      &   0.13 &  0.08 &  0.06  &   0.00 &  0.00 &  0.00  \\
person        &   0.01 &  0.00 &  0.00  &   0.00 &  0.00 &  0.00  \\
road          &  66.63 & 62.21 & 49.99  &  69.92 & 63.06 & 50.49  \\
sidewalk      &  29.46 & 25.17 & 18.85  &  42.35 & 37.13 & 29.13  \\
building      &  18.64 & 22.82 & 17.66  &  23.03 & 27.05 & 21.40  \\
fence         &   9.29 &  6.03 &  3.96  &   8.82 &  6.40 &  4.61  \\
vegetation    &  32.76 & 26.49 & 20.89  &  30.42 & 24.96 & 19.75  \\
terrain       &  24.80 & 22.43 & 18.00  &  30.73 & 23.85 & 17.93  \\
pole          &   0.25 &  0.24 &  0.14  &   0.46 &  0.40 &  0.28  \\
traffic-sign  &   0.50 &  0.17 &  0.09  &   0.00 &  0.00 &  0.00  \\
other-obj.    &   0.26 &  0.07 &  0.04  &   0.00 &  0.00 &  0.00  \\

\specialrule{0.5pt}{1.5pt}{2pt}
\multicolumn{7}{c}{\textit{Geometric validation}} \\
\specialrule{0.5pt}{1.5pt}{2pt}
\cellcolor{tud0c!20}\textbf{IoU}           & \cellcolor{tud0c!20}49.34 & \cellcolor{tud0c!20}42.26 & \cellcolor{tud0c!20}37.61 & \cellcolor{tud0c!20}\bfseries 49.77 & \cellcolor{tud0c!20}\bfseries 43.19 & \cellcolor{tud0c!20}\bfseries 38.55 \\
Precision     &  52.83 & 45.95 & 41.55  &  52.76 & 46.46 & 42.11  \\
Recall        &  88.21 & 84.05 & 79.88  &  89.76 & 85.99 & 82.02  \\
\bottomrule
\end{tabularx}

    \label{tab:probingresultsfull}
    \vspace{-0.6em}
\end{table}

\cref{tab:sscbenchfull} provides additional SSC results of our S4C~\cite{Hayler:2024:S4C} + STEGO~\cite{Hamilton:2022:USS} baseline and \MethodName using DINOv2 features~\cite{Oquab:2023:DLR}. In particular, we train STEGO with DINOv2 features and lift the resulting unsupervised semantic predictions using S4C. For \MethodName, we use DINOv2 target features and perform distillation and clustering. Training S4C + STEGO using DINOv2 features leads to improvements for close range (\SI{12.8}{m}) over using DINO features (\cf \cref{tab:sscbenchfull}). For larger ranges (\eg, \SI{51.2}{m}), S4C + STEGO with DINOv2 features drops in accuracy compared to S4C + STEGO with DINO features. We attribute this drop in accuracy to the coarser feature resolution of DINOv2 (larger ViT patch size). This behavior has also been observed for the task of 2D unsupervised semantic segmentation~\cite{Hahn:2024:BUS}. Note that \MethodName overcomes the coarse features using a learnable downsampler and multi-view training, learning high-resolution 3D features.

\inparagraph{Class-wise semantic results.} To further assess the segmentation accuracy of \MethodName, we report the class-wise IoU metric in 3D (\cf \cref{tab:sscbench}, \ref{tab:sscbenchfull}, and \ref{tab:sscbenchdinov2}) and 2D (\cf \cref{tab:semantic_segmentation_2d_results_classes}). We generally observe that \MethodName performs well in segmenting frequent classes, such as ``road'', ``building'', and ``sky''. Less frequent classes, such as ``fence'' and ``pole'', are less well segmented. Classes including very small and fine structures (\eg, ``pole'') are completely missed by \MethodName. This trend can also be observed for our 3D unsupervised baseline S4C + STEGO and 2D STEGO. We also observe that class-wise metrics strongly correlate between 2D and 3D.

\Cref{tab:confusion_matrix_2d} reports confusion matrices of \MethodName and STEGO for 2D semantic segmentation on KITTI-360. Both approaches share a similar confusion pattern. We attribute this to the fact that both approaches rely on the feature representation of DINO. In particular, we observe confusion between semantically close classes, such as ``pole'', ``traffic light'', and ``traffic sign''. Interestingly, for the semantic classes ``person'', ``rider'', ``car'', ``truck'', ``bus'', ``motorcycle'', and ``bicycle'', we see a strong confusion. We suspect this correlation is potentially caused by the fact that these classes often appear on the ``road'' and ``sidewalk'' and are rare in KITTI-360.

We also provide class-wise SSC results of \MethodName using 2D-supervised linear probing in \cref{tab:probingresultsfull}. Linear probing provides an upper bound for clustering our features, improving the segmentation accuracy for almost all classes. However, rare classes like ``motorcycle'' are still not captured using linear probing. This suggests that the DINO feature space fails to express these classes accurately, limiting the segmentation accuracy of \MethodName. Still, our approach is agnostic to the utilized target features and can potentially profit from better 2D features.

\inparagraph{Camera pose analysis.} Training \MethodName requires accurate camera poses. While KITTI-360 offers ground-truth camera poses, these poses are obtained using additional cues, including LiDAR data~\cite{Liao:2023:KND}. To adhere to our fully unsupervised setting, we provide results training with unsupervised camera poses, estimated using stereo visual SLAM. In particular, \cref{tab:analysis} reports results of \MethodName trained using unsupervised camera poses estimated by ORB-SLAM3~\cite{Campos:2021:ORB}. \Cref{tab:posesresults} extends this and reports detailed SSC results using two different unsupervised stereo visual SLAM approaches---SOFT2~\cite{soft2} and ORB-SLAM3~\cite{Campos:2021:ORB}. Using unsupervised and visually estimated poses leads to a minor drop in both semantic and geometric SSC validation. While ORB-SLAM3 poses lead to slightly better semantic accuracy than SOFT2 poses, SOFT2 estimated poses result in higher geometric accuracy. Still, both SOFT2 and ORB-SLAM3 provide poses accurate enough for training \MethodName, reaching a similar accuracy to employing KITTI-360 ground-truth poses.

\inparagraph{Out-of-domain results.} We illustrate results for out-of-domain prediction in \cref{fig:features_2d_ood}. While our \MethodName model is trained on the KITTI-360 dataset, we still obtain plausible features when inferring 2D features for vastly different scenes. The 2D rendered features still show a strong correlation with semantically uniform regions, showcasing the generalization of our feature field.

\begin{table}[t]
    \centering
    \caption{\textbf{Camera pose analysis on SSCBench-KITTI-360.} We extend the camera pose analysis in \cref{tab:analysis} and report detailed results of \MethodName with unsupervised camera poses estimated by SOFT2~\cite{soft2} and ORB-SLAM3~\cite{Campos:2021:ORB}. For reference, we also provide results obtained using the KITTI-360 ground-truth poses. Semantic results using mIoU and class IoU, and geometric results using IoU, Precision, and Recall, and (all in \%, $\uparrow$) on SSCBench-KITTI-360 test using three depth ranges.}
    \vspace{-0.3em}
    \renewcommand\tabcolsep{1.76pt}
\scriptsize
\renewcommand{\arraystretch}{0.865}
\begin{tabularx}{\columnwidth}{lS[table-format=2.2]S[table-format=2.2]S[table-format=2.2]|S[table-format=2.2]S[table-format=2.2]S[table-format=2.2]|S[table-format=2.2]S[table-format=2.2]S[table-format=2.2]}
\toprule
\textbf{Method} & \multicolumn{9}{c}{\textbf{\MethodName (Ours)}} \\
\midrule
\textbf{Poses} & \multicolumn{3}{c|}{\textbf{SOFT2}} & \multicolumn{3}{c|}{\textbf{ORB-SLAM3}} & \multicolumn{3}{c}{\textbf{KITTI-360 (GT)}} \\
\midrule
\textbf{Range} & {\SI{12.8}{\meter}} & {\SI{25.6}{\meter}} & {\SI{51.2}{\meter}} & {\SI{12.8}{\meter}} & {\SI{25.6}{\meter}} & {\SI{51.2}{\meter}} & {\SI{12.8}{\meter}} & {\SI{25.6}{\meter}} & {\SI{51.2}{\meter}} \\
\specialrule{0.5pt}{1.5pt}{2pt}
\multicolumn{10}{c}{\textit{Semantic validation}} \\
\specialrule{0.5pt}{1.5pt}{2pt}
\cellcolor{tud0c!20}\textbf{mIoU} & \cellcolor{tud0c!20}10.58 & \cellcolor{tud0c!20}9.58 & \cellcolor{tud0c!20}7.72 & \cellcolor{tud0c!20}\bfseries 10.88 & \cellcolor{tud0c!20}9.86 & \cellcolor{tud0c!20}7.88 & \cellcolor{tud0c!20}10.76 & \cellcolor{tud0c!20}\bfseries 10.01 & \cellcolor{tud0c!20}\bfseries 8.00 \\
car           & 18.47 & 13.98 & 10.44 & 19.37 & 14.09 &  9.72 & 21.24 & 15.94 & 11.21 \\
bicycle       &  0.04 &  0.03 &  0.03 &  0.06 &  0.03 &  0.02 & 0.00 & 0.00 & 0.00 \\
motorcycle    &  0.00 &  0.00 &  0.00 &  0.01 &  0.01 &  0.00 & 0.00 & 0.00& 0.00 \\
truck         &  0.00 &  0.00 &  0.00 &  0.05 &  0.02 &  0.01 & 0.00 & 0.00& 0.00 \\
other-v.      &  0.01 &  0.02 &  0.04 &  0.08 &  0.06 &  0.05 & 0.00 & 0.00& 0.00 \\
person        &  0.02 &  0.01 &  0.01 &  0.00 &  0.00 &  0.00 & 0.00 & 0.00& 0.00 \\
road          & 44.48 & 44.50 & 36.06 & 44.74 & 40.58 & 31.86 & 51.10 & 49.12 & 39.82 \\
sidewalk      & 16.55 & 16.79 & 14.38 & 21.45 & 23.56 & 19.88 & 20.26 & 22.31 & 18.97 \\
building      & 19.40 & 23.40 & 18.56 & 19.19 & 24.87 & 20.02 & 12.33 & 18.27 & 14.32 \\
fence         &  1.79 &  1.00 &  0.68 &  1.62 &  1.21 &  0.91 & 1.91 & 0.90 & 0.58 \\
vegetation    & 32.10 & 25.65 & 20.67 & 32.60 & 24.91 & 19.49 & 31.22 & 25.57 & 19.85 \\
terrain       & 25.59 & 18.11 & 14.79 & 23.98 & 18.41 & 16.16 & 23.26 & 18.02 & 15.22 \\
pole          &  0.18 &  0.11 &  0.09 &  0.00 &  0.00 &  0.00 & 0.05 & 0.05 & 0.05 \\
traffic-sign  &  0.00 &  0.01 &  0.00 &  0.03 &  0.03 &  0.02 & 0.00 & 0.00 & 0.00 \\
other-obj.    &  0.08 &  0.05 &  0.03 &  0.08 &  0.05 &  0.03 & 0.00 & 0.00 & 0.00 \\
\specialrule{0.5pt}{1.5pt}{2pt}
\multicolumn{10}{c}{\textit{Geometric validation}} \\
\specialrule{0.5pt}{1.5pt}{2pt}
\cellcolor{tud0c!20}\textbf{IoU} & \cellcolor{tud0c!20}\bfseries 49.91 & \cellcolor{tud0c!20}41.85 & \cellcolor{tud0c!20}37.25 & \cellcolor{tud0c!20}45.42 & \cellcolor{tud0c!20}40.21 & \cellcolor{tud0c!20}36.65 & \cellcolor{tud0c!20}49.54 & \cellcolor{tud0c!20}\bfseries 42.27 & \cellcolor{tud0c!20}\bfseries 37.60 \\
Precision     & 54.74 & 45.66 & 40.79 & 54.42 & 45.54 & 40.98 & 53.27 & 46.10 & 41.59 \\
Recall        & 84.98 & 83.40 & 81.12 & 73.33 & 77.46 & 77.62 & 87.61 & 83.59 & 79.67 \\
\bottomrule
\end{tabularx}

    \label{tab:posesresults}
    \vspace{-0.6em}
\end{table}

\section{Limitations and Future Work \label{supp:limitations}}

\paragraph{Target features.} Our method builds on DINO~\cite{Caron:2021:EPS} to obtain target features. While we learn to lift these features into 3D and improve multi-view feature consistency, we cannot improve the discriminative power of the target features \emph{per se}. However, \MethodName can be trained using arbitrary 2D target features and can profit from future advances in SSL representations. Note that training \MethodName requires only 2 days on a single GPU and our training transfers seamlessly to different target features (\eg, DINOv2), thus, utilizing \MethodName differently is straightforward.

\inparagraph{Dynamic objects.} Our loss does not model dynamic objects and relies on a static scene assumption. This can potentially cause inaccurate predictions for dynamic classes such as \emph{person} in our experiments. Recent works in depth estimation have explicitly modeled the probability of areas being dynamic~\cite{prodepth} and even their motion within the scene~\cite{dynamodepth}, which might be extended to \MethodName. 

\inparagraph{View sampling and camera poses.} For sampling views during training, we rely on the sampling scheme of S4C~\cite{Hayler:2024:S4C}. This is not directly applicable to other non-driving datasets, where the sampling needs to be tuned. In addition, our approach requires accurate camera poses for each view. We demonstrated that these can be obtained in an unsupervised way for KITTI-360 (\cf \cref{tab:analysis} \& \cref{tab:posesresults}). However, obtaining unsupervised camera poses in more complex scenarios and conditions is still a challenge~\cite{slam}.

\begin{figure}[t]
    \centering
    \scriptsize
\sffamily
\setlength{\tabcolsep}{0pt}
\renewcommand{\arraystretch}{0.66}
\begin{tabular}{>{\centering\arraybackslash} m{0.5\columnwidth} 
                >{\centering\arraybackslash} m{0.5\columnwidth}}

\textbf{Input Image} & \textbf{\MethodName} \\[2pt]

\includegraphics[width=0.5\columnwidth]{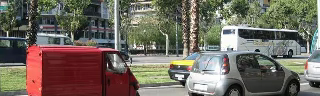} &
\includegraphics[width=0.5\columnwidth]{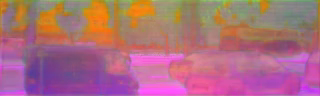} \\[-1.5pt]

\includegraphics[width=0.5\columnwidth]{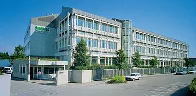} &
\includegraphics[width=0.5\columnwidth]{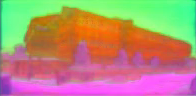} \\[-1.5pt]

\includegraphics[width=0.5\columnwidth]{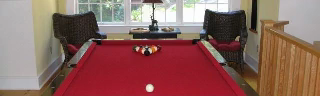} &
\includegraphics[width=0.5\columnwidth]{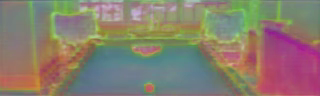} \\

\end{tabular}

    \vspace{-0.3em}
    \caption{\textbf{2D \MethodName features on out-of-domain images.} We visualize our 2D rendered features \emph{(right)} given an out-of-domain image \emph{(left)} from ADE20K~\cite{ade20k}. We use the first three principal components for feature visualization. While not trained on such scenes, \MethodName still produces plausible feature maps.}
    \label{fig:features_2d_ood}
    \vspace{-0.6em}
\end{figure}

\inparagraph{Future work.} \MethodName is only trained using a single dataset to be comparable to existing SSC approaches. However, scaling our approach to multiple datasets of more variable scenes could lead to more general feature representations. Ultimately, scaling \MethodName to internet-scale videos might enable strong zero-shot and cross-domain 3D scene understanding.

\end{document}